\documentclass[11pt]{memtensor}
\usepackage{enumitem}
\usepackage[utf8]{inputenc}
\usepackage{url}
\usepackage{graphicx}
\usepackage{booktabs}
\usepackage{amsfonts}
\usepackage{nicefrac}
\usepackage{makecell}
\usepackage{microtype}
\usepackage{amsmath}
\usepackage{etoolbox}
\usepackage{lipsum}
\usepackage{minitoc}
\usepackage{tablefootnote}
\usepackage{threeparttable}
\usepackage{wrapfig}
\usepackage{appendix}
\usepackage{multirow}
\usepackage{ulem}
\useunder{\uline}{\ul}{}
\usepackage{colortbl}
\usepackage{adjustbox}
\usepackage{array}
\usepackage{tabularx}
\usepackage{siunitx}
\usepackage{placeins}
\usepackage{float}
\usepackage{xcolor}
\usepackage{mdframed}
\usepackage{listings}
\usepackage{bigstrut}
\usepackage{amsthm}

\usepackage{newtxtext,newtxmath}
\usepackage[most]{tcolorbox}
\usepackage{microtype}
\usepackage{setspace}

\usepackage[table]{xcolor}

\usepackage{booktabs}
\usepackage[table]{xcolor}
\usepackage{siunitx}
\usepackage{adjustbox}
\usepackage{threeparttable}

\definecolor{MetisBlue}{HTML}{6F8FEF}
\definecolor{MetisBlueDark}{HTML}{536FC8}
\definecolor{MetisBlueLight}{HTML}{EAF1FF}
\definecolor{MetisBluePale}{HTML}{F6F9FF}
\definecolor{MetisBlueRule}{HTML}{C9D7F6}
\definecolor{MetisHighlight}{HTML}{EDF3FF}
\definecolor{MetisHighlightStrong}{HTML}{DFE9FF}
\definecolor{MetisText}{HTML}{263247}

\definecolor{MetisHeader}{HTML}{DCE6F0}
\definecolor{MetisHeaderText}{HTML}{31475E}

\definecolor{InsightBlue}{HTML}{2F5D7C}
\definecolor{InsightBg}{HTML}{F3F6F8}
\definecolor{InsightText}{HTML}{24313A}

\definecolor{MetisDeltaDown}{HTML}{2E8B57}
\definecolor{MetisDeltaUp}{HTML}{C94C4C}

\newcommand{\increase}[1]{%
    \multicolumn{1}{c}{%
        \textcolor{MetisDeltaUp}{\num{+#1}}%
    }%
}

\newtcolorbox{keyinsight}[1][Key insight]{
  enhanced,
  breakable,
  colback=InsightBg,
  colframe=InsightBlue,
  coltext=InsightText,
  boxrule=0pt,
  leftrule=2.8pt,
  arc=1.2pt,
  outer arc=1.2pt,
  left=13pt,
  right=13pt,
  top=9pt,
  bottom=10pt,
  before skip=11pt,
  after skip=11pt,
  title={\footnotesize\bfseries\color{InsightBlue}\MakeUppercase{#1}},
  attach title to upper={\par\vspace{4pt}},
  fontupper=\normalsize,
}

\DeclareMathOperator{\Softmax}{Softmax}
\DeclareMathOperator{\Norm}{Norm}

\title{
  \titlefont
  \parbox{\textwidth}{
    \centering
    \LARGE
    \raisebox{-0.4em}{\includegraphics[height=1.6em]{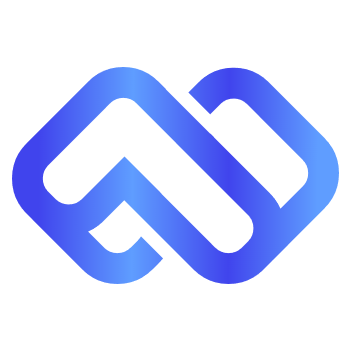}}\hspace{0.15em}
    MemT-1.0: Towards Memory-Native Evolution
  }
}

\author[]{\makebox[\linewidth][c]{MemTensor $\times$ NVIDIA}}

\abstract{
Memory-native evolution requires making memory an intrinsic capability of foundation models, enabling them to continuously accumulate and utilize information.
To this end, we introduce \textbf{MemT-1.0}, a new architecture of memory foundation models that internalize memory as part of model computation.
Building on Metis, MemT-1.0 employs Gated DeltaNet-2 with a stable token-level memory update mechanism within each turn. 
We further introduce Mixture-of-Storage to scale native memory capacity.
Following a deliberate backbone preparation, we develop a two-stage training pipeline consisting of continued pre-training and post-training, supported by 34B tokens of training data and efficient training and inference infrastructure.
We release a model family comprising \textbf{MemT-1.0-Nano}, \textbf{MemT-1.0-Flash}, and \textbf{MemT-1.0-Pro}, and conduct a comprehensive evaluation of memory and general capabilities.
The results show that they outperform the parametric memory baselines on average and remain competitive with retrieval-augmented methods, while largely preserving general capabilities.
At a history length of 128K tokens, MemT-1.0-Nano achieves approximately $1.86\times$ end-to-end inference speedup and a $66\times$ reduction in persistent storage relative to full context KV cache.
By internalizing retrieved procedural experience, MemT-1.0 improves task success on agentic tasks, providing a practical step toward memory-native evolution.
}

\checkdata[\raisebox{-0.7ex}{\includegraphics[height=1.05em]{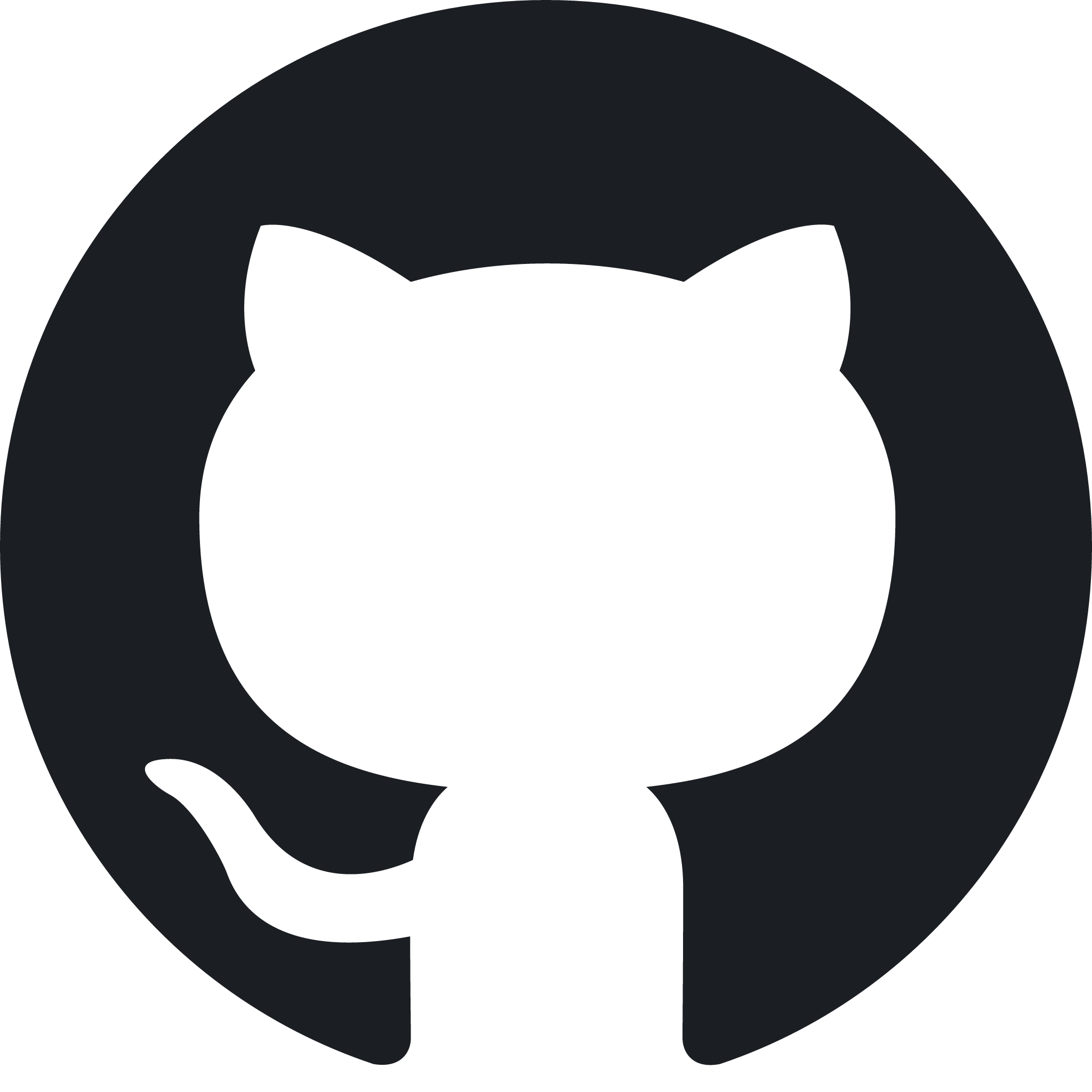}}\hspace{0.3em}Code]{%
  \url{https://github.com/MemTensor/MemT-1.0}%
}
\checkdata[\raisebox{-0.7ex}{\includegraphics[height=1.05em]{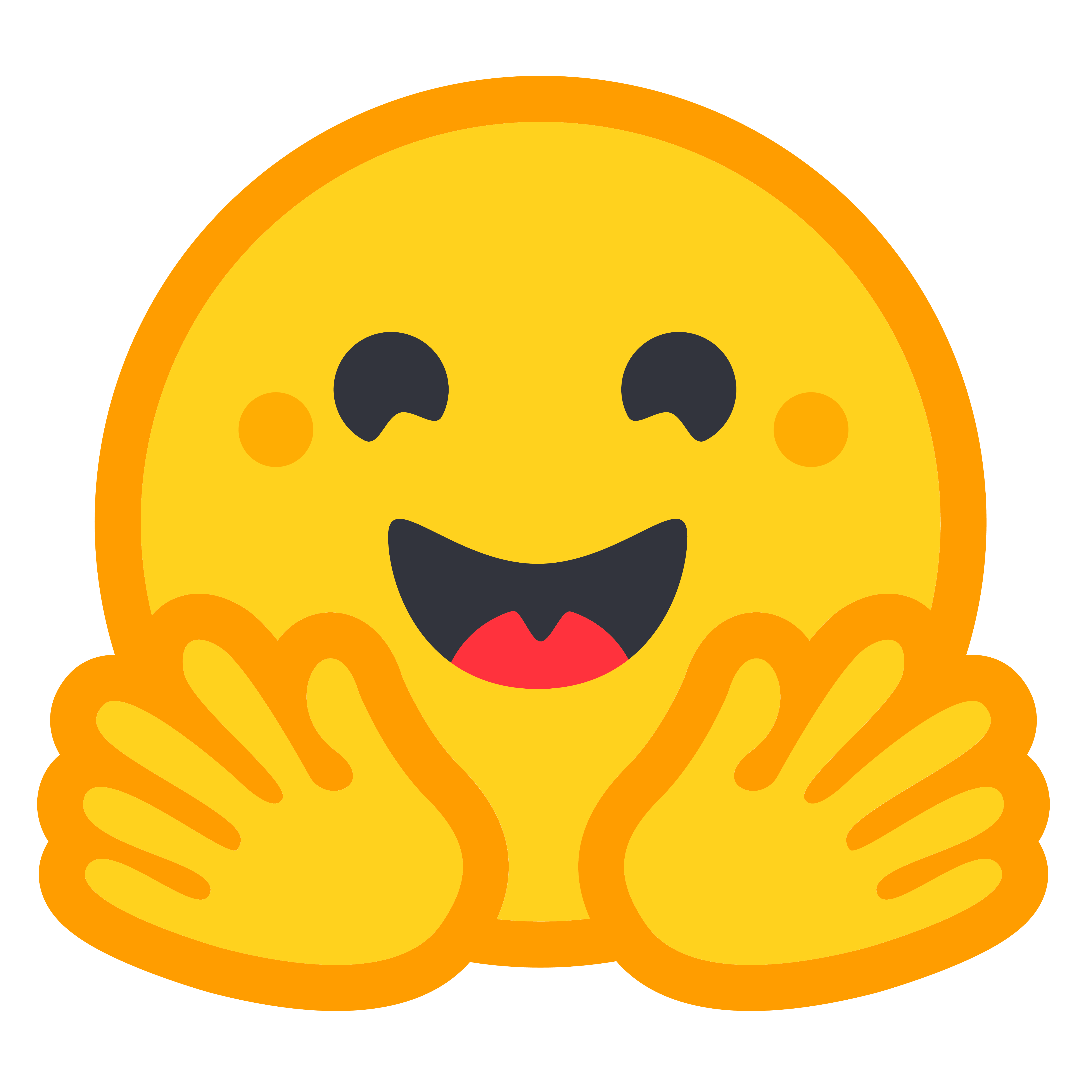}}\hspace{0.3em}Model]{%
\url{https://huggingface.co/collections/MemTensor/memt-1}%
}

\theoremstyle{definition}

\begin{document}
\maketitle

\begin{center}
    \vspace{-0.4cm}
    \includegraphics[width=\textwidth]
        {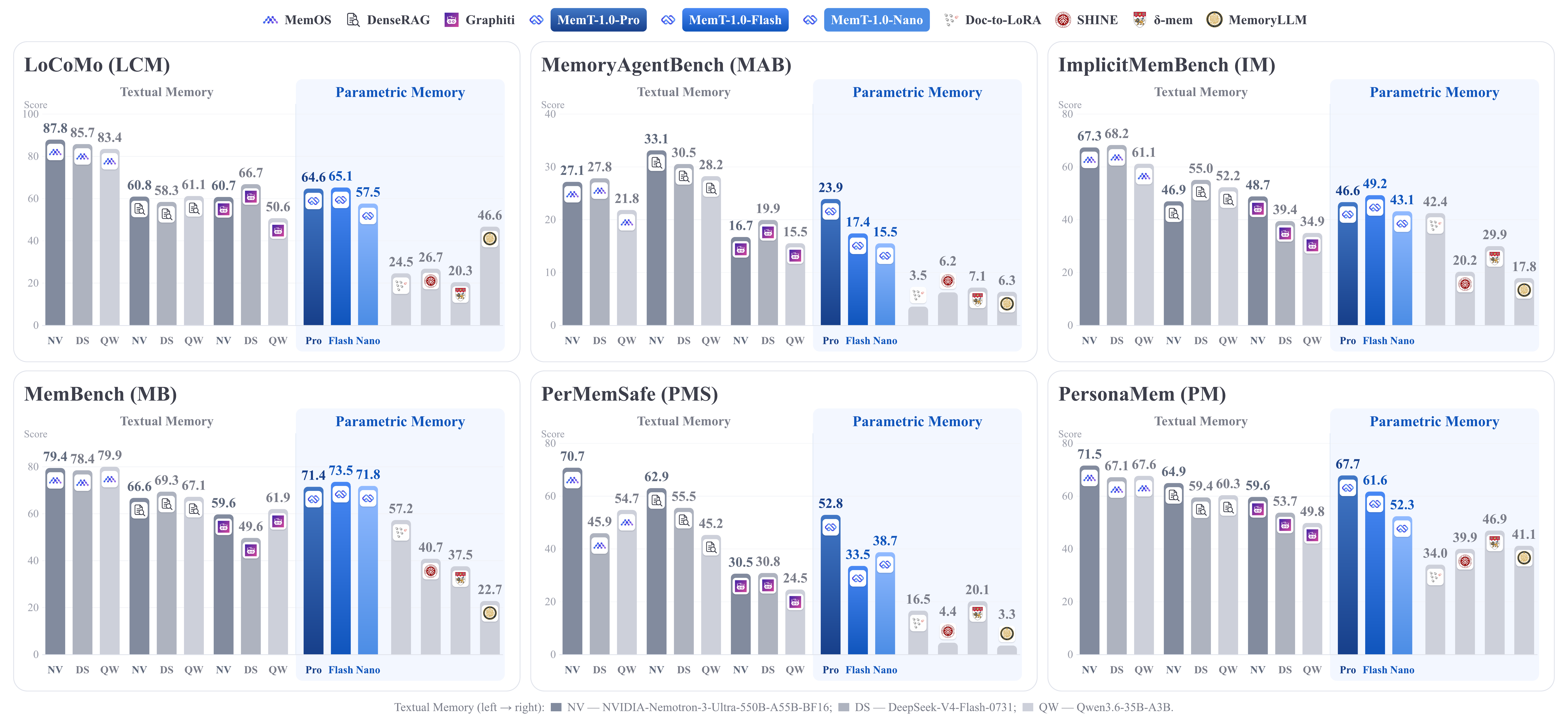}

    \vspace{-0.2cm}
    \captionof{figure}{
        Memory performance of MemT-1.0 on representative benchmarks.
    }
    \label{fig:overall_results}

    \vspace{-0.6cm}
\end{center}

\begin{center}
    \includegraphics[width=\textwidth]
        {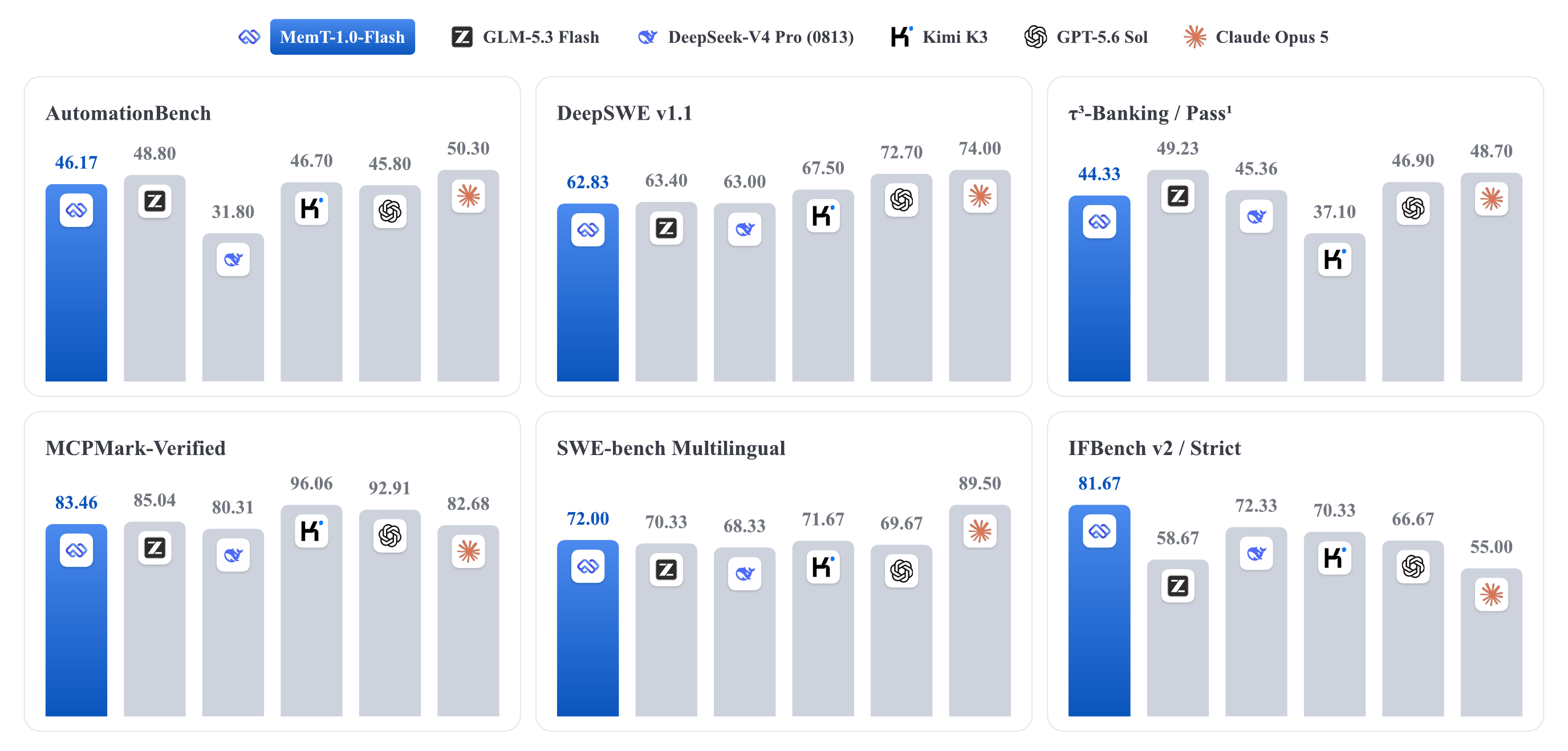}

    \captionof{figure}{
        Agentic performance of MemT-1.0-Flash after backbone preparation without native memory.
    }
    \label{fig:backbone_preparation_results}

\end{center}

\begin{figure*}[h]
    \centering
    \vspace{0.8em}

    \includegraphics[width=\textwidth]
        {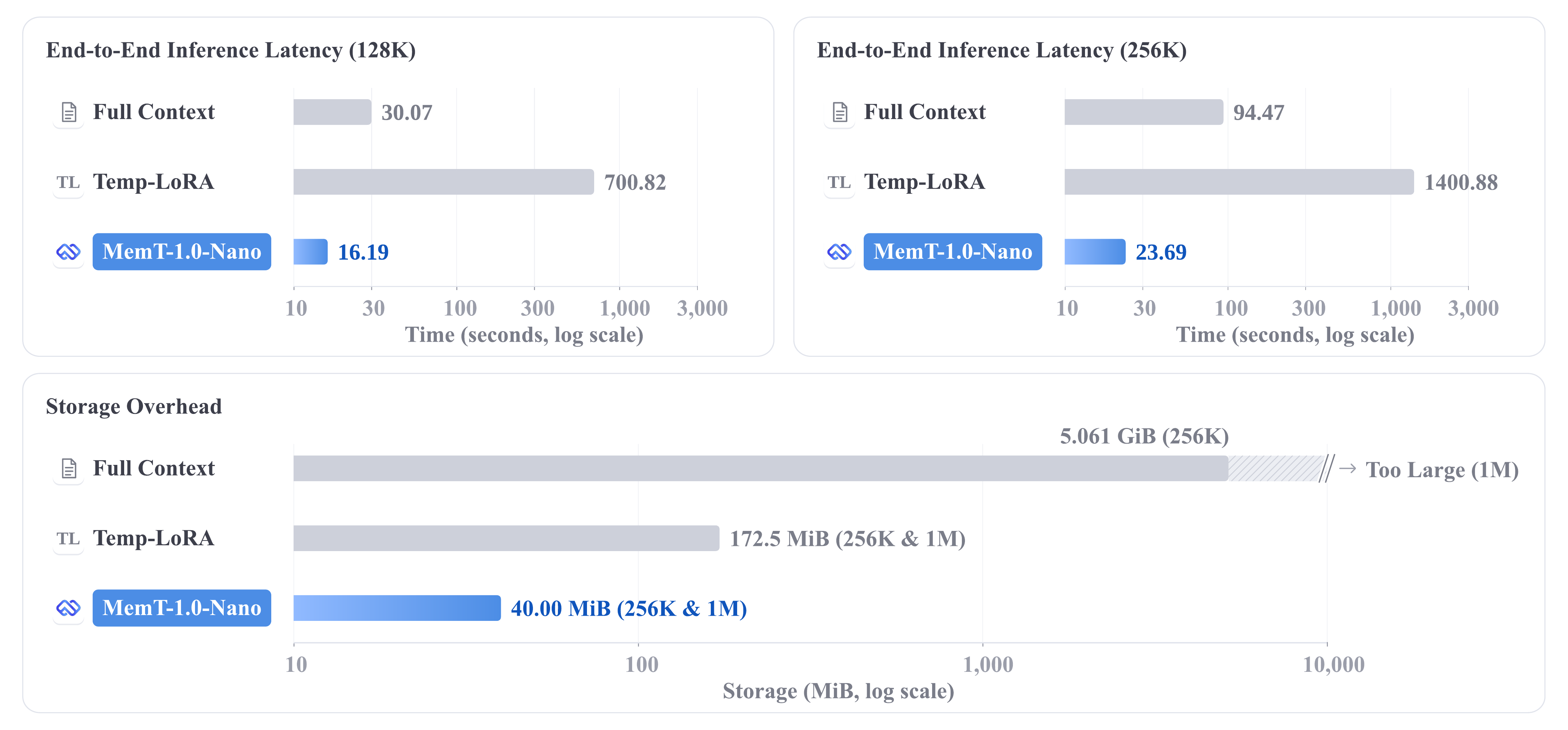}

    \caption{
    Inference efficiency and storage overhead of MemT-1.0-Nano.
    }
    \label{fig:overall_efficiency}

    \vspace{-0.8em}
\end{figure*}

\newpage
\tableofcontents

\newpage

\section{Introduction}

The rapid evolution of foundation models is expanding the boundary of what can be natively learned and performed within a single model.
Many capabilities are progressively internalized from external systems into the model itself, such as reasoning~\cite{guo2025deepseek,jaech2024openai} and multimodality~\cite{xu2026deepseek,team2024gemini}.
Memory represents another fundamental capability, enabling models to accumulate experience and adapt over extended interactions~\cite{zhang2025survey,hu2025memory}.
Our previous work, Metis~\cite{zhang2026metis}, first introduced memory foundation models to internalize memory as a native capability of foundation models.
This offers a path beyond external context and enables models to continuously accumulate information and experience, which marks a critical step towards memory-native evolution.

To improve memory capabilities, we introduce \textbf{MemT-1.0}, a new architecture with substantial upgrades over Metis, while retaining the structure of dynamic local memory and trainable hyper memory.
It replaces token selection with a stable token-level update mechanism based on Gated DeltaNet-2 (GDN-2)~\cite{hatamizadeh2026gated}.
To further scale native memory capability, we introduce Mixture-of-Storage (MoS) to expand the capacity of native memory.
We release a model family comprising \textbf{MemT-1.0-Nano} with 36B parameters (3B activated, 10M dynamic memory), \textbf{MemT-1.0-Flash} with 288B parameters (17B activated, 336M dynamic memory), and \textbf{MemT-1.0-Pro} with 554B parameters (59B activated, 13M dynamic memory).

Following a deliberate backbone preparation, we develop a two-stage training pipeline for MemT-1.0.
The first stage performs continued pre-training (CPT) to establish language modeling capabilities with native memory~\cite{gururangan2020don}.
The post-training stage further adapts the model to downstream memory tasks while preserving its general capabilities~\cite{ouyang2022training}.
To support this training pipeline, we construct a large-scale dataset of 34B tokens.
Across 11 memory benchmarks, our models outperform parametric memory baselines on average, showing effective utilization of historical information without accessing the original context during inference.
They also largely preserve general capabilities under the noise setting.
Our evaluation shows an increasing end-to-end inference latency and storage overhead advantage over full context on longer histories.

To support training and deployment, we develop a distributed infrastructure, which preserves memory isolation across independent trajectories.
We organize stateful inference around separate memory storage and utilization procedures, with shared model parameters and trajectory-specific memory states.
For a fixed memory configuration, the persistent state size remains constant as interactions accumulate.
In addition, we explore the scaling behavior of MemT-1.0.
Unlike general foundation models~\cite{kaplan2020scalinglawsneurallanguage}, memory foundation models introduce memory capacity as a critical scaling dimension, so we conduct scaling studies over memory capacity and training data.
These findings provide valuable guidance for building stronger memory foundation models.

We further apply MemT-1.0 to experience-guided coding agents, where retrieved procedural experience is internalized into native memory to guide subsequent actions.
On SWE-bench Verified~\cite{chowdhury2024swebenchverified} and SWE-bench Multilingual~\cite{Yang2025SWEsmithSD}, our method improves average task success and consistency across repeated runs over agents without experience, while achieving performance comparable to explicit textual experience.
These findings extend the role of native memory from information recall to procedural guidance in multi-step execution.

Our main contributions are summarized as follows: \\
$\bullet$ We introduce MemT-1.0, a memory foundation model architecture with improved memory storage and utilization mechanisms, together with MoS for scalable memory capacity.\\
$\bullet$ Following a deliberate backbone preparation, we design a two-stage training pipeline including CPT and post-training stages, with a corresponding infrastructure design.
We construct a training corpus of 34B tokens, and systematically evaluate MemT-1.0 on both memory and general benchmarks. The results show that MemT-1.0 achieves superior memory performance while largely preserving general capabilities. \\
$\bullet$ We explore the scaling behavior of memory capabilities in MemT-1.0.
We further deploy MemT-1.0 in practical agentic coding tasks to demonstrate its effectiveness and potential in enabling memory-native evolution.
We also discuss its current limitations and promising directions for future development.

\section{Model Architecture}
\label{sec:model_arc}

\begin{figure}[t]
    \centering
    \includegraphics[width=\linewidth]{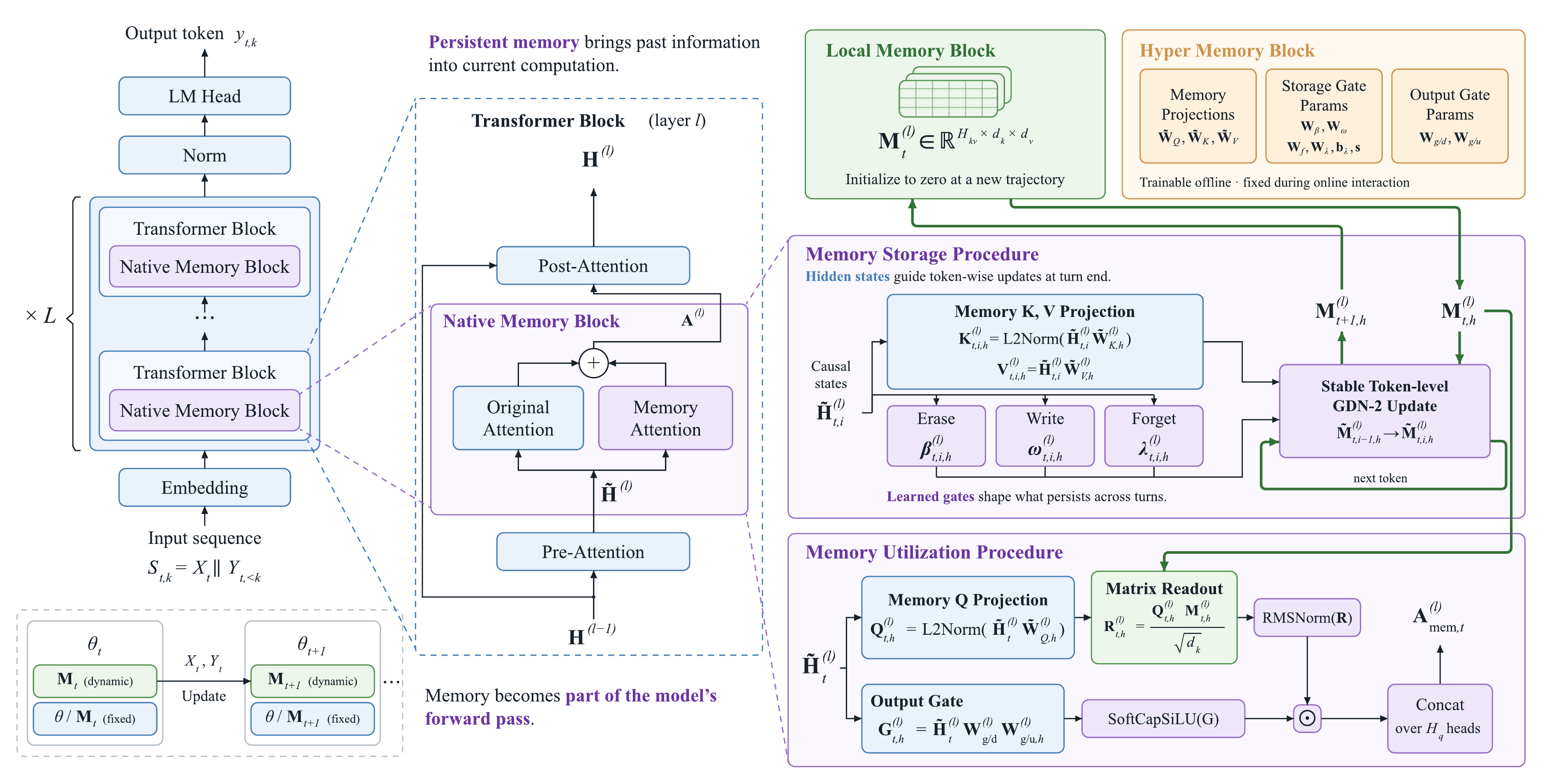}
    \vspace{-0.6cm}
    \caption{Architecture of MemT-1.0. MemT-1.0 equips causal language models with native memory that evolves across interaction steps. Each selected Transformer block is augmented with a native memory block that performs parallel original and memory attention. The memory mechanism comprises dynamic local memory and trainable hyper memory, together with native memory storage and utilization procedures for updating and retrieving persistent information.}
    \label{fig:memt_arc}
    \vspace{-0.4cm}
\end{figure}

\subsection{Overview}
MemT-1.0 is formulated under a multi-step scenario with discrete steps $t \in \{1,2, \cdots, T\}$, as presented in Figure~\ref{fig:memt_arc}.
At step $t$, let $\theta_t$ denote the current model parameters.
Given the input $X_t$, the autoregressive generation process can be formulated as
\begin{equation*}
    y_{t,k} \sim P \left( \cdot | X_t,Y_{t,<k} ; \theta_t \right),
\end{equation*}
where $Y_{t,<k}$ denotes the tokens generated before $y_{t,k}$.
Meanwhile, the stateful model parameter is updated by
\begin{equation*}
    \theta_{t+1} \sim P\left( \cdot | X_t,Y_t ; \theta_t \right),
\end{equation*}
where $Y_{t}$ is all generated tokens at step $t$.
MemT-1.0 is built upon general causal language models.
We denote the input sequence as $S_{t,k} = X_t||Y_{t,<k} = \left(x_{t,1}, \cdots, y_{t,1}, \cdots , y_{t,k-1}\right)$ with length $L_{t,k}$.
First, the input is mapped to the initial hidden state $\mathbf{H}^{(0)}_{t,k} \in \mathbb{R}^{L_{t,k} \times d}$ through the embedding layer, where $d$ is the dimension of the hidden state.
Then, it is processed sequentially by $L$ stacked Transformer blocks~\cite{vaswani2017attention} as follows:
\begin{equation*}
    \mathbf{H}^{(l)}_{t,k} = \operatorname{TransformerBlock}\left( \mathbf{H}^{(l-1)}_{t,k} \right), \quad \text{for $l=1, 2, \cdots L$}.
\end{equation*}
After that, the last hidden state $\mathbf{H}^{(L)}_{t,k}$ is normalized and projected into the vocabulary space through the language modeling head $\mathbf{W}_{\text{LM}} \in \mathbb{R}^{d \times |\mathcal{V}|}$, and a softmax function is then applied to obtain the probability distribution by
\begin{equation*}
    P \left( \cdot | X_t,Y_{t,<k} ; \theta_t \right) = \Softmax \left( \Norm\left( \mathbf{h}^{(L)}_{t,k} \right) \mathbf{W}_{\text{LM}} \right),
\end{equation*}
where $|\mathcal{V}|$ is the vocabulary size and $\mathbf{h}^{(L)}_{t,k}$ is the last position of $\mathbf{H}^{(L)}_{t,k}$.

MemT-1.0 augments each selected standard Transformer block with a native memory block, which is responsible for maintaining the memory from previous information.
For the $l$-th Transformer block, we denote the computation of attention output as 
\begin{equation*}
    \mathbf{A}^{(l)}_{\text{ori},t,k} = \operatorname{Attention}\left(\tilde{\mathbf{H}}^{(l)}_{t,k} \right), \quad \tilde{\mathbf{H}}^{(l)}_{t,k} = \operatorname{Pre-Attention}\left({\mathbf{H}}^{(l-1)}_{t,k} \right),
\end{equation*}
where $\operatorname{Attention}\left( \cdot \right)$ represents the backbone attention and $\tilde{\mathbf{H}}^{(l)}$ indicates the hidden state after pre-attention operations, such as pre-normalization.
MemT-1.0 introduces a parallel memory attention mechanism to utilize the parametric memory accumulated from previous steps for targeted layers, by
\begin{equation*}
    \mathbf{A}^{(l)}_{t,k} = \mathbf{A}^{(l)}_{\text{ori},t,k} + \mathbf{A}^{(l)}_{\text{mem},t,k}, \quad \mathbf{A}^{(l)}_{\text{mem},t,k} = \operatorname{MemoryAttention}\left( \tilde{\mathbf{H}}^{(l)}_{t,k}; \cdot \right),
\end{equation*}
where $\mathbf{A}^{(l)}_{\text{mem},t,k}$ indicates the output of memory attention.
After that, the combined attention output $\mathbf{A}^{(l)}_{t,k}$ is used in the subsequent computation as
\begin{equation*}
    \mathbf{H}^{(l)}_{t,k} = \operatorname{Post-Attention}\left( \mathbf{A}^{(l)}_{t,k}, {\mathbf{H}}^{(l-1)}_{t,k} \right).
\end{equation*}
For different backbones, we preserve the original $\operatorname{Pre-Attention}\left( \cdot \right)$, $\operatorname{Attention}\left( \cdot \right)$, and $\operatorname{Post-Attention}\left( \cdot \right)$ unchanged to retain the backbone capabilities, and introduce memory through $\operatorname{MemoryAttention}\left( \cdot \right)$ for different backbones.
Specifically, we use Qwen3.6-35B-A3B~\cite{qwen36_35b_a3b}, as released, as the backbone of MemT-1.0-Nano and a further post-trained DeepSeek-V4-Flash~\cite{deepseekai2026deepseekv4} (Section~\ref{sec:backbone_prep}) as the backbone of MemT-1.0-Flash.
In addition, we use NVIDIA-Nemotron-3-Ultra-550B-A55B-BF16~\cite{nvidia_nemotron_3_ultra_2026} as the backbone of MemT-1.0-Pro.

\subsection{Native Memory Block}

\subsubsection{Native Memory State}

Following our previous work Metis~\cite{zhang2026metis}, a native memory block in MemT-1.0 comprises a local memory block and a hyper memory block.
The former stores online information, while the latter is trained to equip general foundation models with memory capabilities in a data-driven manner.

\textbf{Local Memory Block.}
In MemT-1.0, the dynamic memory state at the $t$-th step and $l$-th layer can be represented as $\mathbf{M}_t^{(l)} \in \mathbb{R}^{H_{kv} \times d_k \times d_v}$, where $d_k, d_v$ represent the dimensions of memory keys and values, and $H_{kv}$ denotes the number of memory channels.
Compared with Metis, we remove the query-key normalization vector and extend the memory state to multiple channels. 
We initialize $\mathbf{M}_t^{(l)} = \mathbf{0}$ at the start of new trajectories.

\textbf{Hyper Memory Block.}
In the hyper memory block, the multi-channel projections of memory query, key, and value are denoted as $\tilde{\mathbf{W}}_Q^{(l)} \in \mathbb{R}^{H_{q} \times d \times d_k}$, $\tilde{\mathbf{W}}_K^{(l)} \in \mathbb{R}^{H_{kv} \times d \times d_k}$ and $\tilde{\mathbf{W}}_V^{(l)} \in \mathbb{R}^{H_{kv} \times d \times d_v}$ for each layer $l$.
Beyond these core memory projections, this block includes all other trainable parameters for memory processing.
First, the gating projections include the erase gate $\mathbf{W}_{\beta}^{(l)} \in \mathbb{R}^{d \times H_{kv} \times d_k}$ and the write gate $\mathbf{W}_{\omega}^{(l)} \in \mathbb{R}^{d \times H_{kv} \times d_v}$, along with the forget gate components consisting of $\mathbf{W}_{f}^{(l)} \in \mathbb{R}^{d \times d_{\text{gate}}}$, $\mathbf{W}_{\lambda}^{(l)} \in \mathbb{R}^{d_{\text{gate}} \times H_{kv} \times d_k}$, $\mathbf{b}_{\lambda}^{(l)} \in \mathbb{R}^{H_{kv} \times d_k}$, and a learnable vector $\mathbf{s}^{(l)} \in \mathbb{R}^{H_{kv}}$.
Finally, the memory output gating is parameterized by a shared down-projection $\mathbf{W}_{g/d}^{(l)} \in \mathbb{R}^{d \times d_{g/d}}$ and a multi-channel up-projection $\mathbf{W}_{g/u}^{(l)} \in \mathbb{R}^{H_q \times d_{g/d} \times d_v}$.
Generally, we have $H_q = H_{kv}$ to align query channel and key-value channels. 
These trainable parameters are denoted as $\phi$, which can be optimized during the training phase and remain fixed during online interaction.

In summary, the model parameters $\theta_t$ at step $t$ include the backbone parameters $\theta_b$, the static but trainable parameters $\phi$, and the dynamic memory states $\{\mathbf{M}_t^{(l)}\}_{l=1}^{L}$.
During online inference, these dynamic memory states are responsible for storing online information and evolve over interactions.

\subsubsection{Native Memory Procedures}
MemT-1.0 consists of two parallel native procedures, including memory storage and memory utilization.
The memory storage procedure updates the dynamic memory states to preserve information from the current interaction, while the memory utilization procedure retrieves information from previously accumulated memory states to support autoregressive generation.

\textbf{Native Memory Storage Procedure.}
MemT-1.0 replaces the token selection with a stable token-level update mechanism based on GDN-2~\cite{hatamizadeh2026gated} to update the dynamic memory state in each turn.
As the memory storage procedure is triggered at the turn level, we denote the hidden states as $\mathbf{H}^{(l)}_t$ of $S_{t} = X_t||Y_t$.
For the pre-attention input $\tilde{\mathbf{H}}^{(l)}_t$, we compute the memory key and value of the $h$-th channel by
\begin{equation*}
    \mathbf{K}_{t,h}^{(l)} = \operatorname{L2Norm}\left(\tilde{\mathbf{H}}^{(l)}_{t} \tilde{\mathbf{W}}_{K,h}^{(l)}\right), \quad \mathbf{V}_{t,h}^{(l)} = \tilde{\mathbf{H}}^{(l)}_{t} \tilde{\mathbf{W}}_{V,h}^{(l)},
\end{equation*}
where $\tilde{\mathbf{W}}_{K,h}^{(l)}$ and
$\tilde{\mathbf{W}}_{V,h}^{(l)}$ are the key and value projection matrices for the $h$-th memory channel.
Then, the erase gate and write gate can be obtained by
\begin{equation*}
    \bm{\beta}_{t}^{(l)} = \sigma\left( \tilde{\mathbf{H}}^{(l)}_{t} \mathbf{W}_{\beta}^{(l)} \right),
    \quad \bm{\omega}_{t}^{(l)} = \sigma\left( \tilde{\mathbf{H}}^{(l)}_{t} \mathbf{W}_{\omega}^{(l)} \right),
\end{equation*}
where $\mathbf{W}_{\beta}^{(l)}, \mathbf{W}_{\omega}^{(l)}$ are gating projection matrices.
In addition, the forget gate is calculated by
\begin{equation*}
    \bm{\lambda}_{t,h}^{(l)} = - \exp \left(\mathbf{s}^{(l)}_{h} \right) \odot \operatorname{Softplus}\left( \mathbf{F}_{t}^{(l)} \mathbf{W}_{\lambda,h}^{(l)} +\mathbf{b}_{\lambda,h}^{(l)} \right),
    \quad \mathbf{F}_{t}^{(l)} = \tilde{\mathbf{H}}^{(l)}_{t} \mathbf{W}_{f}^{(l)} ,
\end{equation*}
where $\mathbf{W}_{f}^{(l)}, \mathbf{W}_{\lambda}^{(l)}, \mathbf{b}_{\lambda}^{(l)}, \mathbf{s}_h^{(l)}$ are forget-gate parameters.
To prevent the divergence of memory states during updates, we introduce a stable mechanism for the token-level update, where it first calculates
\begin{equation*}
    \mathbf{U}_{t,i}^{(l)} = \sqrt{\bm{\beta}_{t,i}^{(l)}}
\odot
\mathbf{K}_{t,i}^{(l)}.
\end{equation*}
Then, the memory state can be updated by
\begin{equation*}
    \tilde{\mathbf{M}}_{t, i, h}^{(l)} = \left[
\mathbf{I}
-
\mathbf{U}_{t,i,h}^{(l)\top}
\mathbf{U}_{t,i,h}^{(l)}
\right]
\cdot
\mathbf{D}_{t,i,h}^{(l)}
\tilde{\mathbf{M}}_{t,i-1,h}^{(l)}
+
\mathbf{U}_{t,i,h}^{(l)\top}
\left(
\bm{\omega}_{t,i,h}^{(l)}
\odot
\mathbf{V}_{t,i,h}^{(l)}
\right),
\end{equation*}
where the initialization is given by $\tilde{\mathbf{M}}_{t, 0, h}^{(l)} = \mathbf{M}_{t,  h}^{(l)}$, and the state transition is defined as $\mathbf{M}_{t+1,  h}^{(l)} = \tilde{\mathbf{M}}_{t, L_t, h}^{(l)}$, with
\begin{equation*}
    \mathbf{D}_{t,i,h}^{(l)} = \operatorname{Diag}
\left[
\exp
\left(
\bm{\lambda}_{t,i,h}^{(l)}
\right)
\right].
\end{equation*}
The detailed analysis of our stable mechanism on GDN-2 is presented in Appendix~\ref{appendix:stable_gdn2}.

\textbf{Native Memory Utilization Procedure.}
In MemT-1.0, we remove the key-value normalization from Metis.
Because the memory utilization operates at the token level in parallel, we omit the decoding index $k$ for simplicity.
For the step $t$, we read out information from the current memory state $\mathbf{M}_{t,h}^{(l)}$ by
\begin{equation*}
    \mathbf{R}_{t,h}^{(l)} = \frac{\mathbf{Q}_{t,h}^{(l)} \mathbf{M}_{t,h}^{(l)}}{\sqrt{d_k}}, 
    \quad \mathbf{Q}_{t,h}^{(l)} = \operatorname{L2Norm}\left(\tilde{\mathbf{H}}^{(l)}_{t} \tilde{\mathbf{W}}_{Q,h}^{(l)}\right),
\end{equation*}
where $\tilde{\mathbf{W}}_{Q,h}^{(l)}$ is the memory query projection matrix.
After that, we calculate the memory output gate by 
\begin{equation*}
    \mathbf{G}_{t,h}^{(l)} = \tilde{\mathbf{H}}^{(l)}_{t} \mathbf{W}_{g/d}^{(l)} \mathbf{W}_{g/u,h}^{(l)},
\end{equation*}
where $\mathbf{W}_{g/d}^{(l)},\mathbf{W}_{g/u,h}^{(l)}$ are parameters of the readout gate.
Finally, the memory attention is calculated by
\begin{equation*}
    \begin{aligned}
        \mathbf{A}_{\text{mem},t}^{(l)} & = \operatorname{MemoryAttention}\left( \tilde{\mathbf{H}}^{(l)}_{t,k}; \mathbf{M}_{t}^{(l)} \right) \\
        & = \operatorname{Concat}_{h=1}^{H_{q}}\left[ \operatorname{RMSNorm}\left( \mathbf{R}_{t,h}^{(l)} \right) \odot \operatorname{SoftCapSiLU}\left( \mathbf{G}_{t,h}^{(l)}\right) \right],
    \end{aligned}
\end{equation*}
where
\begin{equation*}
    \operatorname{SoftCapSiLU}(x)
    =
    \tau \tanh\left(
        \frac{\operatorname{SiLU}(x)}{\tau}
    \right),
    \qquad
    \operatorname{SiLU}(x) = x\,\sigma(x),
\end{equation*}
and $\tau > 0$ controls the soft-cap threshold, with $\tau = 2$ by default.
Target layers, projection strategies, and other settings vary slightly to accommodate different backbones, and more details can be found in Appendix~\ref{appendix:backbone_adaptation}.

\subsection{Scaling Native Memory Capacity}

\begin{figure}[t]
    \centering
    \vspace{-0.4cm}
    \includegraphics[width=\linewidth]{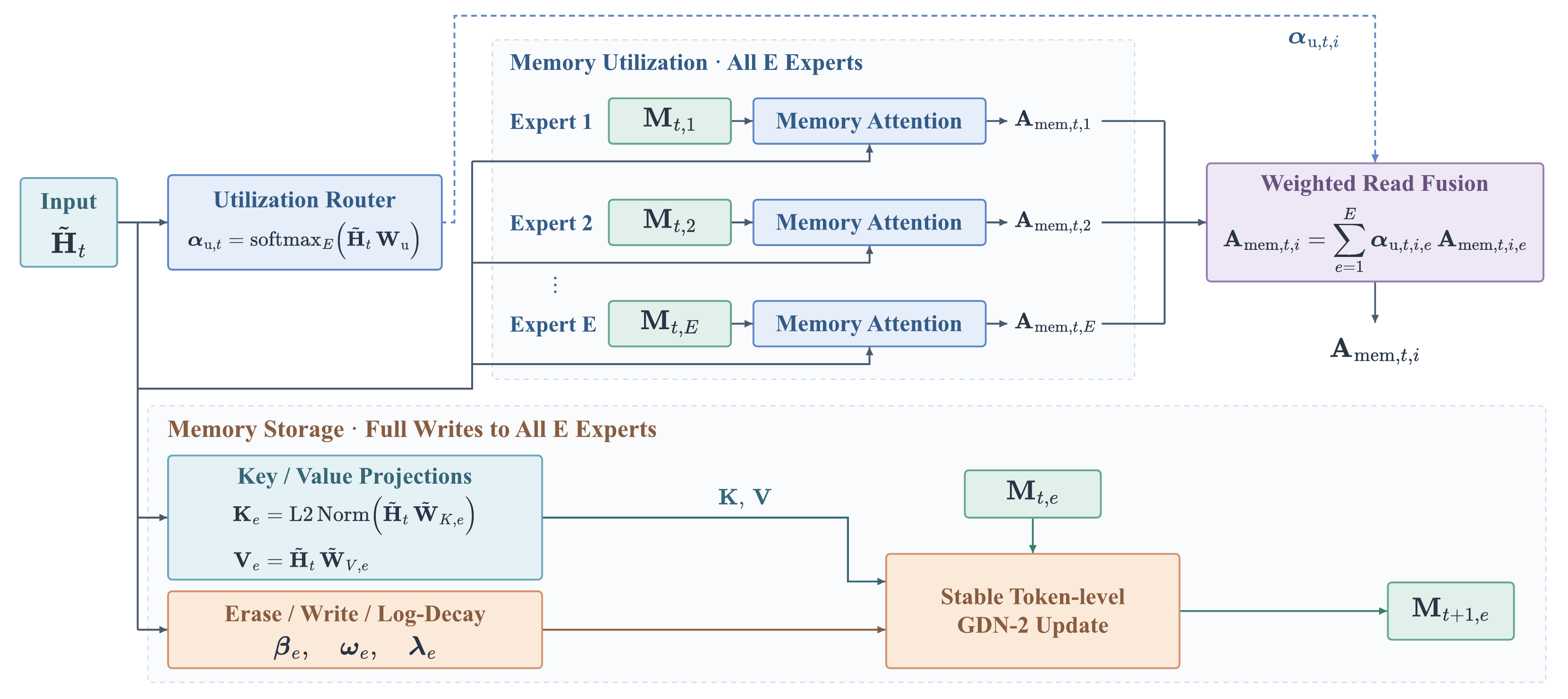}
    \caption{Scaling native memory capacity. MoS expands each native memory block into multiple memory experts, performs full writes to all experts, and uses a dense utilization router to combine their readouts.}
    \vspace{-0.4cm}
    \label{fig:mos_dse}
\end{figure}

To jointly scale the online memory capacity and the trainable memory parameters, we introduce Mixture-of-Storage (MoS), as presented in Figure~\ref{fig:mos_dse}, which expands a native memory block into $E$ parallel memory experts.
Unlike conventional sparse Mixture-of-Experts (MoE)~\cite{shazeer2017outrageouslylargeneuralnetworks}, MoS adopts dense routing over all memory experts, so every expert participates in the memory computation and the routing probabilities determine how information is read from different memory experts.

\textbf{Memory Expert Expansion.}
For the $l$-th MoS layer, we extend the local memory state with an additional expert dimension as
\begin{equation*}
    \bm{\mathcal{M}}_{t,\mathrm{MoS}}^{(l)}
    =
    \left\{
    \mathbf{M}_{t,e}^{(l)}
    \right\}_{e=1}^{E}
    \in
    \mathbb{R}^{E \times H_{kv} \times d_k \times d_v},
\end{equation*}
where $\mathbf{M}_{t,e}^{(l)} \in \mathbb{R}^{H_{kv} \times d_k \times d_v}$ is the dynamic memory state of the $e$-th expert, and each expert has independent hyper memory parameters.
All experts and utilization routers receive the same pre-attention hidden states~$\tilde{\mathbf{H}}_t^{(l)}$.

\textbf{Dense Storage and Weighted Utilization.}
MoS combines dense memory storage with weighted memory utilization.
During memory storage, all $E$ experts process the input tokens and update their respective memory states using expert-specific parameters.
During memory utilization, a dense router computes token-dependent weights to combine the experts' readouts.
Specifically, we introduce a trainable router projection
$\mathbf{W}_{\text{route/u}}^{(l)} \in \mathbb{R}^{d \times E}$,
which maps hidden representations to expert logits.
The routing weights are calculated as
\begin{equation*}
    \bm{\alpha}_{\text{u},t}^{(l)}
    =
    \Softmax
    \left(
    \tilde{\mathbf{H}}_{t}^{(l)}
    \mathbf{W}_{\text{route/u}}^{(l)}
    \right)
    \in \mathbb{R}^{L_t \times E},
\end{equation*}
where the softmax is applied along the expert dimension.
The memory readouts from all $E$ experts are then combined through a weighted sum using $\bm{\alpha}_{\text{u},t}^{(l)}$, allowing each token to adaptively utilize information stored across the experts.

\textbf{Mixture-of-Storage Fusion.}
For memory utilization, each expert independently follows the native memory utilization procedure described above.
We denote the memory-attention output of the $e$-th memory expert by
\begin{equation*}
    \mathbf{A}_{\mathrm{mem},t,e}^{(l)}
    =
    \operatorname{MemoryAttention}_{e}
    \left(
    \tilde{\mathbf{H}}_{t}^{(l)};
    \mathbf{M}_{t,e}^{(l)}
    \right).
\end{equation*}
The final memory attention at the $i$-th token in sequence $S_t$ is obtained by
\begin{equation*}
    \mathbf{A}_{\mathrm{mem},t,i}^{(l)}
    =
    \sum_{e=1}^{E}
    \bm{\alpha}_{\text{u},t,i,e}^{(l)}
    \mathbf{A}_{\mathrm{mem},t,i,e}^{(l)}.
\end{equation*}

For memory storage, all experts process the input tokens and update their respective memory states.
Each expert independently computes its memory key, value, erase gate, write gate, and log-decay using expert-specific parameters.

For the $e$-th expert, we define
\begin{equation*}
    \mathbf{U}_{t,e}^{(l)}
    =
    \sqrt{\bm{\beta}_{t,e}^{(l)}}
    \odot
    \mathbf{K}_{t,e}^{(l)}.
\end{equation*}

The memory state is then updated by

\begin{equation*}
    \tilde{\mathbf{M}}_{t,i,e,h}^{(l)}
    =
    \left[
    \mathbf{I}
    -
    \mathbf{U}_{t,i,e,h}^{(l)\top}
    \mathbf{U}_{t,i,e,h}^{(l)}
    \right]
    \mathbf{D}_{t,i,e,h}^{(l)}
    \tilde{\mathbf{M}}_{t,i-1,e,h}^{(l)}
    +
    \mathbf{U}_{t,i,e,h}^{(l)\top}
    \left(
    \bm{\omega}_{t,i,e,h}^{(l)}
    \odot
    \mathbf{V}_{t,i,e,h}^{(l)}
    \right),
\end{equation*}
where the recurrence is initialized with
$\tilde{\mathbf{M}}_{t,0,e,h}^{(l)} = \mathbf{M}_{t,e,h}^{(l)}$,
and the updated memory state is
$\mathbf{M}_{t+1,e,h}^{(l)} = \tilde{\mathbf{M}}_{t,L_t,e,h}^{(l)}$.
The decay matrix is defined as
\begin{equation*}
    \mathbf{D}_{t,i,e,h}^{(l)}
    =
    \operatorname{Diag}
    \left[
    \exp
    \left(
    \bm{\lambda}_{t,i,e,h}^{(l)}
    \right)
    \right].
\end{equation*}

\section{Training}
\label{sec:training}
\subsection{Backbone Preparation}
\label{sec:backbone_prep}
A memory foundation model inherits from its backbone both the knowledge fixed at pre-training and the reliability with which that knowledge becomes behavior, and post-training treats the two very differently. Training on the backbone's own verified samples fits its distribution conditioned on verifier acceptance~\cite{zelikman2022star,singh2024beyond}: it reweights what the backbone already produces and, by itself, leaves tasks the backbone almost never solves untouched. Our per-item analysis bears this out. Gains concentrate on tasks the backbone solves occasionally. Tasks it never solved in eight attempts end at only 1.6--4.0\% accuracy across four benchmarks after the preparation described below, in line with observations for RL~\cite{yue2025does}, whereas on a separate probe set of hard knowledge questions, supplying the relevant facts in context raises accuracy from 27\% to 91\%, which suggests that the remaining gap is largely informational. In our setting, reliability is therefore what post-training mainly improves, and new information is what native memory must supply.

Accordingly, we post-train DeepSeek-V4-Flash for MemT-1.0-Flash before attaching native memory, with data from a three-step pipeline applied within each of three tracks: general capabilities (instruction following, knowledge, mathematics, and long-context reasoning), agentic coding, and agentic working (tool use and workflow automation). First, capability profiling: each task is attempted repeatedly and classified as reliably, occasionally, or not solved, a profile that stands in for the pre-training data we cannot see. Second, attribution by intervention: a failure is re-run with the suspected missing piece, a grader convention, a procedural hint, or a fact, placed in context, and whether it disappears tells a missing convention or procedure apart from missing knowledge; failures that persist under every intervention are excluded from synthesis. Third, capability-conditioned synthesis assigns each class its own data source. Reliably solved tasks supply replay. Occasionally solved tasks supply the backbone's own attempts that pass programmatic verifiers (constraint checkers, unit tests, or sandboxed agent runs), which keeps most of the data on the backbone's own distribution. Failures traced to a missing convention or procedure are learned either from guided sampling, where the guidance appears in the sampling context but not in the target, or from style-matched demonstrations by stronger models. Knowledge gaps, which synthesis has so far closed only item by item, are left mainly to retrieval and, in MemT-1.0, to native memory. Transfer is checked on held-out task families.

The recipe was settled over five rounds of full-mixture training and more than fifty training runs in total. The final mixture holds 126.5K trajectories (2.4B tokens, 1.1B loss-bearing), 56.4\% general, 18.0\% agentic coding, and 25.3\% agentic working by count; agentic trajectories account for about 90\% of loss-bearing tokens, so general data anchors the backbone's existing capabilities while the long agentic trajectories carry the multi-turn behavior that memory tasks build on. The prepared backbone improves on all six benchmarks we monitor across the three tracks, by 6.3 points on average over the public release (Figure~\ref{fig:backbone_preparation_results}), and on SWE-bench Multilingual and IFBench~v2, at 284B total and 13B active parameters, it exceeds the reported scores of larger open models, including GLM-5.3 Flash (320B total, 18B active)~\cite{glm53flash} and the far larger DeepSeek-V4 Pro (1.6T total, 49B active)~\cite{deepseekai2026deepseekv4}.

The backbone is then frozen to serve as both the host model and the full-context distillation teacher of Section~\ref{sec:objectives}. Backbone preparation and native memory are two routes by which experience shapes a model's later behavior, through the weights before deployment and through memory states during interaction; closing the loop between them is left to future work.

\subsection{Data Construction}

To support the training of MemT-1.0, we construct a large-scale training corpus with 34B tokens, which is composed of continued pre-training (CPT) and post-training data.
We collect 20B tokens of CPT data, which mainly consists of large-scale general-purpose language modeling corpora.
It contains long-form and sequential text to expose the model to diverse linguistic patterns and extensive contextual dependencies.

For post-training, we organize 14B tokens of data into three major categories.
First, we construct memory-native language modeling data.
This data encourages MemT-1.0 to maintain linguistic coherence, stylistic consistency, and instruction-following capability when information is accessed through native memory.
Second, we construct data for native memory storage and utilization, including tasks such as summarization, continuation, and compression.
They provide explicit supervision for storing information that should be retained in native memory.
They train the model to recover and utilize the stored information during subsequent generation.
Third, we introduce memory task data to further strengthen the model's long-range memory capability.
They further improve the model's ability to selectively retain critical information and resolve long-range dependencies, which improves the utilization of memory when answering downstream queries.
Part of the training corpus is based on real-world interaction patterns in MemOS's anonymized internal data.

All three categories are formulated in a unified context-query-answer format.
The context is first processed and stored through the native memory mechanism, after which the model receives the corresponding query and is trained to generate the target answer conditioned on the resulting memory state.
This unified formulation allows heterogeneous memory-related tasks to be incorporated into the same post-training pipeline.

\subsection{Training Objectives}
\label{sec:objectives}
The core training objective of MemT-1.0 is to align its capability with the backbone given full context, which enables the model to retain and utilize historical information without preserving the context.
To this end, we introduce an off-policy distillation framework for training native memory models, where the backbone serves as the teacher model for acquiring memory capabilities~\cite{hinton2015distillingknowledgeneuralnetwork}.

Let $\theta_{b}$ denote the frozen backbone parameters and $\phi$ denote the trainable memory parameters, including the hyper memory blocks and extra parameters of MoS.
For a training trajectory $\{(X_t,Y_t)\}_{t=1}^{T}$, we define the preceding textual history as $C_{<t}=S_1\Vert\cdots\Vert S_{t-1}$, where $S_t=X_t\Vert Y_t$.
We use the original backbone without the additional memory modules as the teacher, whose next-token distribution is
\begin{equation*}
    q_{t,k}(v)
    = P\left(v \mid C_{<t},X_t,Y_{t,<k};\theta_{b}\right),
    \qquad v\in\mathcal{V}.
\end{equation*}
The MemT-1.0 student receives the current input and the same observed response prefix, with previous steps represented by the incoming local memory state $\mathbf{M}_t$, where
\begin{equation*}
    p_{t,k}(v)
    = P\left(v \mid X_t,Y_{t,<k};\theta_{b},\phi,\bm{\mathcal{M}}_t\right).
\end{equation*}
Here, $\bm{\mathcal{M}}_t$ denotes the collection of dynamic states
across all memory layers and experts.
The state is initialized to zero for each independent trajectory
and evolves through the native storage procedure after each completed step.
Both models are evaluated under teacher forcing on
fixed teacher-generated trajectories rather than student-generated
rollouts, making the distillation objective off-policy
with respect to the student policy.
To reduce the size of the teacher targets, we retain its Top-$K$ tokens at each supervised position.
Let $\mathcal{K}_{t,k}\subseteq\mathcal{V}$ be this token set and $T_{d}>0$ be the distillation temperature.
Given $q^{T_{d}}$ and $p^{T_{d}}$ for distributions obtained by dividing the respective logits by $T_{d}$ before softmax, we define
\begin{equation*}
    \hat{q}_{t,k}(v)
    = \frac{q^{T_{d}}_{t,k}(v)}
    {\sum_{u\in\mathcal{K}_{t,k}}q^{T_{d}}_{t,k}(u)},
    \qquad v\in\mathcal{K}_{t,k},
\end{equation*}
and set $\hat{q}_{t,k}(v)=0$ outside this set.
The per-token distillation loss is the forward KL divergence from this truncated teacher distribution to the student by
\begin{equation*}
    \ell^{\mathrm{KL}}_{t,k}
    = T_{d}^{2}
    \sum_{v\in\mathcal{K}_{t,k}}
    \hat{q}_{t,k}(v)
    \log\frac{\hat{q}_{t,k}(v)}{p^{T_{d}}_{t,k}(v)}.
\end{equation*}
The teacher is renormalized over its selected support, whereas the student remains normalized over the full vocabulary.
Consequently, the loss also penalizes student probability mass assigned outside the teacher's selected tokens.
We additionally use the observed target token $y_{t,k}$ for next-token cross-entropy supervision
\begin{equation*}
    \ell^{\mathrm{CE}}_{t,k}=-\log p_{t,k}(y_{t,k}).
\end{equation*}
For a training batch $\mathcal{B}$, let $m^{(b)}_{t,k}\in\{0,1\}$ indicate whether a position in trajectory $b$ is supervised.
The training loss is
\begin{equation*}
    \mathcal{L}(\phi)
    = \frac{1}{N_{\mathrm{spt}}}
    \sum_{b\in\mathcal{B}}\sum_{t,k}m^{(b)}_{t,k}
    \left[
        \lambda_{\mathrm{KD}}\ell^{\mathrm{KL},(b)}_{t,k}
        +\lambda_{\mathrm{CE}}\ell^{\mathrm{CE},(b)}_{t,k}
    \right],
\end{equation*}
where $N_{\mathrm{spt}}=\sum_{b\in\mathcal{B}}\sum_{t,k}m^{(b)}_{t,k}$ is the number of supervised tokens, and $\lambda_{\mathrm{KD}}$ and $\lambda_{\mathrm{CE}}$ are nonnegative weights with at least one nonzero value.
Cross-entropy uses the untempered student distribution.
The mask selects the prediction targets appropriate to each training stage and excludes padding.
History tokens may participate in memory construction without receiving direct loss.
Normalization is performed over all supervised tokens in the optimization batch, including those distributed across workers and accumulation steps.
Specifically, we use CE supervision during CPT and KL-based distillation during post-training.
For MoS, we do not add a MoE-style load-balancing auxiliary loss.
In ablations where expert routing weights modulate the memory-update gates, we observe a self-balancing trend in expert loads consistent with a spontaneous negative-feedback effect.
With full-write storage, all experts process the same set of tokens, balancing the write workload by construction.
These observations motivate training MoS without an auxiliary load-balancing objective.
For MemT-1.0-Flash, the backbone prepared in Section~\ref{sec:backbone_prep} is frozen during native-memory training.

\subsection{Continued Pre-Training}

We first conduct CPT to establish the basic capability of language modeling with native memory.
This stage trains MemT-1.0 to predict subsequent tokens using both the current textual context and information accumulated from preceding segments.
For each training sequence, we divide the tokens into consecutive, non-overlapping segments and treat each segment as a step $t$.
The student performs causal language modeling within the current segment, with the preceding history $C_{<t}$ represented by the incoming memory states $\bm{\mathcal{M}}_t$.
Its backbone context is restricted to the current segment, so information from earlier segments is accessed through native memory.
After processing a segment, the native storage procedure applies token-level updates to obtain $\bm{\mathcal{M}}_{t+1}$ for the next segment.
The final segment is retained even when it is shorter than the specified segment length.
Memory states are carried across segments within the same sequence and initialized to zero at the beginning of each independent sequence.

We use CE loss during CPT, with the ground-truth next tokens serving as the supervision targets. The student performs causal language modeling within the current segment while relying on native memory to access information from preceding segments. This encourages the model to establish language modeling capabilities under native memory without requiring the full historical context as textual input.

\subsection{Post-Training}

Post-training further adapts MemT-1.0 to native memory-based response generation.
We initialize the memory parameters from the CPT checkpoint and train on instances containing historical context, queries, and reference responses.
The historical context is processed chronologically through the native storage procedure, producing the memory states used to answer subsequent queries. The responses are rollouts from the full-context teacher, and serve to condition both models under teacher forcing rather than as direct supervision targets.
At step $t$, the student receives the current query $X_t$ and the observed response prefix $Y_{t,<k}$, with the preceding history $C_{<t}$ represented by $\bm{\mathcal{M}}_t$.
The teacher retains access to the corresponding textual history.
Memory states are initialized independently for each trajectory, preserving the association between a query and its own history.

We use the KL loss in the post-training stage.
The supervision mask $m^{(b)}_{t,k}$ selects response tokens, while history, query, and padding positions receive no direct token loss.
Both teacher and student are evaluated under teacher forcing with the same observed response prefix.
Reference responses thus determine the prefixes on which the models are evaluated, while the teacher's probability distributions provide the distillation targets.
The loss is averaged over all supervised response tokens in the training batch.
This focuses optimization on the outputs required by the task while training the student to approximate the full-context teacher through native memory.
Although historical tokens receive no direct loss, gradients from response predictions propagate through memory readout and earlier storage operations.
This allows the model to learn which historical information to retain and how to utilize it for the current query.
The full-context teacher provides a reference for both the use of historical information and the backbone's response behavior, encouraging MemT-1.0 to acquire task-specific memory capabilities while preserving its general capabilities.

\subsection{Training Setups}
For MemT-1.0-Nano, training consists of 1 CPT epoch followed by post-training.
The global batch size is 64 packed sequences.
Each pack contains at most 64K tokens, while the segment length is 2K tokens.
Incomplete final global batches are dropped.
For the distillation objective, we retain the teacher's top-$K$ tokens at each supervised position with $K=64$, and set the distillation temperature to $T_{d}=1.0$.
We optimize only the memory parameters using AdamW with $\beta_1=0.9$, $\beta_2=0.999$, $\epsilon=10^{-8}$, weight decay $0.01$, and a gradient clipping norm of $1.0$.
We use mixed-precision training.
The backbone runs in BF16, while numerically sensitive components (e.g., RMSNorm, Softmax, and memory states) and optimizer states are kept in FP32.
A soft cap is applied to the memory output gate, with a limit of 2.0, for training stability.
We use a WSD learning rate scheduler for MemT-1.0-Nano training. During CPT, we warm up for 600 steps to a learning rate of $1\times10^{-4}$, and keep the peak rate constant for the remaining steps. During post-training, we warm up for 500 steps to $1\times10^{-4}$, keep it constant, and apply cosine decay over the final 1,000 steps to a minimum learning rate of $5\times10^{-6}$.
For MemT-1.0-Flash, we initialize from the prepared backbone and use a global batch of 192 packs, with the same 64K maximum pack length and 2K-token storage steps.
Memory training uses the online backbone teacher with Top-64 targets, temperature 1, and KL loss.
The learning-rate schedule has a peak of $8\times10^{-5}$, a 100-step warmup, and a planned cosine decay over the final steps.

For MemT-1.0-Pro, the CPT recipe freezes the Nemotron backbone and optimizes only the memory parameters using CE loss.
The global batch contains 128 packs of at most 65,536 tokens each, with 2,048-token storage steps and independent memory states for different sequences. AdamW uses $\beta_1=0.9$, $\beta_2=0.999$, $\epsilon=10^{-8}$, weight decay $0.01$, and a gradient clipping norm of $1.0$.
The WSD schedule is configured for 10,729 optimizer steps, with a 100-step linear warmup to $8\times10^{-5}$, 9,556 steps at the peak rate, and a 1,073-step cosine decay to $8\times10^{-6}$.
For MemT-1.0-Pro post-training, we initialize the memory parameters from the CPT checkpoint and keep the Nemotron backbone frozen.
We retain the same batch configuration, storage step length, and AdamW hyperparameters as in CPT.
We use the online full-context backbone teacher with Top-64 targets, temperature $T_d=1.0$, and KL loss.
Post-training runs for two epochs with a WSD schedule, including a 100-step linear warmup to $8\times10^{-5}$, followed by a constant learning rate and cosine decay over the final 20\% of steps to $8\times10^{-6}$.

\section{System and Infrastructure}
\label{sec:infra}

Each interaction in MemT-1.0 depends on memory accumulated from earlier steps.
Training must combine these dependencies with batched GPU computation, while inference must maintain an independent evolving state for each trajectory.
Our infrastructure executes the model one decoder layer at a time, handles state recurrence within memory blocks, and manages model weights separately from trajectory-specific states.
We describe the resulting training schedule, memory operators, and online state organization below.

\subsection{Distributed Training}
\label{sec:distributed_training}

Training packs combine independent trajectories, each containing the steps defined in Section~\ref{sec:training}.
Attention stays within each step, while memory carries information between steps of the same trajectory and resets at trajectory boundaries.
Teacher forcing supplies all tokens in the pack, so the model processes the complete pack one decoder layer at a time.
Within each layer, attention and memory projections are batched across tokens.
The memory block alone advances state along each trajectory.
This organization retains large backbone matrix multiplications while preserving the order of memory writes and reads.

We use Megatron-Core~\cite{narayanan2021efficientlargescalelanguagemodel} and Bridge~\cite{nvidia2026megatronbridge} for distributed training, including expert parallelism for the backbone MoE layers.
Although backbone weights are frozen during memory training, gradients still pass through backbone computations and memory transitions, allowing later predictions to train earlier writes.
The remaining infrastructure challenge is therefore the cost of memory computation and its training activations, which we address below.

\subsection{Efficient Operators}
\label{sec:infra_operators}

Native memory introduces two related costs: ordered state updates and activations that grow with the number of memory experts.
Our execution separates state updates from reads of step snapshots, then groups complete steps into windows to limit the lifetime of expert projections.
This lets the memory branch control its working set while the backbone continues to process the full token pack layer by layer.

\textbf{Chunked updates and snapshot reads.}
The stable update has a rank-one structure.
Suppressing layer, step, and channel indices, let $\mathbf{k}_i$ denote the normalized key, and define $\mathbf{u}_i=\sqrt{\boldsymbol{\beta}_i}\odot\mathbf{k}_i$, $\mathbf{z}_i=\boldsymbol{\omega}_i\odot\mathbf{v}_i$, and $\mathbf{D}_i=\operatorname{Diag}(\exp(\boldsymbol{\lambda}_i))$.
The update in Section~\ref{sec:model_arc} can then be written as
\begin{equation*}
\mathbf{S}_i=\mathbf{D}_i\mathbf{S}_{i-1}
+\mathbf{u}_i^\top\bigl(\mathbf{z}_i-\mathbf{u}_i\mathbf{D}_i\mathbf{S}_{i-1}\bigr),
\qquad
\mathbf{S}_0=\mathbf{M}_t,\quad \mathbf{M}_{t+1}=\mathbf{S}_{L_t}.
\end{equation*}
Each token requires a vector--matrix multiplication and an outer product, with leading work of $O(d_kd_v)$ per channel.
We use chunked WY~\cite{NEURIPS2024_d13a3eae} to group these small dependent updates into triangular solves and matrix multiplications, carrying state between chunks and reconstructing intermediate quantities during backward.
The adaptation to MemT-1.0 is to pair this writer with a separate snapshot reader: all queries in step $t$ read the same incoming state $\mathbf{M}_t$, even as the writer processes several chunks to form $\mathbf{M}_{t+1}$.
The reads can therefore be batched independently, followed by RMSNorm and SoftCapSiLU output gating.
Gradients from later reads propagate through the writes that produced their snapshots.
Writer chunks determine computation granularity, and model steps determine when updated memory becomes visible.

\textbf{Complete-step windows for MoS.}
Dense MoS evaluates every expert, so materializing all expert projections at once requires tensors that grow with both pack length and expert count.
Whole-pack execution does exactly this: it projects every token in the pack before processing the resulting memory computations.
Windowed execution instead performs expert projections, state updates, and readout fusion for one group of complete steps at a time, passing a differentiable FP32 final state to the next window.
Both schedules use the same expert parameters, utilization router, and update kernel.
Each expert updates its memory using its own keys, values, erase gates, write gates, and log-decay.
Utilization probabilities mix the gated expert readouts.

A window budget limits how many tokens are projected together, while keeping each step intact even if it exceeds the budget.
Recomputing internal intermediates during backward~\cite{korthikanti2023reducing} avoids retaining all windows' expert projections for the full training pass.
Smaller windows reduce live intermediates but require more launches and recomputation.
Outputs, routing probabilities, and boundary states still occupy memory, so the total remains dependent on pack length as well as window size.

\textbf{The memory--time tradeoff of window size.} We measure one complete MoS memory block on a single GPU using synthetic hidden-state inputs.
Table~\ref{tab:infra_window_tradeoff} fixes a 32K-token pack, BF16 parameters and projections, and 2048-token steps, comparing peak allocated memory and forward-plus-backward (F+B) time for two window sizes.
Arrows denote whole-pack execution $\rightarrow$ windowed execution; all values are medians.

\begin{table}[htbp]
    \centering
    \caption{Memory-time tradeoff for one MoS memory block on a single GPU at a 32K-token pack length.
    Arrows denote whole-pack to windowed execution. The values are medians.}
    \label{tab:infra_window_tradeoff}
    \small
    \setlength{\tabcolsep}{5pt}
    \renewcommand{\arraystretch}{1.15}
    \begin{tabular}{lrrrr}
        \toprule
        \makecell[l]{Window size\\(complete steps)} & \makecell{Peak memory\\(GiB)} & \makecell{Memory\\reduction} & \makecell{F+B\\(ms)} & \makecell{Time\\increase} \\
        \midrule
        2048 tokens (1 step) & $47.83 \rightarrow 6.03$ & 87.4\% & $462.7 \rightarrow 493.5$ & 6.6\% \\
        4096 tokens (2 steps) & $47.83 \rightarrow 9.49$ & 80.1\% & $463.0 \rightarrow 470.6$ & 1.6\% \\
        \bottomrule
    \end{tabular}
\end{table}

In this configuration, processing two steps at a time reduces peak memory by approximately 80\%, with a 1.6\% increase in median F+B time.
Shrinking the window to one step saves additional memory but raises the time increase to 6.6\%.
The window budget thus provides a direct resource tradeoff: smaller windows reduce live intermediates, while larger windows reduce execution overhead.
Appendix~\ref{appendix:infra_experiment_details} provides the block configuration, measurement protocol, and full results and variability from 8K to 32K.

\subsection{Stateful Inference and Serving}
\label{sec:infra_serving}

During online interaction, backbone and hyper memory weights remain fixed and are shared across trajectories, while each trajectory maintains its own dynamic memory states.
Within a step, generation reads the incoming memory snapshot, and attention over the current text uses that step's KV cache.
Once the interaction completes, storage incorporates the input and response into memory for the next interaction.
The serving runtime manages states by trajectory so that subsequent requests continue their respective histories.

For a fixed configuration, the number of memory-state elements per trajectory is
\begin{equation*}
N_{\mathrm{state}}=\sum_l E_l H_{kv,l}d_{k,l}d_{v,l},
\end{equation*}
where the sum covers memory layers and $E_l=1$ without MoS.
Growing history updates these elements while their number remains determined by configured memory capacity.
Total storage for multiple trajectories grows with trajectory count.
Active requests additionally require their current-step KV caches and computational workspaces.
Persistent-state storage can thus be estimated directly from the memory configuration, whereas runtime GPU memory also includes shared weights and active requests.
The effect of finite capacity on retained historical detail is examined in the capability evaluations and Discussion.

This organization separates history preparation from subsequent queries.
In Evaluation's two-GPU, single-request BEAM experiment, MemT-1.0-Nano's query latency after history preparation remains nearly constant as history grows.
At 64K tokens of history, MemT-1.0-Nano achieves an approximately \textbf{33$\times$ reduction} in logical per-trajectory state size compared with full context.
The storage metric measures the logical state size after history preparation and before querying, and excludes shared weights.
When subsequent queries reuse that state, the initial preparation cost is amortized over multiple interactions.
The full protocol and curves appear in Section~\ref{sec:evaluation} and Appendix~\ref{appendix:eval_storage_overhead}.

\section{Memory Scaling Behavior}

\subsection{Overview}

Scaling has been a key driver behind the continuous improvement of foundation models, where increasing model capacity, training data, and computation often leads to predictable gains in general capabilities~\cite{kaplan2020scalinglawsneurallanguage}.
However, for memory foundation models, it remains unclear whether native memory capabilities exhibit similar scaling behavior.
Exploring such scaling behavior is critical, as it reveals whether memory capability can continue to improve with scale and provides guidance for the design of larger memory foundation models.
In particular, memory-related training and the capacity of native memory states are two important factors that may influence memory capability, motivating us to explore scaling behavior along these two aspects.

In this work, we study memory scaling along both dimensions.
Data and training scaling explores how memory capability changes with continued exposure to memory-related post-training data, while memory capacity scaling studies the impact of increasing the capacity of native memory states.
They allow us to characterize how native memory capability evolves with additional training and expanded memory capacity.

\subsection{Scaling Protocol}

We conduct all scaling experiments based on MemT-1.0-Nano under a controlled experimental protocol.
We explore two scaling aspects, including data and training scaling and memory capacity scaling.
The backbone architecture, training objectives, and evaluation protocol are kept consistent across different configurations unless otherwise specified.
We evaluate memory capability using the same memory benchmarks to characterize performance changes across different scales.

Specifically, for data and training scaling, we evaluate checkpoints along a continuous post-training trajectory spanning two consecutive epochs over the same corpus.
The model architecture and training corpus remain fixed, while cumulative training exposure and optimization steps increase throughout training.
For memory capacity scaling, we vary the number of memory experts in MoS to expand native memory capacity, while keeping the remaining architectural components and training configurations consistent.
By examining the resulting performance curves, we investigate how native memory capability evolves with additional post-training and increased memory capacity.

\subsection{Data and Training Scaling}

We explore whether additional exposure to memory-related post-training data improves memory capability at a fixed model capacity.
Starting from the CPT checkpoint, we follow a single MemT-1.0-Nano training trajectory for two epochs over the same post-training corpus.
\begin{figure}[t]
    \centering
    \includegraphics[width=0.8\linewidth]{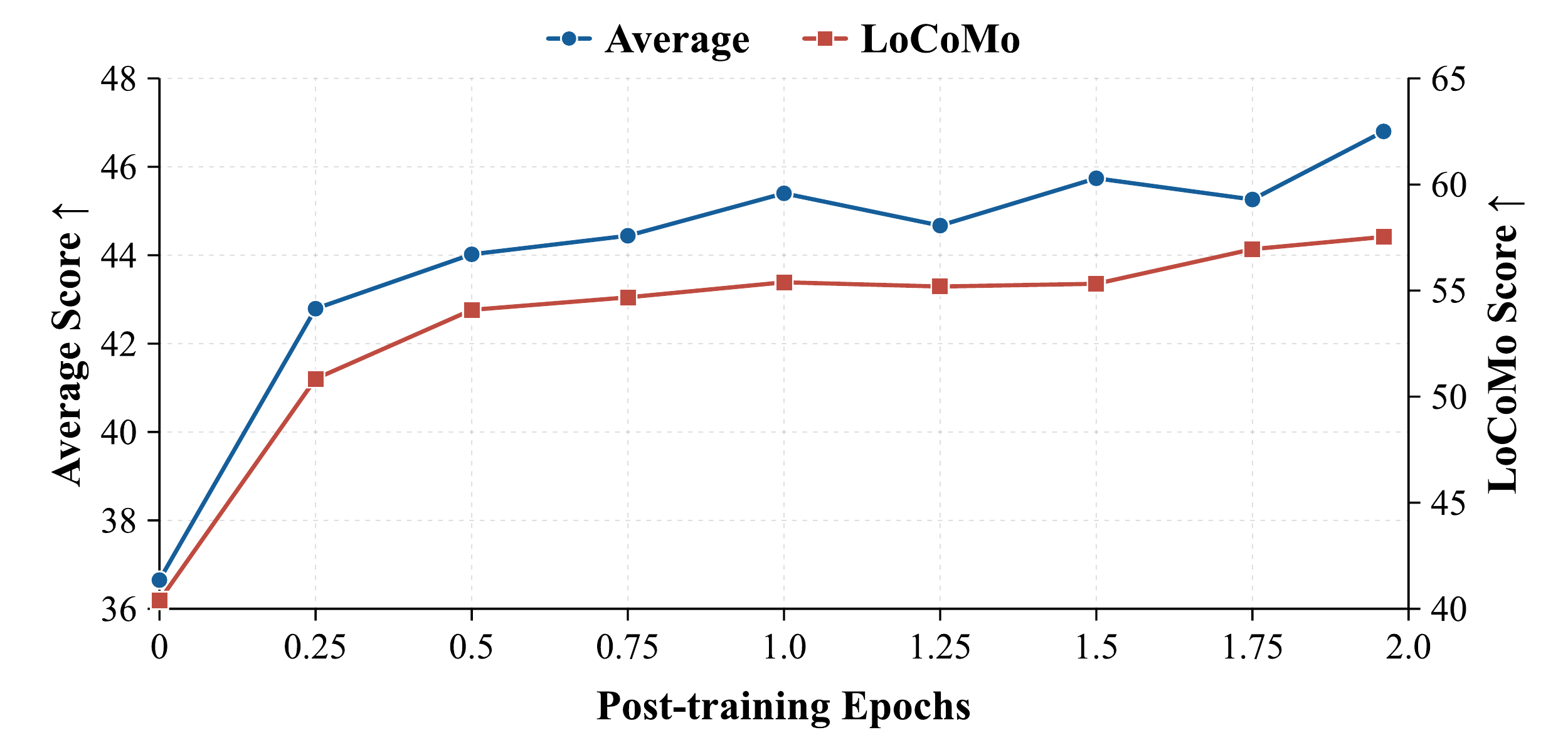}
    \vspace{-0.2cm}
    \caption{Data and training scaling of MemT-1.0-Nano.
    Memory performance over two consecutive post-training epochs.
    Blue circles show the average score on the left axis, and red squares show LoCoMo accuracy on the right axis.}
    \vspace{-0.4cm}
    \label{fig:training_data_scaling}
\end{figure}
Figure~\ref{fig:training_data_scaling} shows that memory performance generally improves with additional post-training.
Across the sampled checkpoints, the average score increases from 36.65 to 46.80, with intermediate fluctuations.
Most of this gain arrives early: the average score rises steeply within the first quarter epoch and continues to climb over the remainder of the first epoch, reaching 45.40 by its end, while LoCoMo improves from 40.39 to 55.39 over the same span.
This pattern is consistent with rapid adaptation of the CPT-initialized memory modules to downstream query answering, after which the increments shrink steadily but do not disappear.
In the second epoch, the final sampled checkpoint reaches 46.80 on average and 57.53 on LoCoMo, an improvement over the last sampled checkpoint of the first epoch despite intermediate fluctuations.
These gains indicate remaining room for optimization at fixed memory capacity, even when training revisits previously seen data.
In summary, continued exposure to memory-related post-training data can improve native memory capability, with further gains achievable across epochs without increasing model capacity.

\subsection{Memory Capacity Scaling}
\begin{table*}[!t]
    \centering
    \caption{
    We compare MoS configurations with 2 and 4 memory experts across 11 benchmarks.
    The evaluation settings follow those described in Section~\ref{sec:evaluation}.
    Bold and underlined values indicate the best and second-best results.
    }
    \label{tab:mos_table}

    \begin{threeparttable}

    \renewcommand{\arraystretch}{1.16}
    \setlength{\tabcolsep}{3pt}

    \sisetup{
        detect-weight = true,
        detect-family = true,
        round-mode = places,
        round-precision = 2,
        round-pad = true,
        table-alignment-mode = none,
        table-number-alignment = center,
        table-text-alignment = center
    }

    \newcommand{\best}[1]{{\bfseries\num{#1}}}
    \newcommand{\secondbest}[1]{\underline{\num{#1}}}

    \newcolumntype{Y}{
        S[
            table-fixed-width = true,
            table-column-width = 2.7 em,
            table-alignment-mode = none,
            table-number-alignment = center,
            table-text-alignment = center
        ]
    }

    \begin{adjustbox}{max width=\textwidth}
    \begin{tabular}{
        l
        Y Y Y Y Y Y Y Y Y Y Y Y
    }

    \toprule[1.15pt]

    \rowcolor{MetisBlueDark}
    \color{white}\bfseries Method
    &
    \multicolumn{2}{c}{\color{white}\bfseries $<16$K}
    &
    \multicolumn{3}{c}{\color{white}\bfseries 16--32K}
    &
    \multicolumn{3}{c}{\color{white}\bfseries 32--64K}
    &
    \multicolumn{3}{c}{\color{white}\bfseries $>64$K}
    &
    \multicolumn{1}{c}{\color{white}\bfseries All}
    \\

    \rowcolor{MetisBlueLight}
    &
    \multicolumn{1}{c}{\bfseries MB}
    &
    \multicolumn{1}{c}{\bfseries IM}
    &
    \multicolumn{1}{c}{\bfseries PM}
    &
    \multicolumn{1}{c}{\bfseries PSB}
    &
    \multicolumn{1}{c}{\bfseries LCM}
    &
    \multicolumn{1}{c}{\bfseries PMS}
    &
    \multicolumn{1}{c}{\bfseries MMR}
    &
    \multicolumn{1}{c}{\bfseries MO}
    &
    \multicolumn{1}{c}{\bfseries LME}
    &
    \multicolumn{1}{c}{\bfseries MAB}
    &
    \multicolumn{1}{c}{\bfseries BEAM}
    &
    \multicolumn{1}{c}{\bfseries AVG}
    \\

    \addlinespace[1pt]

    \rowcolor{MetisHighlight}
    \textcolor{MetisBlueDark}{\bfseries MemT-1.0-Nano}
    & 71.75 & 43.11
    & \secondbest{52.29} & 95.71 & \secondbest{57.53}
    & \secondbest{38.67} & {\best{23.92}} & 49.85
    & \best{32.60} & \secondbest{15.48} & 35.78
    & 46.97
    \\

    \rowcolor{MetisHighlight}
    \textcolor{MetisBlueDark}{\bfseries MemT-1.0-Nano-E2}
    & {\secondbest{74.40}} & {\best{50.01}}
    & {{50.08}} & {{\secondbest{96.67}}} & 57.21
    & {{37.60}} & {\secondbest{20.11}} & {\secondbest{51.65}}
    & {\best{32.60}} & {{15.20}} & {\secondbest{36.98}}
    & {\secondbest{47.50}}
    \\

    \rowcolor{MetisHighlight}
    \textcolor{MetisBlueDark}{\bfseries MemT-1.0-Nano-E4}
    & {{\best{75.12}}} & {{\secondbest{45.65}}}
    & {{\best{54.33}}} & {{\best{97.80}}} & {\best{59.03}}
    & {\best{41.07}} & {{19.96}} & {\best{53.69}}
    & {\secondbest{29.20}} & {\best{15.91}} & {{\best{37.41}}}
    & {\best{48.11}}
    \\

    \bottomrule[1.15pt]

    \end{tabular}
    \end{adjustbox}

    \vspace{2pt}

    \begin{minipage}{\textwidth}
        \footnotesize
        \textit{Abbreviations.}
        \textbf{MB}: MemBench;
        \textbf{IM}: ImplicitMemBench;
        \textbf{PM}: PersonaMem;
        \textbf{PSB}: PS-Bench;
        \textbf{LCM}: LoCoMo;
        \textbf{PMS}: PerMemSafe;
        \textbf{MMR}: Memora;
        \textbf{MO}: MemOps;
        \textbf{LME}: LongMemEval-S;
        \textbf{MAB}: MemoryAgentBench;
        \textbf{BEAM}: BEAM-1M.
    \end{minipage}

    \end{threeparttable}
    \vspace{-0.4cm}
\end{table*}

We compare the final checkpoints of MemT-1.0-Nano and MoS configurations with 2 and 4 memory experts, trained on the same data with identical hyperparameters.
As shown in Table ~\ref{tab:mos_table}, the 4-expert configuration achieves an average score of 48.11, compared with 47.50 for 2 experts and 46.97 for MemT-1.0-Nano, and obtains the strongest results on eight of the eleven benchmarks.
Its gains over the 2-expert configuration are largest on PersonaMem and PerMemSafe, where scores increase from 50.08 to 54.33 and from 37.60 to 41.07, respectively.
On LoCoMo, 4 experts achieve 59.03, compared with 57.21 for 2 experts and 57.53 for MemT-1.0-Nano.
MemOps similarly improves from 49.85 for MemT-1.0-Nano to 51.65 and 53.69 for 2 and 4 experts.
However, the improvement is not uniform. The 2-expert configuration remains strongest on ImplicitMemBench with 50.01, compared with 45.65 for 4 experts, while MemT-1.0-Nano retains advantages on Memora and LongMemEval-S.
These results support memory capacity scaling while showing distinct strengths across configurations.
Across MemBench, PS-Bench, MemOps, and BEAM-1M, relative improvements over MemT-1.0-Nano range from 1.00\% to 3.69\% for 2 experts and from 2.18\% to 7.70\% for 4 experts, with mean per-benchmark relative gains of 2.92\% and 4.78\%, respectively.


\section{Evaluation}
\label{sec:evaluation}

\subsection{Overall Evaluation Setup}
We evaluate MemT-1.0 from the perspectives of memory and general capabilities.
The memory evaluation focuses on whether MemT-1.0 can answer questions based on its dynamic memory states, without access to prior textual context.
The evaluation on general capabilities examines the impact of the additional native memory mechanism on the model’s general capabilities.
These aspects provide an important foundation for self-evolving memory-native models.

\textbf{Benchmarks.}
We evaluate memory capabilities on 11 benchmarks and divide them into four groups according to trajectory length.
Specifically, the group below 16K includes MemBench~\cite{tan2025membench} and ImplicitMemBench~\cite{qin2026implicitmembench}, and the 16K-32K group includes PersonaMem~\cite{jiang2025know}, PS-Bench~\cite{guo2026personalization}, and LoCoMo~\cite{maharana2024evaluating}.
For longer trajectories, the 32K-64K group includes PerMemSafe~\cite{an2026permemsafe}, Memora~\cite{uddin2026recall}, and MemOps~\cite{hao2026memops}, and the group above 64K includes LongMemEval-S~\cite{wu2024longmemeval}, MemoryAgentBench~\cite{hu2026evaluating}, and BEAM-1M~\cite{tavakoli2025beyond}.
These benchmarks focus on long-term information retention and retrieval, dynamic memory updating and management, and personalized understanding.
In addition, we evaluate general capabilities on six benchmarks spanning four major dimensions.
Specifically, we evaluate knowledge and scientific reasoning with MMLU-Pro~\cite{wang2024mmlu} and GPQA Diamond~\cite{rein2023gpqa}, instruction following with IFEval~\cite{zhou2023instruction},
long-context understanding with LongBench v2~\cite{bai2025longbench}, and code generation with BigCodeBench-Hard~\cite{zhuo2025bigcodebench} and LiveCodeBench v6~\cite{jain2025livecodebench}.
These benchmarks collectively assess a wide range of general capabilities, thereby reflecting the impact of the additional native memory mechanism on the underlying backbone model.
More details are available in Appendix~\ref{appendix:benchmark_details}.

\textbf{Baselines.}
We compare MemT-1.0 with textual memory and parametric memory approaches.
Textual memory methods perform inference based on the original backbone model with full context and partial context settings.
Full context provides the complete historical text as memory in the input context, while DenseRAG retrieves a subset of relevant information as contextual evidence~\cite{karpukhin2020densepassageretrievalopendomain}.
We also include representative memory systems for comparison, including MemOS~\cite{li2025memos}, Mem0~\cite{chhikara2025mem0}, OpenViking~\cite{fu2026vikingmem}, and Graphiti~\cite{rasmussen2025zep}.
For parametric memory methods, no textual history is provided in the context, and the model performs inference using its own parametric memory.
Our parametric memory baselines include Temp-LoRA~\cite{wang2024greater}, Doc-to-LoRA~\cite{charakorn2026doc}, SHINE~\cite{liu2026shine}, $\delta$-mem~\cite{lei2026delta}, LiveMem~\cite{liu2026livemem}, MemoryLLM~\cite{wang2024memoryllm} and ICAE~\cite{ge2023context}.
More details of the baselines are provided in Appendix~\ref{appendix:baseline_details}.

\textbf{Evaluation Protocol.}
For memory capability evaluation, in each trajectory, we first reset the memory and write the historical information into memory following the chronological order.
Then, the model is required to answer multiple questions related to the historical information, and the responses are compared against the ground-truth answers to compute accuracy.
Specifically, memory is strictly isolated across different users and histories. 
Ground-truth answers, evidence annotations, and evaluation labels are only used for scoring and are never exposed to the evaluated models.
For LLM-as-a-Judge, we utilize \texttt{gpt-4.1-mini} with the OmniMemEval~\cite{memtensor2026omnimemeval} prompt.
For general capability evaluation, we introduce irrelevant information before inference, which aims to examine how the native memory mechanism affects the model’s general capabilities under the noise setting.
More details are available in Appendix~\ref{appendix:eval_general_pipeline_details}.

\begin{table*}[!t]
    \centering
    \caption{Memory capability evaluation across 11 benchmarks.
        Benchmarks are grouped according to trajectory length.
        NV denotes NVIDIA-Nemotron-3-Ultra-550B-A55B-BF16.
        DS indicates DeepSeek-V4-Flash, while QW means Qwen3.6-35B-A3B.
        Full Context (256K) uses a joint budget of 256K tokens, and Full Context (Max) uses each backbone's maximum context window.
        Bold and underlined values indicate the best and second-best results within each memory category, respectively.
    }
    \label{tab:memory_capability_main}

    \begin{threeparttable}

    \renewcommand{\arraystretch}{1.16}
    \setlength{\tabcolsep}{3pt}

    \sisetup{
        detect-weight = true,
        detect-family = true,
        round-mode = places,
        round-precision = 2,
        round-pad = true,
        table-alignment-mode = none,
        table-number-alignment = center,
        table-text-alignment = center
    }

    \newcommand{\best}[1]{{\bfseries\num{#1}}}
    \newcommand{\secondbest}[1]{\underline{\num{#1}}}

    \newcolumntype{Y}{
        S[
            table-fixed-width = true,
            table-column-width = 2.7 em,
            table-alignment-mode = none,
            table-number-alignment = center,
            table-text-alignment = center
        ]
    }

    \begin{adjustbox}{max width=\textwidth}
    \begin{tabular}{
        l
        c
        Y Y Y Y Y Y Y Y Y Y Y Y
    }

    \toprule[1.15pt]

    \rowcolor{MetisBlueDark}
    \color{white}\bfseries Method
    &
    \color{white}\bfseries Backbone
    &
    \multicolumn{2}{c}{\color{white}\bfseries $<16$K}
    &
    \multicolumn{3}{c}{\color{white}\bfseries 16--32K}
    &
    \multicolumn{3}{c}{\color{white}\bfseries 32--64K}
    &
    \multicolumn{3}{c}{\color{white}\bfseries $>64$K}
    &
    \multicolumn{1}{c}{\color{white}\bfseries All}
    \\

    \rowcolor{MetisBlueLight}
    &
    &
    \multicolumn{1}{c}{\bfseries MB}
    &
    \multicolumn{1}{c}{\bfseries IM}
    &
    \multicolumn{1}{c}{\bfseries PM}
    &
    \multicolumn{1}{c}{\bfseries PSB}
    &
    \multicolumn{1}{c}{\bfseries LCM}
    &
    \multicolumn{1}{c}{\bfseries PMS}
    &
    \multicolumn{1}{c}{\bfseries MMR}
    &
    \multicolumn{1}{c}{\bfseries MO}
    &
    \multicolumn{1}{c}{\bfseries LME}
    &
    \multicolumn{1}{c}{\bfseries MAB}
    &
    \multicolumn{1}{c}{\bfseries BEAM}
    &
    \multicolumn{1}{c}{\bfseries AVG}
    \\

    \midrule[0.65pt]

    \rowcolor{MetisBluePale}
    \multicolumn{14}{c}{
        \textcolor{MetisBlueDark}{\bfseries Textual Memory}
    }
    \\

    \addlinespace[1pt]

    \multirow{3}{*}{MemOS}
    & NV
    & {\num{79.37}} & {\secondbest{67.28}}
    & {\num{71.48}} & {\num{92.49}} & {\best{87.79}}
    & {\best{70.67}} & {\best{55.57}} & {\num{82.73}}
    & {\num{80.60}} & {\num{27.12}} & {\best{77.76}}
    & {\best{72.08}}
    \\

    & DS
    & 78.38 & {\best{68.15}}
    & 67.06 & 92.86 & {\secondbest{85.65}}
    & 45.87 & {\secondbest{55.52}} & 76.95
    & {\best{83.20}} & 27.78 & {\num{75.43}}
    & {\num{68.80}}
    \\

    & QW
    & 79.87 & 61.14
    & 67.57 & 97.32 & 83.38
    & 54.67 & 52.61 & 82.56
    & 80.80 & 21.78 & {\num{74.28}}
    & {\num{68.73}}
    \\

    \addlinespace[2pt]

    \multirow{3}{*}{Mem0}
    & NV
    & {\num{79.73}} & {\num{63.92}}
    & {\secondbest{72.33}} & {\num{94.04}} & {\num{81.69}}
    & {\secondbest{68.80}} & {\num{45.69}} & {\num{80.31}}
    & {\num{77.80}} & {\num{32.80}} & {\num{75.44}}
    & {\secondbest{70.23}}
    \\

    & DS
    & 75.79 & 45.85
    & 67.23 & 89.43 & 78.96
    & 66.00 & 47.52 & 75.67
    & 72.60 & 30.30 & {\num{74.19}}
    & {\num{65.78}}
    \\

    & QW
    & 78.52 & 60.41
    & 63.16 & 97.32 & 80.91
    & 54.93 & 41.40 & 79.42
    & 78.90 & 20.56 & {\num{72.95}}
    & {\num{66.23}}
    \\

    \addlinespace[2pt]

    \multirow{3}{*}{OpenViking}
    & NV
    & {\num{82.21}} & {\num{65.54}}
    & {\num{62.65}} & {\num{93.34}} & {\num{84.94}}
    & {\num{60.13}} & {\num{52.77}} & {\num{82.34}}
    & {\num{81.00}} & {\num{25.45}} & {\secondbest{77.72}}
    & {\num{69.83}}
    \\

    & DS
    & 75.72 & 41.39
    & 65.37 & 95.44 & 79.09
    & 53.20 & 53.43 & 77.27
    & 75.00 & 28.15 & {\num{74.21}}
    & {\num{65.30}}
    \\

    & QW
    & 76.77 & 53.03
    & 61.63 & 95.92 & 72.92
    & 53.73 & 51.45 & 81.74
    & {\secondbest{82.00}} & 23.20 & {\num{70.05}}
    & {\num{65.68}}
    \\

    \addlinespace[2pt]

    \multirow{3}{*}{Graphiti}
    & NV
    & {\num{59.61}} & {\num{48.74}}
    & {\num{59.59}} & {\num{91.68}} & {\num{60.65}}
    & {\num{30.53}} & {\num{30.18}} & {\num{52.81}}
    & {\num{42.60}} & {\num{16.68}} & {\num{63.38}}
    & {\num{50.59}}
    \\

    & DS
    & 49.58 & 39.41
    & 53.65 & 89.00 & 66.69
    & 30.80 & 26.22 & 51.08
    & 57.00 & 19.85 & {\num{64.41}}
    & {\num{49.79}}
    \\

    & QW
    & 61.93 & 34.94
    & 49.75 & 96.51 & 50.58
    & 24.53 & 23.04 & 47.43
    & 38.10 & 15.52 & {\num{64.33}}
    & {\num{46.06}}
    \\

    \addlinespace[2pt]

    \multirow{3}{*}{Full Context (256K)}
    & NV
    & {\best{84.36}} & {\num{51.71}}
    & {\best{73.68}} & {\num{95.49}} & {\num{85.19}}
    & {\num{56.13}} & {\num{48.73}} & {\num{88.60}}
    & {\num{59.80}} & {\num{38.33}} & {\num{38.61}}
    & {\num{65.51}}
    \\

    & DS
    & 83.23 & 58.04
    & 65.20 & 97.37 & 79.87
    & 32.27 & 51.65 & {\secondbest{88.73}}
    & 77.80 & 35.98 & 44.65
    & 64.98
    \\

    & QW
    & {\secondbest{84.32}} & 53.63
    & 65.20 & {\best{98.66}} & 83.77
    & 44.80 & 49.79 & {\best{92.40}}
    & 76.80 & 35.94 & 43.66
    & 66.27
    \\

    \addlinespace[2pt]

    \multirow{3}{*}{Full Context (Max)}
    & NV
    & {\best{84.36}} & {\num{51.71}}
    & {\best{73.68}} & {\num{95.49}} & {\num{85.19}}
    & {\num{56.13}} & {\num{48.73}} & {\num{88.60}}
    & {\num{59.80}} & {\secondbest{38.57}} & {\num{45.66}}
    & {\num{66.17}}
    \\

    & DS
    & {\num{83.23}} & {\num{58.04}}
    & {\num{65.20}} & {\num{97.37}} & {\num{79.87}}
    & {\num{32.27}} & {\num{51.65}} & {\secondbest{88.73}}
    & {\num{77.80}} & {\best{42.85}} & {\num{57.35}}
    & {\num{66.76}}
    \\

    & QW
    & {\secondbest{84.32}} & {\num{53.63}}
    & {\num{65.20}} & {\best{98.66}} & {\num{83.77}}
    & {\num{44.80}} & {\num{49.79}} & {\best{92.40}}
    & {\num{76.80}} & {\num{35.77}} & {\num{52.22}}
    & {\num{67.03}}
    \\

    \addlinespace[2pt]

    \multirow{3}{*}{DenseRAG}
    & NV
    & {\num{66.64}} & {\num{46.88}}
    & {\num{64.86}} & {\num{92.70}} & {\num{60.84}}
    & {\num{62.93}} & {\num{32.04}} & {\num{71.78}}
    & {\num{71.80}} & {\num{33.09}} & {\num{46.13}}
    & {\num{59.06}}
    \\

    & DS
    & 69.30 & 55.03
    & 59.42 & 89.05 & 58.31
    & 55.47 & 31.40 & 68.53
    & 69.40 & 30.48 & 43.71
    & 57.28
    \\

    & QW
    & 67.09 & 52.15
    & 60.27 & {\secondbest{97.69}} & 61.10
    & 45.20 & 31.28 & 70.55
    & 70.80 & 28.17 & 37.09
    & 56.49
    \\

    \addlinespace[2pt]
    \midrule[0.5pt]

    \rowcolor{MetisBluePale}
    \multicolumn{14}{c}{
        \textcolor{MetisBlueDark}{\bfseries Parametric Memory}
    }
    \\

    \addlinespace[1pt]

    \rowcolor{MetisHighlightStrong}
    \textcolor{MetisBlueDark}{\bfseries MemT-1.0-Pro}
    & --
    & {\num{71.37}} & {\secondbest{46.57}}
    & {\best{67.74}} & {\best{99.03}} & {\secondbest{64.61}}
    & {\best{52.80}} & {\num{18.63}} & {\secondbest{63.62}}
    & {\secondbest{33.80}} & {\best{23.87}} & {\best{41.80}}
    & {\best{53.08}}
    \\

    \rowcolor{MetisHighlightStrong}
    \textcolor{MetisBlueDark}{\bfseries MemT-1.0-Flash}
    & --
    & \best{73.48} & {\best{49.21}}
    & {\secondbest{61.63}} & {97.48} & {\best{65.13}}
    & 33.47 & {\secondbest{20.54}} & {\best{64.06}}
    & {\best{37.20}} & {\secondbest{17.36}} & {\secondbest{37.04}}
    & {\secondbest{50.60}}
    \\

    \rowcolor{MetisHighlight}
    \textcolor{MetisBlueDark}{\bfseries MemT-1.0-Nano}
    & --
    & \secondbest{71.75} & 43.11
    & 52.29 & 95.71 & 57.53
    & {\secondbest{38.67}} & {\best{23.92}} & 49.85
    & 32.60 & 15.48 & 35.78
    & 46.97
    \\

    \addlinespace[2pt]

    Temp-LoRA
    & --
    & 44.25 & 32.03
    & 57.39 & {\secondbest{98.01}} & 25.13
    & 35.73 & 16.17 & 52.50
    & 13.40 & 11.98 & 32.32
    & 38.08
    \\

    Doc-to-LoRA
    & --
    & 57.19 & 42.42
    & 33.96 & 70.21 & 24.48
    & 16.53 & 15.31 & 36.01
    & 14.60 & 3.54 & 25.99
    & 30.93
    \\

    SHINE
    & --
    & 40.70 & 20.24
    & 39.90 & 63.29 & 26.69
    & 4.40 & 16.74 & 39.59
    & 13.80 & 6.15 & 9.58
    & 25.55
    \\

    $\delta$-mem
    & --
    & 37.48 & 29.92
    & 46.86 & 90.82 & 20.26
    & 20.13 & 17.89 & 40.92
    & 14.80 & 7.13 & 17.03
    & 31.20
    \\

    LiveMem-RL
    & --
    & 35.67 & 23.31
    & 46.01 & 82.02 & 24.42
    & 10.53 & 16.37 & 39.83
    & 14.60 & 7.27 & 27.08
    & 29.74
    \\

    MemoryLLM
    & --
    & 22.71 & 17.76
    & 41.09 & 37.41 & 46.62
    & 3.33 & 16.26 & 46.80
    & 15.60 & 6.26 & 24.01
    & 25.26
    \\

    ICAE
    & --
    & 31.90 & 24.49
    & 17.83 & 70.37 & 22.40
    & 2.80 & 11.35 & 40.41
    & 11.00 & 7.35 & 11.69
    & 22.87
    \\

    \bottomrule[1.15pt]

    \end{tabular}
    \end{adjustbox}

    \vspace{2pt}

    \begin{minipage}{\textwidth}
        \footnotesize
        \textit{Abbreviations.}
        \textbf{MB}: MemBench;
        \textbf{IM}: ImplicitMemBench;
        \textbf{PM}: PersonaMem;
        \textbf{PSB}: PS-Bench;
        \textbf{LCM}: LoCoMo;
        \textbf{PMS}: PerMemSafe;
        \textbf{MMR}: Memora;
        \textbf{MO}: MemOps;
        \textbf{LME}: LongMemEval-S;
        \textbf{MAB}: MemoryAgentBench;
        \textbf{BEAM}: BEAM-1M.
    \end{minipage}

    \end{threeparttable}
    \vspace{-0.4cm}
\end{table*}

\subsection{Memory Capabilities}
The results for memory capabilities are presented in Table~\ref{tab:memory_capability_main}.
MemT-1.0-Pro achieves an average score of 53.08, ranking first among the evaluated parametric memory methods.
It obtains the strongest results in this group on PersonaMem, PS-Bench, PerMemSafe, MemoryAgentBench, and BEAM-1M.
Its gains over MemT-1.0-Flash are largest on PerMemSafe and also extend to personalized understanding and long-history tasks.
However, the improvement is not uniform, where MemT-1.0-Flash remains stronger on several benchmarks like MemBench, ImplicitMemBench, and MemOps.
These results indicate complementary strengths across the model family.
MemT-1.0-Flash and MemT-1.0-Nano achieve average scores of 50.60 and 46.97, respectively, ranking second and third among the evaluated parametric memory methods and exceeding the strongest non-MemT parametric baseline, Temp-LoRA, by 12.52 and 8.89 points.
The results support native memory as a shared representation for diverse memory tasks without providing historical text in the query context.
The comparison between MemT-1.0-Flash and MemT-1.0-Nano reveals distinct strengths across tasks, though the aggregate gap between them remains modest and is driven mainly by memory operations.
MemT-1.0-Flash improves most clearly on memory operations, personalized understanding, conversational recall, and implicit learning.
MemT-1.0-Nano remains stronger on PerMemSafe and Memora, whereas MemT-1.0-Flash scores higher on Persona-grounded PS-Bench.
The advantage over parametric memory baselines also extends to long histories, where MemT-1.0 leads this group on LongMemEval-S, MemoryAgentBench, and BEAM-1M.
However, the remaining errors on these benchmarks show that processing an extended trajectory does not by itself ensure reliable access to its contents.
Answering later queries requires earlier information to survive successive updates and remain distinguishable from other stored associations.
The results motivate further improvements in selective retention and query-dependent readout.

The strongest textual memory systems still achieve substantially higher performance, particularly on LongMemEval-S and BEAM-1M.
These systems provide explicit historical evidence at query time, whereas MemT-1.0 must recover the required information from its evolving states.
This gap is consistent with the difficulty of preserving detailed evidence through recurrent memory updates and highlights memory fidelity as an important target for further development.

\subsection{General Capabilities}
We evaluate general capabilities under the noise setting.
In this setting, we inject irrelevant information into memory before the evaluation tasks, which simulates the impact of distracting memories during real-world online interactions.
We report the performance gap between MemT-1.0 and its corresponding backbone model under the noise setting.
The results are presented in Table~\ref{tab:general_capability}.

MemT-1.0-Pro has an average decrease of 0.11 percentage points, compared with its backbone under the same noise setting.
It shows improvements on MMLU-Pro, GPQA Diamond, and IFEval.
The largest decrease occurs on LongBench v2, followed by smaller decreases on BigCodeBench-Hard and LiveCodeBench v6.
The minor decrease therefore masks task-dependent changes, with gains in knowledge, scientific reasoning, and instruction following largely offsetting regressions in long-context understanding and code generation.

MemT-1.0-Flash performs comparably to its DeepSeek-V4-Flash backbone on average under the same noise setting.
The largest gain occurs on BigCodeBench-Hard, with smaller improvements on IFEval and LongBench v2.
Performance decreases modestly on MMLU-Pro, GPQA Diamond, and LiveCodeBench v6.
These results suggest that MemT-1.0-Flash broadly preserves general capabilities under distracting memories.

We next examine MemT-1.0-Nano, whose average decrease is around one point.
Under the noise setting, most benchmarks remain close to the backbone, with gains on IFEval and BigCodeBench-Hard.
The degradation is concentrated rather than uniform.
The largest decrease occurs on LiveCodeBench v6, followed by smaller decreases on LongBench v2, MMLU-Pro, and GPQA Diamond.
The contrasting changes on BigCodeBench-Hard and LiveCodeBench v6 further suggest that the impact varies even across benchmarks within the same capability domain.

Overall, the results support broad capability preservation alongside native memory, while highlighting that resistance to distracting memories should be evaluated at the task level.

\begin{table*}[t]
    \centering
    \caption{
        General capability evaluation under the backbone and noise settings.
    }
    \vspace{-0.2cm}
    \label{tab:general_capability}

    \begin{threeparttable}

    \renewcommand{\arraystretch}{1.16}

    \sisetup{
        round-mode = places,
        round-precision = 2,
        round-pad = true,
        table-number-alignment = center
    }

    \newcommand{\gcDown}[1]{%
        \multicolumn{1}{c}{%
            \textcolor{MetisDeltaDown}{\ensuremath{-#1}}%
        }%
    }
    \newcommand{\gcUp}[1]{%
        \multicolumn{1}{c}{%
            \textcolor{MetisDeltaUp}{\ensuremath{+#1}}%
        }%
    }
    \newcommand{\gcZero}{\multicolumn{1}{c}{0.0}}

    \begin{adjustbox}{max width=\textwidth}
    \begin{tabular}{
        l
        *{7}{S[
            table-format = 2.2,
            table-fixed-width = true,
            table-column-width = 4.5 em
        ]}
    }

    \toprule[1.15pt]

    \rowcolor{MetisBlueDark}
    \color{white}\bfseries Model
    & \multicolumn{1}{c}{\color{white}\bfseries MP}
    & \multicolumn{1}{c}{\color{white}\bfseries GPQA}
    & \multicolumn{1}{c}{\color{white}\bfseries IFE}
    & \multicolumn{1}{c}{\color{white}\bfseries LB}
    & \multicolumn{1}{c}{\color{white}\bfseries BCB-H}
    & \multicolumn{1}{c}{\color{white}\bfseries LCB-v6}
    & \multicolumn{1}{c}{\color{white}\bfseries AVG}
    \\

    \midrule[0.65pt]

    \rowcolor{MetisBluePale}
    \multicolumn{8}{c}{
        \textcolor{MetisBlueDark}{\bfseries Backbone}
    }
    \\

    \addlinespace[1pt]

    Nemotron-3-Ultra-550B-A55B
    & {\num{84.04}}
    & {\num{84.91}}
    & {\num{97.23}}
    & {\num{59.64}}
    & {\num{29.73}}
    & {\num{92.23}}
    & {\num{74.63}}
    \\

    DeepSeek-V4-Flash
    & 85.68 & 91.16 & 93.53 & 63.02 & 35.81 & 87.49 & 76.12
    \\

    Qwen3.6-35B-A3B
    & 84.58 & 83.90 & 90.57 & 57.26 & 27.03 & 87.96 & 71.88
    \\

    \addlinespace[2pt]
    \midrule[0.5pt]

    \rowcolor{MetisBluePale}
    \multicolumn{8}{c}{
        \textcolor{MetisBlueDark}{\bfseries Noise Setting}
    }
    \\

    \addlinespace[1pt]

    Nemotron-3-Ultra-550B-A55B
    & {\num{81.29}}
    & {\num{84.66}}
    & {\num{90.76}}
    & {\num{60.83}}
    & {\num{29.73}}
    & {\num{91.75}}
    & {\num{73.17}}
    \\

    \rowcolor{MetisHighlightStrong}
    \textcolor{MetisBlueDark}{\bfseries MemT-1.0-Pro}
    & \gcUp{4.22}
    & \gcUp{0.88}
    & \gcUp{1.11}
    & \gcDown{5.17}
    & \gcDown{1.35}
    & \gcDown{0.38}
    & \gcDown{0.11}
    \\

    DeepSeek-V4-Flash
    & 83.81 & 89.58 & 92.98 & 64.21 & 32.43 & 89.19 & 75.37
    \\

    \rowcolor{MetisHighlightStrong}
    \textcolor{MetisBlueDark}{\bfseries MemT-1.0-Flash}
    & \gcDown{1.8}
    & \gcDown{0.6}
    & \gcUp{0.9}
    & \gcUp{0.2}
    & \gcUp{2.0}
    & \gcDown{0.7}
    & \gcZero
    \\

    Qwen3.6-35B-A3B
    & 84.42 & 80.05 & 89.28 & 55.67 & 25.68 & 88.06 & 70.53
    \\

    \rowcolor{MetisHighlight}
    \textcolor{MetisBlueDark}{\bfseries MemT-1.0-Nano}
    & \gcDown{1.6}
    & \gcDown{1.5}
    & \gcUp{3.3}
    & \gcDown{2.0}
    & \gcUp{2.0}
    & \gcDown{5.8}
    & \gcDown{0.9}
    \\

    \bottomrule[1.15pt]

    \end{tabular}
    \end{adjustbox}

    \vspace{2pt}

    \begin{minipage}{\textwidth}
        \footnotesize
        \textit{Abbreviations.}
        \textbf{MP}: MMLU-Pro;
        \textbf{GPQA}: GPQA Diamond;
        \textbf{IFE}: IFEval;
        \textbf{LB}: LongBench v2;
        \textbf{BCB-H}: BigCodeBench-Hard;
        \textbf{LCB-v6}: LiveCodeBench v6.
        \textcolor{MetisDeltaDown}{Green} indicates a decrease,
        while \textcolor{MetisDeltaUp}{red} indicates an increase.
    \end{minipage}

    \end{threeparttable}
    \vspace{-0.4cm}
\end{table*}

\subsection{Ablation Studies}
In this section, we further present several ablation studies based on MemT-1.0-Nano during model development.
As for the model architecture, we focus on the update rule and update granularity in the native memory procedure.
We also design a bidirectional encoding mechanism to aggregate information from the current turn.
In addition, we study the contribution of different training stages.
These experiments further verify the effectiveness of our current architecture.

\subsubsection{Architecture Ablation}

For architecture ablation, we evaluate four variants.
First, we remove the stable mechanism to recover the GDN-2 memory update (w/ GDN-2) by directly using
\begin{equation*}
    \tilde{\mathbf{M}}_{t, i, h}^{(l)} = \left[
\mathbf{I}
-
\mathbf{K}_{t,i,h}^{(l)\top}
\left(
\bm{\beta}_{t,i,h}^{(l)}
\odot
\mathbf{K}_{t,i,h}^{(l)}
\right)
\right]
\cdot
\mathbf{D}_{t,i,h}^{(l)}
\tilde{\mathbf{M}}_{t,i-1,h}^{(l)}
+
\mathbf{K}_{t,i,h}^{(l)\top}
\left(
\bm{\omega}_{t,i,h}^{(l)}
\odot
\mathbf{V}_{t,i,h}^{(l)}
\right).
\end{equation*}
We also change the update rule into GDN~\cite{yang2025gateddeltanetworksimproving} as another ablation baseline (w/ GDN).
Second, we change the update granularity from token-level to turn-level (w/ TL).
The storage procedure can be defined as
\begin{equation*}
    \mathbf{M}_{t+1,h}^{(l)} = \left[
\mathbf{I}
-
\frac{1}{L_t}
\mathbf{U}_{t,h}^{(l)\top}
\mathbf{U}_{t,h}^{(l)}
\right]
\cdot
\mathbf{D}_{t,h}^{(l)}
\mathbf{M}_{t,h}^{(l)}
+
\frac{1}{L_t}
\mathbf{U}_{t,h}^{(l)\top}
\left(
\bm{\omega}_{t,h}^{(l)}
\odot
\mathbf{V}_{t,h}^{(l)}
\right),
\end{equation*}
where we have 
\begin{equation*}
    \mathbf{D}_{t,h}^{(l)} = \operatorname{Diag}
\left[
\exp
\left(
\frac{1}{L_t}
\sum_{i=1}^{L_t}
\bm{\lambda}_{t,i,h}^{(l)}
\right)
\right].
\end{equation*}
We also evaluate a bidirectional encoding (w/ BE) for information aggregation.
Specifically, during the storage procedure in each turn, it first projects the pre-attention hidden states $\tilde{\mathbf{H}}_t^{(l)}$ into a lower-dimensional space, and applies multi-head self-attention with rotary position embeddings without a causal mask.
After that, it projects the result back to the backbone dimension through a residual connection.
The enriched representations are used to predict the erase, write, and forget gates, while the memory keys and values are still projected from the original pre-attention hidden states.

As shown in Table~\ref{tab:ablation_results}, the update granularity has the largest effect among the architecture variants.
Replacing token-level with turn-level aggregation reduces the average from 46.97 to 40.98, a far larger drop than any change to the update rule.
This may be because successive token-level updates can better account for memory decay and revise associations established earlier within the same turn.
The update-rule variants, by contrast, stay close to the full model, with removing the stable mechanism giving 46.43 and substituting GDN giving 46.72.
Both fall within a narrow band around the baseline, so the aggregate capability is not determined by the precise form of the update rule.
Adding bidirectional encoding reduces the average to 45.43.
Its effect is benchmark-dependent rather than uniformly negative: MB, IM, PM, and PMS improve, whereas the remaining benchmarks regress.
Under the evaluated training budget, bidirectional encoding does not improve aggregate memory performance, and its benchmark-specific gains do not offset the losses elsewhere.

\subsubsection{Training Ablation}
We further conduct ablation studies on our training procedures.
For the w/o CPT setting, we skip CPT and directly train the memory parameters through post-training alone.
For the w/o PT setting, we evaluate the checkpoint after continued pre-training without subsequent post-training.
From the results in Table~\ref{tab:ablation_results}, we find that post-training plays an important role in memory-specific tasks, where it increases the average score by approximately 10.69 points over the CPT checkpoint.
These results indicate that language-modeling CPT alone does not provide the same memory-conditioned response capability as the complete pipeline.
In addition, the CPT stage improves post-training as well.
Therefore, post-training accounts for the larger observed gain in this comparison, while CPT provides an additional aggregate benefit as an initialization stage.

\begin{table*}[t]
    \centering
    \caption{
        Architecture and training ablations of MemT-1.0-Nano.
        Bold and underlined values indicate the best and
        second-best results, respectively.
        $\Delta$ AVG denotes the score difference relative
        to MemT-1.0-Nano.
    }
    \label{tab:ablation_results}

    \begin{threeparttable}

    \setlength{\tabcolsep}{3.2pt}
    \renewcommand{\arraystretch}{1.16}

    \sisetup{
        detect-weight = true,
        detect-family = true,
        round-mode = places,
        round-precision = 2,
        round-pad = true,
        table-alignment-mode = none,
        table-number-alignment = center,
        table-text-alignment = center
    }

    \newcommand{\abfirst}[1]{{\bfseries\num{#1}}}
    \newcommand{\absecond}[1]{{\underline{\num{#1}}}}

    \newcolumntype{Y}{
        S[
            table-fixed-width = true,
            table-column-width = 2.8em,
            table-alignment-mode = none,
            table-number-alignment = center,
            table-text-alignment = center
        ]
    }

    \newcolumntype{Z}{
        S[
            table-fixed-width = true,
            table-column-width = 3.8em,
            table-alignment-mode = none,
            table-number-alignment = center,
            table-text-alignment = center
        ]
    }

    \begin{adjustbox}{max width=\textwidth}
    \begin{tabular}{
        l
        Y Y Y Y Y Y Y Y Y Y Y Y Z
    }

    \toprule[1.15pt]

    \rowcolor{MetisBlueDark}
    \color{white}\bfseries Variant
    &
    \multicolumn{2}{c}{\color{white}\bfseries $<16$K}
    &
    \multicolumn{3}{c}{\color{white}\bfseries 16--32K}
    &
    \multicolumn{3}{c}{\color{white}\bfseries 32--64K}
    &
    \multicolumn{3}{c}{\color{white}\bfseries $>64$K}
    &
    \multicolumn{2}{c}{\color{white}\bfseries All}
    \\

    \rowcolor{MetisBlueLight}
    &
    \multicolumn{1}{c}{\bfseries MB}
    &
    \multicolumn{1}{c}{\bfseries IM}
    &
    \multicolumn{1}{c}{\bfseries PM}
    &
    \multicolumn{1}{c}{\bfseries PSB}
    &
    \multicolumn{1}{c}{\bfseries LCM}
    &
    \multicolumn{1}{c}{\bfseries PMS}
    &
    \multicolumn{1}{c}{\bfseries MMR}
    &
    \multicolumn{1}{c}{\bfseries MO}
    &
    \multicolumn{1}{c}{\bfseries LME}
    &
    \multicolumn{1}{c}{\bfseries MAB}
    &
    \multicolumn{1}{c}{\bfseries BEAM}
    &
    \multicolumn{1}{c}{\bfseries AVG}
    &
    \multicolumn{1}{c}{\bfseries $\Delta$ AVG}
    \\

    \midrule[0.65pt]

    \rowcolor{MetisHighlight}
    \textcolor{MetisBlueDark}{\bfseries MemT-1.0-Nano}
    & \num{71.75}
    & 43.11
    & 52.29
    & \num{95.71}
    & {\abfirst{57.53}}
    & 38.67
    & {\abfirst{23.92}}
    & 49.85
    & {\abfirst{32.60}}
    & {\abfirst{15.48}}
    & {\abfirst{35.78}}
    & \abfirst{46.97}
    & 0.00
    \\

    \addlinespace[2pt]
    \midrule[0.5pt]

    \rowcolor{MetisBluePale}
    \multicolumn{14}{c}{
        \textcolor{MetisBlueDark}{
            \bfseries Architecture Ablation
        }
    }
    \\

    \addlinespace[1pt]

    w/ GDN-2
    & \num{71.58}
    & {\absecond{43.74}}
    & {\abfirst{55.69}}
    & \num{96.24}
    & 57.34
    & 36.93
    & {\absecond{21.71}}
    & 49.15
    & 29.40
    & 14.38
    & 34.59
    & \num{46.43}
    & \num{-0.54}
    \\

    \addlinespace[2pt]

    w/ GDN
    & \absecond{72.27}
    & 41.92
    & 54.50
    & \abfirst{98.34}
    & {\absecond{57.47}}
    & {\abfirst{40.93}}
    & 20.30
    & {\abfirst{50.60}}
    & {\absecond{30.00}}
    & 13.43
    & 34.15
    & \absecond{46.72}
    & \num{-0.25}
    \\

    \addlinespace[2pt]

    w/ BE
    & \abfirst{73.88}
    & {\abfirst{44.73}}
    & 54.33
    & \num{94.85}
    & 55.39
    & {\absecond{40.80}}
    & 16.27
    & 45.51
    & 25.60
    & {\absecond{14.44}}
    & 33.97
    & \num{45.43}
    & \num{-1.54}
    \\

    \addlinespace[2pt]

    w/ TL
    & \num{69.12}
    & 39.46
    & 52.12
    & \absecond{97.48}
    & 52.86
    & 36.40
    & 17.15
    & 38.29
    & 25.00
    & 5.48
    & 17.39
    & \num{40.98}
    & \num{-5.99}
    \\

    \addlinespace[2pt]
    \midrule[0.5pt]

    \rowcolor{MetisBluePale}
    \multicolumn{14}{c}{
        \textcolor{MetisBlueDark}{
            \bfseries Training Ablation
        }
    }
    \\

    \addlinespace[1pt]

    w/o CPT
    & \num{69.80}
    & 37.42
    & {\absecond{54.67}}
    & \num{96.78}
    & 55.52
    & 35.33
    & 18.68
    & {\absecond{50.15}}
    & 27.00
    & 13.37
    & {\absecond{34.66}}
    & \num{44.85}
    & \num{-2.12}
    \\

    \addlinespace[2pt]

    w/o PT
    & \num{66.19}
    & 38.69
    & 41.26
    & \num{90.23}
    & 40.39
    & 26.27
    & 16.09
    & 24.63
    & 13.80
    & 12.21
    & 29.36
    & \num{36.28}
    & \num{-10.69}
    \\

    \bottomrule[1.15pt]

    \end{tabular}
    \end{adjustbox}

    \end{threeparttable}
    \vspace{-0.4cm}
\end{table*}

\subsection{End-to-End Inference Efficiency}
\label{sec:inference_efficiency}
We further evaluate the end-to-end inference efficiency of MemT-1.0 against baseline methods, with a particular focus on the results as the history length increases.
Specifically, we compare MemT-1.0-Nano with full context, $\delta$-mem, and Temp-LoRA on BEAM.
Full context, Temp-LoRA and MemT-1.0-Nano use HF inference on two GPUs exclusively allocated to each run, with backbone layers distributed across devices, a concurrency of 1, and a batch size of 1. For $\delta$-mem, we use its publicly released model and implementation.

\begin{figure}[t]
    \centering
    \includegraphics[width=\linewidth]{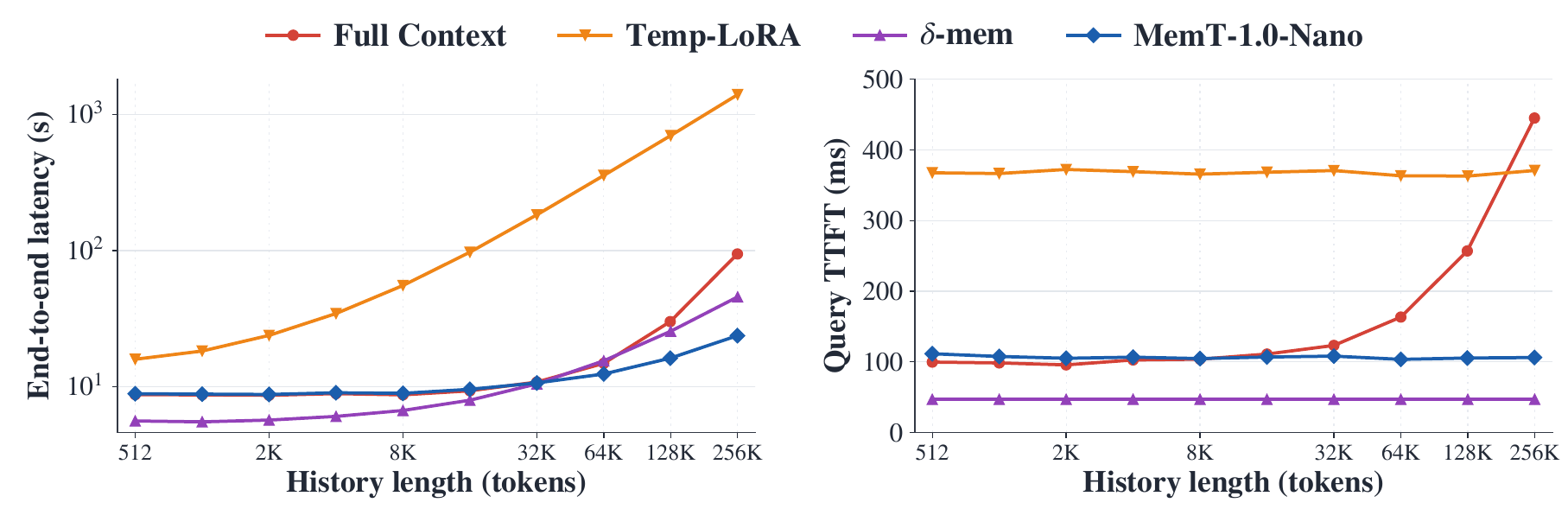}
    \vspace{-0.8cm}
    \caption{Left: End-to-end latency includes history preparation and generation of a 128-token response. Right: Query TTFT measures the time from query execution to the first generated token, excluding history preparation.}
    \label{fig:results_efficiency}
    \vspace{-0.4cm}
\end{figure}

We construct input sequences with varying history lengths based on BEAM.
The query uses shared 512-token content, with the chat template added separately, rather than a benchmark question, and each sequence is treated independently.
We evaluate a fixed generation length of 128 tokens by forcing the model to decode exactly 128 tokens while ignoring the end-of-sequence signal.
For the HF evaluations, synchronized wall-clock timers exclude model loading, input tokenization, and evaluation-related scoring overhead.
For each history and length, we measure preparation once, discard one query warm-up, and average five timed queries. Reported curves average the resulting values across 35 histories.
For MemT-1.0-Nano, history content is processed in 2,048-token slices, with each chat-formatted chunk capped at 2,048 tokens.

The reported end-to-end latency includes history preparation and response generation, accounting for indexing, test-time adaptation, or memory writes as applicable.
We additionally report query time to first token (TTFT), measured from the start of query execution to the first generated token, including query prefilling but excluding history preparation.
For full context and MemT-1.0-Nano, both end-to-end latency and TTFT include state restoration.
The HF full-context baseline reuses the prepared history cache.
We report results over the common history-length range from 512 to 256K tokens.
Figure~\ref{fig:results_efficiency} presents the results.
MemT-1.0-Nano achieves an end-to-end advantage over full context at long histories.
It is generally slower at short histories and approximately matches full context at 32K tokens, reaching speedups of approximately $1.20\times$ at 64K, $1.86\times$ at 128K, and $3.99\times$ at 256K.
Its memory storage procedure processes history incrementally; generation then accesses the resulting state without retaining historical tokens in the attention context.
In comparison, the full-context baseline incurs both history prefilling and continued attention over historical KV entries during generation.
The measured long-history end-to-end advantage primarily reflects lower history-preparation cost.

The query TTFT curves reveal an additional benefit after history preparation.
MemT-1.0-Nano maintains broadly stable mean query TTFT of approximately 104--112\,ms across the evaluated range, while full-context TTFT rises to approximately 257\,ms at 128K tokens and 445\,ms at 256K tokens.
MemT-1.0-Nano's behavior is consistent with a fixed-size memory readout whose state dimensions do not grow with the number of historical tokens processed.
For persistent interactions, it allows accumulated history to influence responses without progressively increasing the attention context used for each new query.
The benefit concerns access to the stored representation, and the memory-capability results above already show that this representation preserves useful information.
At 128K and 256K tokens, MemT-1.0-Nano achieves query TTFT speedups of approximately $2.44\times$ and $4.19\times$, respectively, over cached full-context inference.

Previously collected results for the parametric memory baselines highlight the importance of accounting for preparation costs.
Temp-LoRA maintains stable query TTFT, but its end-to-end cost grows substantially because history-specific adaptation requires optimization.
Although $\delta$-mem has the lowest query TTFT, its end-to-end latency exceeds Nano's from 64K tokens onward as history-preparation cost grows.
Absolute latency differences across these systems also reflect their backbone, runtime, and tokenization configurations.
These results show the value of efficient memory construction together with predictable query TTFT.
Reusing a prepared state amortizes its construction cost over multiple queries.

\subsection{Storage Overhead}
\label{sec:storage_overhead}
\begin{figure}[t]
    \centering
    \includegraphics[width=0.75\linewidth]{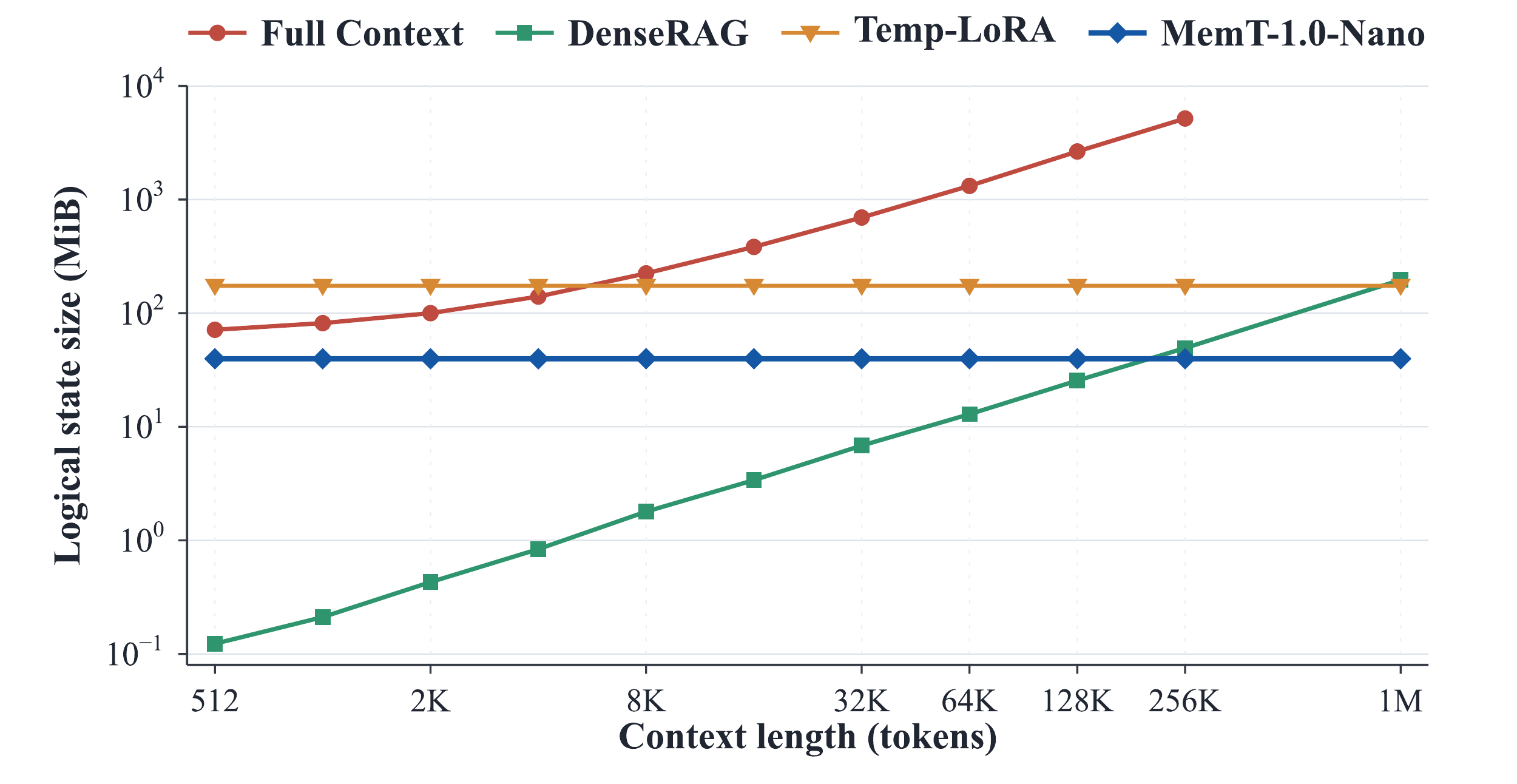}
    \vspace{-0.2cm}
    \caption{Results of storage overhead. Per-trajectory persistent storage on BEAM histories of increasing length.}
    \label{fig:storage_overhead}
    \vspace{-0.4cm}
\end{figure}

We evaluate the per-trajectory storage overhead on BEAM histories, after history preparation and before querying.
The reported storage size is the logical size in bytes of the per-trajectory state.
The shared backbone weights, embedders and hyper memory parameters are excluded from the measurement.
More details can be found in Appendix~\ref{appendix:eval_storage_overhead}, and the results are presented in Figure~\ref{fig:storage_overhead}.

Figure~\ref{fig:storage_overhead} shows a clear difference between retaining historical records and maintaining a fixed-size state.
Full context incurs increasing storage as its logical state size grows with the visible history.
DenseRAG has a much smaller storage overhead at short histories, but its stored texts and retrieval vectors also accumulate as more information is added.
In contrast, MemT-1.0-Nano's memory state storage remains constant throughout the trajectory, because new information updates memory states without enlarging them.
At a history length of 128K tokens, MemT-1.0-Nano achieves an approximately $66\times$ reduction in per-trajectory storage compared with the full context setting.
Temp-LoRA also exhibits constant storage, since its adapter dimensions are fixed, although its storage overhead is larger than MemT-1.0-Nano's in the evaluated configuration.

The comparison with DenseRAG illustrates how the storage advantage depends on interaction length.
For short histories, storing a small collection of text chunks and embeddings is cheaper than allocating MemT-1.0-Nano's full memory state.
As histories grow, DenseRAG's accumulated records eventually exceed that fixed allocation.
MemT-1.0-Nano therefore becomes more attractive for persistent sessions with extended interaction histories, where additional turns do not increase the size of the saved state.
The full context curve ends at its context-window limit, so its reported storage does not represent retaining the entire longest history.
For deployment of MemT-1.0, a constant state size makes per-trajectory logical-state budgets predictable, and bounds the logical payload that must be saved or transferred when resuming an interaction.
Total storage still grows with the number of independent trajectories and the configured memory capacity.

\subsection{Low-Rank Decomposition}
To further explore the compressibility of memory states based on MemT-1.0-Nano, we perform online low-rank projection for memory states to study their performance impacts.
Specifically, we apply singular value decomposition (SVD) to the FP32 state matrix of each layer after each storage procedure, and reconstruct a dense state from the retained components for subsequent computation.
For a retained rank $k$, the reconstructed state is computed as 
\begin{equation*}
    \mathbf{M}_{t,h}^{(l)}=U_{:,1:k}\Sigma_{1:k,1:k}V_{:,1:k}^{\top}.
\end{equation*}
We set the rank $k \in \{1, 2, 4, 8, 16, 32, 64, 128, 256\}$, with the full rank $k=256$ retaining the original dense state without projection.
The evaluation is conducted on PersonaMem and LoCoMo.
For each retention ratio, we independently initialize the memory state, write the complete history, and apply low-rank reconstruction after every memory storage  when $k<256$.

\begin{figure}[t]
    \centering
    \includegraphics[width=0.95\linewidth]{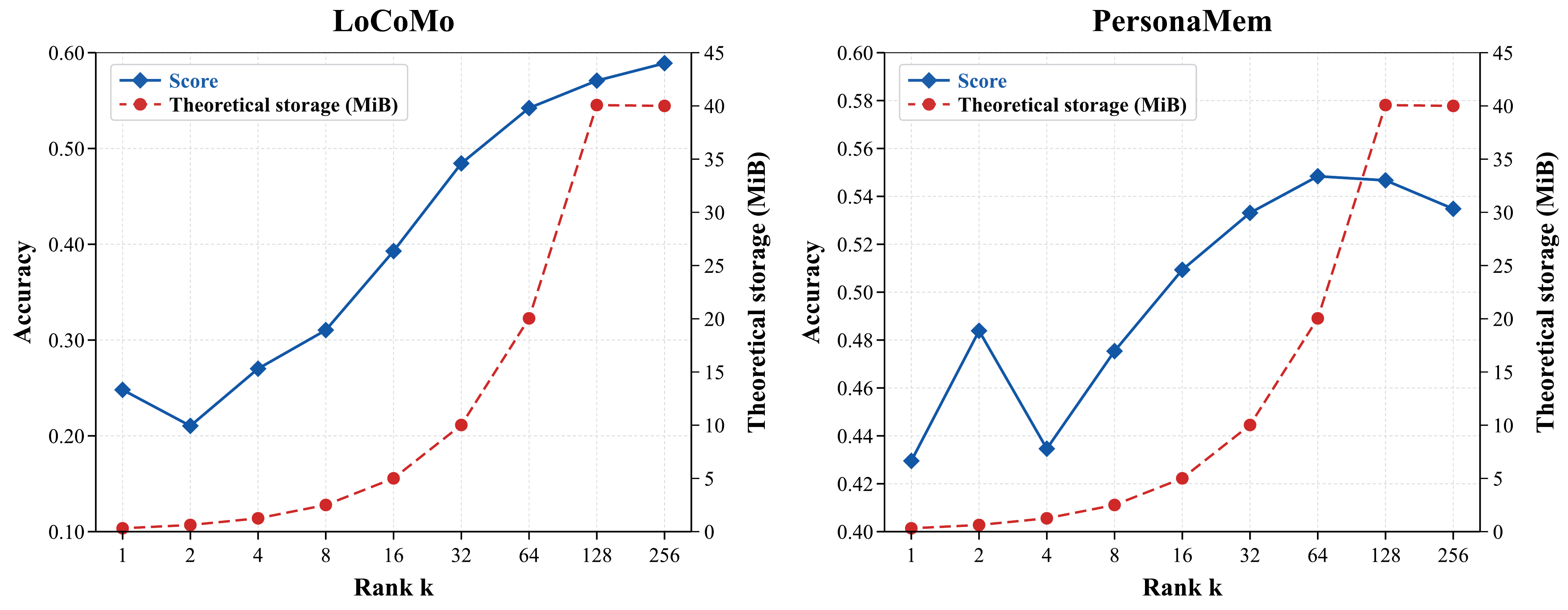}
    \vspace{-0.2cm}
    \caption{Impact of online low-rank decomposition
        on memory performance and storage.}
    \label{fig:low_rank_decomposition}
    \vspace{-0.4cm}
\end{figure}

Figure~\ref{fig:low_rank_decomposition} shows different results across the two tasks.
On LoCoMo, accuracy generally improves as more singular components are retained.
The largest gains fall in the middle of the rank range, and the final doublings add progressively less, although the score still improves up to the full rank.
The one exception is the smallest retained rank, which scores slightly above the next one.
At this level of truncation, the small score difference alone does not establish an advantage of extreme compression.
PersonaMem follows the same shape but saturates much earlier.
Its accuracy rises quickly and then flattens out before reaching full rank, with the highest observed accuracy at an intermediate rank and no further improvement once additional components are retained.
Overall, memory performance improves as the retained rank increases, but the returns diminish quickly, and on PersonaMem they eventually vanish.
Because theoretical factor storage grows roughly linearly with the retained rank, increasing the rank yields diminishing returns in downstream accuracy.

\section{Application in Agentic Tasks}
\label{sec:applications}

\subsection{Overview}

In this section, we investigate the potential of MemT-1.0 for developing self-evolving agents through continual learning. We apply MemT-1.0 to long-horizon agentic tasks, where agents must use environment feedback to guide task execution. Real-world agent interactions generate a wealth of trajectories containing valuable procedural experience. Reusing this experience offers a promising path toward improving agent performance across tasks. Current methods typically extract experience from historical trajectories and include it in the textual context of subsequent tasks~\cite{zhao2024expel}, increasing the burden on an already growing execution context. MemT-1.0 can instead store experience in native memory and use it to guide actions in long-horizon tasks.

We focus on agentic coding and train MemT-1.0-Nano to apply procedural experience extracted from historical trajectories. We evaluate it on SWE-bench Verified and SWE-bench Multilingual against Qwen3.6-35B-A3B, both without experience and with textual experience. With experience stored before execution, MemT-1.0-Nano improves the proportion of tasks solved in all three runs over the backbone without experience by 7.6 percentage points on SWE-bench Verified and 8.3 points on SWE-bench Multilingual. Results show that MemT-1.0 can improve agentic task performance beyond memory-focused evaluations and support its potential as a foundation for self-evolving agents.

\subsection{Experience-Guided Agentic Framework}
\label{sec:app-method}

\begin{figure*}[t]
    \centering
    \includegraphics[
        width=0.95\textwidth
    ]{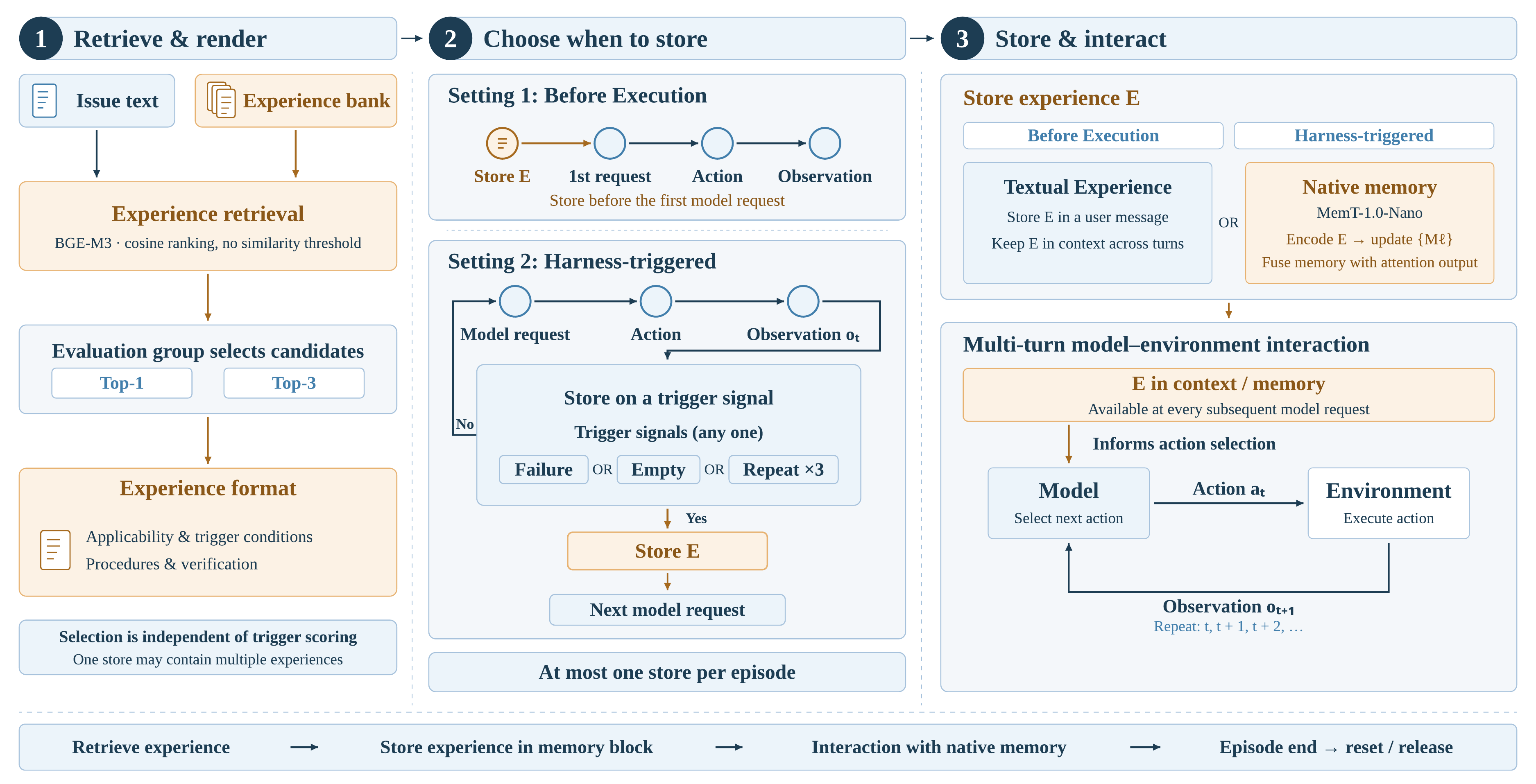}
    \caption{
        Experience-guided agentic framework with MemT-1.0-Nano. Retrieved experience is stored in textual context or native memory to guide subsequent actions.
    }
    \vspace{-0.4cm}
    \label{fig:app-harness}
\end{figure*}

To explore this potential, we develop an agentic framework in which MemT-1.0 stores and reuses procedural experience during long-horizon task execution.
We construct an experience bank by summarizing historical agent trajectories into reusable procedural guidance. As illustrated in Figure~\ref{fig:app-harness}, the framework retrieves relevant experience from the bank and stores it in either the agent's textual context or MemT-1.0-Nano's native memory. The two approaches use the same experience but differ in how it is accessed during task execution.

\textbf{Experience Retrieval and Storage.}
Given a coding task, we retrieve relevant experiences from the experience bank using BGE-M3~\cite{chen2024bgem3}. Each experience describes its applicable conditions, recommended procedures, and verification strategies. The task description serves as the retrieval query, and candidate experiences are ranked according to embedding similarity. The selected experience is converted into a structured textual representation before storage.

We consider two experience storage settings.
In the Before Execution setting, experience is stored before the agent begins execution. In the Harness-Triggered setting, the agent first executes the task independently, and experience is stored when the execution harness detects a critical breakpoint. Such breakpoints are identified using execution feedback, including command failures, empty observations, and repeated unsuccessful operations. The detection only relies on information available during execution, without accessing reference solutions. Each trajectory receives at most one experience store operation.

\textbf{Memory-Based Experience Application.}
In the Textual Experience setting, the retrieved experience is directly inserted into the agent's message history and remains available through the subsequent context. In the Native Memory setting, MemT-1.0-Nano writes the same experience into its native memory state through a separate memory store operation. Specifically, the experience text is processed by a forward pass with memory writing enabled, updating the memory state $M$ without modifying the original task context. After the memory store operation, agent actions access the stored experience through memory attention, allowing the experience to influence action generation without explicitly appearing in the message history.

Each execution trajectory maintains an independent memory state. The state is initialized to empty before the task and updated when experience is stored. After storing experience, it remains fixed and is read during subsequent execution until it is reset at the end of the trajectory. This setup evaluates whether a trained native-memory agent can effectively reuse procedural experience without retaining the experience text in the task context.

\subsection{Experience Construction and Model Training}
\label{sec:app-data-training}

To enable experience-guided agentic execution through native memory, we construct training data that associate reusable procedural experience with successful execution behaviors. The overall data construction consists of two stages, extracting reusable experience from historical trajectories and generating experience-conditioned trajectories.

\textbf{Experience Data Construction.}
We first construct an experience bank from 41,057 successful coding trajectories released by Scale-SWE~\cite{zhao2026scaleswe}. Following the experience summarization framework of MSCE~\cite{tang2026msce}, the construction process consists of trajectory annotation, procedural policy induction, policy clustering, and experience generation with quality auditing. For each successful trajectory, we identify important execution decisions and their corresponding environmental conditions. Trajectories with similar conditions are grouped to induce reusable procedural policies, which are subsequently consolidated and converted into structured experiences. Each experience contains information about its triggering conditions, recommended execution steps, applicability constraints, and verification procedures. After quality filtering, we obtain an experience bank containing 1,040 reusable coding experiences.

\textbf{Training Sample Construction.}
We further construct experience-guided training trajectories through counterfactual execution. Specifically, we generate original coding trajectories using Qwen3.6-35B-A3B with the AweAgent harness~\cite{aweagent2026} in the Scale-SWE environment. For unsuccessful trajectories, we identify candidate breakpoints according to execution errors and other observable failure signals. At each selected breakpoint, we retrieve candidate experiences and resume execution with different experience combinations. The resulting trajectories are evaluated using executable tests, and we retain experience-conditioned continuations that pass the task tests when the corresponding original trajectory failed.

Each training sample is represented as
\begin{equation*}
    x_{b,j} =
    \left(
        \xi_{\leq b},
        g_{b,j},
        \hat{\xi}_{>b}^{+}
    \right),
\end{equation*}
where $b$ indexes the selected breakpoint and $j$ indexes the candidate experience subset. Here, $\xi_{\leq b}$ denotes the original trajectory prefix before and including breakpoint $b$, $g_{b,j}$ denotes the injected experience subset, and $\hat{\xi}_{>b}^{+}$ denotes the successful continuation after experience injection.

The retained samples satisfy the condition that the original trajectory fails while the experience-conditioned continuation succeeds. These continuations provide supervision for predicting successful actions conditioned on procedural experience. We generate 18,820 original trajectories and evaluate 199,500 breakpoint-experience intervention variants. Among these interventions, 45,481 produce successful continuations. After additional quality filtering, we obtain 42,403 training samples. The resulting samples are partitioned by task to avoid leakage between training and validation trajectories. Appendix~\ref{appendix:detail_application} provides details of the experience bank, data construction pipeline, and filtering procedure.

\textbf{Training Strategy.}
We initialize the model from MemT-1.0-Nano. During application-specific post-training, all backbone parameters are frozen, and only the native hyper memory blocks are optimized. Each training sample is processed through separate experience and main-trajectory streams. First, the experience is written into native memory through a memory store operation. The original trajectory prefix is then replayed with memory reading disabled. After the designated breakpoint, memory reading is enabled, and the model is trained to predict the successful continuation conditioned on the stored experience. The experience text and trajectory prefix are excluded from the training loss, while the subsequent assistant actions provide the supervision signal.

We investigate both cross-entropy supervision and knowledge distillation. For knowledge distillation, the frozen backbone teacher receives the experience as text, while the student accesses the same experience only through native memory. Both models use teacher forcing on the same task history and reference continuation. We minimize the forward KL divergence from the teacher distribution to the student distribution over supervised assistant tokens.

We additionally investigate two supervision scopes.
The First-Step setting supervises only the first assistant turn following the breakpoint, and the Full-Suffix setting supervises the complete successful continuation. We use knowledge distillation with Full-Suffix supervision for the application-specific post-training of MemT-1.0-Nano and evaluate the resulting checkpoint in the main experiments. Appendix~\ref{appendix:detail_application} details the sample serialization and supervision masks.

\subsection{Agentic Coding Evaluation}
\label{sec:app-evaluation}

We evaluate MemT-1.0-Nano on two agentic coding benchmarks, including SWE-bench Verified~\cite{chowdhury2024swebenchverified}, containing 500 tasks, and SWE-bench Multilingual~\cite{Yang2025SWEsmithSD}, containing 300 tasks across nine programming languages. Both benchmarks require agents to resolve real-world software issues through repository exploration, code modifications, and executable tests. All baselines use Qwen3.6-35B-A3B as the backbone and follow the official SWE-bench verification protocol.

We compare Qwen3.6-35B-A3B (No Experience), Qwen3.6-35B-A3B (Textual Experience), and MemT-1.0-Nano (Native Memory). The \textbf{No Experience baseline} measures the benefit of experience reuse, while \textbf{Textual Experience} provides a reference for evaluating its effectiveness through native memory. The Harness-Triggered setting uses the same breakpoint-detection logic as training data construction. The Before Execution setting is used to assess robustness to experience injection timing. Each task is executed three times independently.

As shown in Table~\ref{tab:app-main-results}, we report the average success rate, \texttt{pass@3}, representing the proportion of tasks solved at least once across the three runs, and \texttt{pass\textasciicircum{}3}, representing the proportion of tasks solved in all three runs.
\textbf{MemT-1.0-Nano consistently improves the average success rate and repeated-run reliability over the backbone without experience.} On SWE-bench Verified, in the Before Execution setting, MemT-1.0-Nano achieves an average success rate of 69.33\%, improving upon the No Experience baseline of 66.07\%. This slightly exceeds the 68.87\% achieved by Textual Experience. Moreover, \texttt{pass\textasciicircum{}3} increases from 54.00\% to 61.60\%, matching the 61.60\% achieved by Textual Experience. These results indicate that \textbf{native memory can preserve the effectiveness of procedural experience without requiring the corresponding experience text to remain in the task context}.

\begin{table*}[t]
    \centering
    \caption{
        Agentic coding performance on SWE-bench Verified (500 tasks) and SWE-bench Multilingual (300 tasks) under two experience injection settings.
        Bold and underlined values indicate the highest and
        second-highest distinct scores, respectively.
    }
    \label{tab:app-main-results}
    \vspace{-0.1cm}

    \small
    \setlength{\tabcolsep}{5pt}
    \renewcommand{\arraystretch}{1.16}

    \sisetup{
        detect-weight = true,
        detect-family = true,
        round-mode = places,
        round-precision = 2,
        round-pad = true,
        table-alignment-mode = none,
        table-number-alignment = center,
        table-text-alignment = center
    }

    \newcommand{\appfirst}[1]{%
        \multicolumn{1}{c}{\textbf{#1}}%
    }
    \newcommand{\appsecond}[1]{%
        \multicolumn{1}{c}{\underline{#1}}%
    }

    \newcolumntype{Y}{
        S[
            table-format = 2.2,
            table-fixed-width = true,
            table-column-width = 5.5em,
            table-alignment-mode = none,
            table-number-alignment = center
        ]
    }

    \begin{adjustbox}{max width=\textwidth}
    \begin{threeparttable}

    \begin{tabular}{
        l
        l
        l
        Y Y Y
    }
        \toprule

        \rowcolor{MetisBlueDark}
        \color{white}\bfseries Benchmark
        & \color{white}\bfseries Store Timing
        & \color{white}\bfseries Method
        & \multicolumn{1}{c}{\color{white}\bfseries
            \mbox{pass\textasciicircum{}3}}
        & \multicolumn{1}{c}{\color{white}\bfseries Avg. Success}
        & \multicolumn{1}{c}{\color{white}\bfseries pass@3} \\

        \midrule


        \multirow{6}{*}{Verified}
        & \multirow{3}{*}{Before Execution}
        & Qwen3.6-35B-A3B (No Experience)
        & \appsecond{54.00}
        & 66.07
        & \appfirst{77.60} \\

        & & Qwen3.6-35B-A3B (Textual Experience)
        & \appfirst{61.60}
        & \appsecond{68.87}
        & 76.40 \\

        & & \textbf{MemT-1.0-Nano (Native Memory)}
        & \appfirst{61.60}
        & \appfirst{69.33}
        & \appsecond{77.40} \\

        \cmidrule(lr){2-6}

        & \multirow{3}{*}{Harness-Triggered}
        & Qwen3.6-35B-A3B (No Experience)
        & 54.00
        & 66.07
        & \appsecond{77.60} \\

        & & Qwen3.6-35B-A3B (Textual Experience)
        & \appsecond{61.80}
        & \appsecond{69.13}
        & 76.80 \\

        & & \textbf{MemT-1.0-Nano (Native Memory)}
        & \appfirst{62.80}
        & \appfirst{69.73}
        & \appfirst{77.80} \\

        \midrule


        \multirow{6}{*}{Multilingual}
        & \multirow{3}{*}{Before Execution}
        & Qwen3.6-35B-A3B (No Experience)
        & 45.33
        & 60.00
        & \appfirst{73.67} \\

        & & Qwen3.6-35B-A3B (Textual Experience)
        & \appfirst{56.33}
        & \appfirst{64.22}
        & \appsecond{72.67} \\

        & & \textbf{MemT-1.0-Nano (Native Memory)}
        & \appsecond{53.67}
        & \appsecond{62.89}
        & 72.00 \\

        \cmidrule(lr){2-6}

        & \multirow{3}{*}{Harness-Triggered}
        & Qwen3.6-35B-A3B (No Experience)
        & 45.33
        & 60.00
        & \appfirst{73.67} \\

        & & Qwen3.6-35B-A3B (Textual Experience)
        & \appsecond{53.67}
        & \appsecond{63.00}
        & 72.67 \\

        & & \textbf{MemT-1.0-Nano (Native Memory)}
        & \appfirst{55.33}
        & \appfirst{63.56}
        & \appsecond{73.00} \\

        \bottomrule
    \end{tabular}

    \begin{tablenotes}[flushleft]
        \footnotesize
        \item[]
        All values are percentages, based on three
        independent runs per task.
        The \mbox{pass\textasciicircum{}3} denotes the percentage of tasks solved in all three runs.
        The Avg. Success denotes the mean success rate across runs.
        The pass@3 denotes the percentage of tasks solved in at
        least one run.
        Textual Experience uses the top 3 experiences for Verified Before Execution; top 1 otherwise.
    \end{tablenotes}

    \end{threeparttable}
    \end{adjustbox}

    \vspace{-0.4cm}
\end{table*}

In the Harness-Triggered setting, MemT-1.0-Nano further improves the average success rate to 69.73\%, compared with 66.07\% for No Experience and 69.13\% for rank-1 textual experience. The corresponding \texttt{pass\textasciicircum{}3} reaches 62.80\%, representing an improvement of 8.80 percentage points over the backbone and 1.00 percentage point over fixed rank-1 textual experience. This suggests that native memory remains effective when experience is introduced dynamically during agent execution, rather than only at task initialization.

We observe consistent improvements on SWE-bench Multilingual. In the Before Execution setting, MemT-1.0-Nano improves the average success rate from 60.00\% to 62.89\% and \texttt{pass\textasciicircum{}3} from 45.33\% to 53.67\%. In the Harness-Triggered setting, the average success rate reaches 63.56\%, exceeding the fixed rank-1 textual experience result of 63.00\%. These findings provide further evidence that experience internalization generalizes across coding tasks involving different repositories and programming languages.

The gains are most pronounced in repeated-run reliability, suggesting that stored experience helps agents solve tasks more consistently. Although \texttt{pass\textasciicircum{}3} improves substantially, \texttt{pass@3} does not consistently increase.
The comparable behavior of MemT-1.0-Nano and textual experience further suggests that procedural guidance can be represented and utilized through native memory.

We also evaluate several training configurations on a 100-task development subset of SWE-bench Verified. Detailed comparisons and additional experimental results are provided in Appendix~\ref{appendix:detail_application}.

\subsection{Case Study and Discussion}
\label{sec:app-discussion}

To better understand how native memory affects agent behavior, we illustrate a representative task from SWE-bench Verified involving the two-digit year parsing logic in Django's \texttt{parse\_http\_date} function. 

We compare execution trajectories under the No Experience, Textual Experience, and Native Memory settings. Without experience, the agent modifies the relevant parsing logic shortly after locating the implementation, introducing targeted reproduction and verification steps later in the execution. In contrast, agents in both the Textual Experience and Native Memory settings construct reproduction tests before applying modifications. In the MemT-1.0-Nano trajectory, the agent further revisits a boundary condition after its initial modification, corrects the implementation, and performs additional verification.

These observations show how stored procedural experience can guide task execution without appearing in textual context. Together with the benchmark improvements, they demonstrate that MemT-1.0 can reuse procedural experience to improve agentic task performance. This supports its potential for developing self-evolving agents: \textbf{experience from past interactions can be stored in native memory and applied to new tasks.}

Several challenges remain. The current framework relies on externally constructed experience, making its effectiveness dependent on experience quality and retrieval. Storing experience also introduces computation, so reduced textual context does not necessarily imply lower overall execution costs. Extending experience reuse into an autonomous continual learning process remains an open direction.

Future work will explore the co-evolution of the harness and MemT-1.0 through continued interaction with the environment. The harness can improve experience acquisition and selection, while MemT-1.0 learns to reuse that experience more effectively. Together, these advances could support continual learning in self-evolving agents for long-horizon tasks.

\section{Limitations and Discussion}
Despite the progress of MemT-1.0 in native memory, it still has several limitations.
The fixed-size memory state imposes a capacity constraint.
For the fixed memory, MemT-1.0 maintains an $O(1)$ state with respect to the interaction history length $N$.
Compared with methods whose stored representations grow as $O(N)$, this bounded state has limited capacity to preserve detailed historical information.
As interactions accumulate, new information must be incorporated into an existing representation, creating trade-offs between retaining earlier details and adding recent information.
The resulting loss of memory fidelity remains a challenge for tasks requiring precise recall.
At the same time, a fixed-size state offers opportunities beyond reducing storage.
Each update incorporates new information into a representation, which has already been shaped by earlier interactions.
This enables successive compression of accumulated history.
Compared with maintaining a growing collection of historical records, this recurrent organization may provide a basis for latent reflection.
It potentially allows iterative consolidation of experience within the memory state.
Such a mechanism could support online self-evolution by allowing accumulated experience to influence subsequent behavior.

Finally, the relationship between local memory and hyper memory also deserves further study.
Local memory represents information accumulated during interaction, while hyper memory encodes learned capabilities for storing and utilizing that information.
An important question is whether experience accumulated in local memory can be consolidated into hyper-memory parameters through further training, and how knowledge encoded in hyper memory can be instantiated and adapted through local memory.
Therefore, memory capability itself can be viewed as a higher-order form of learned information.
Exploring this relationship may help connect adaptation within individual interactions with improvements in memory capability across tasks.

\section{Conclusion}

In this report, we presented MemT-1.0, a family of memory foundation models that integrates persistent memory into model computation.
Through the token-level stable memory updates, Mixture-of-Storage, a deliberate backbone preparation, and a two-stage training pipeline, MemT-1.0 supports efficient storage and utilization of historical information.
All three models outperform the evaluated parametric memory baselines on average, while largely preserving the general capabilities of their respective backbones under the noise setting.
Applications to coding agents further demonstrate that native memory can internalize procedural experience to improve task success and execution consistency.
Future work will focus on improving long-term memory, scaling memory capacity, and reducing the cost of memory updates.
We will explore autonomous experience acquisition and continual memory consolidation across tasks, moving toward agents learning through memory-native evolution.

\bibliographystyle{unsrt}
\phantomsection
\addcontentsline{toc}{section}{References}
\bibliography{reference}

\newpage

\appendix

\section{Author List and Acknowledgments}

\subsection{Author List}

\begingroup
\renewcommand{\thefootnote}{\ensuremath{\dagger}}

\textbf{Core Contributors:}
Zhiyu Li, Zeyu Zhang, Linyun Liu, Yiwei Bu, Chenyang Xi, Xichong Zhang, Yi Yu, Yu Yu, Zhichen Liu, Yuxiao Yi, Chunyu Li, Haoran Liu, Zhenkai Ding, Le Zhou, Juncheng Zhang, Boyu Chen, Zehao Lin, Guorui Li, Renyu Fu, Ji Li, Youyang Xia, Jiajun Shen, Harvey Li, Jinsong Zhou, Helan Hu, Tong Shen, Tianxiao Xu, Chenye Han, Weijie Li, Zhongquan Zhou, Hongming Xu, Yuekeng Li, Erenkai Zhu, Peng Yu, Wenhao Li, Jingyi Kang, Xixuan Hao, Xuzhou Zhu, Can Chen, Jiawei Liu, Chen Jiang, Tao Jiang, Fangwei Xiang, Zepeng Yao, Kai Chen, Guang Yang, Mingfei Lu, Chunhui Li, Zhengxuan Jiang, Bo Tang.

\textbf{Contributors:}
 Shiyu Zhan, Yihang Sun, Jie Chu, Wenjun Xiong, Zhenxiang Hua, Chen Lin, Detong Su, Dongyu Deng, Qiuping Ding, Guofeng Ding, Si Chen, Guohua Wen.

\textbf{Co-advisors:}
Xu Chen, Zhi-Qin John Xu, Ying-Cong Chen, Ningyu Zhang, Haofen Wang, Muning Wen, Haotian Chen, Wentao Zhang, Xuanhe Zhou, Rita Zhang, Junchi Yan, Feiyu Xiong.

\endgroup

\subsection{Acknowledgments}
We thank NVIDIA for its research collaboration with MemTensor on MemT-1.0-Pro. In addition, some of the training experiments for MemT-1.0-Flash and MemT-1.0-Nano were conducted on NVIDIA’s platform.
Built on NVIDIA-Nemotron-3-Ultra-550B-A55B-BF16, MemT-1.0-Pro incorporates MemTensor’s memory-native modules into 12 of the Nemotron backbone’s full-attention layers while retaining the backbone’s Mamba and MoE components. We are grateful to our NVIDIA collaborators for their valuable technical contributions and insightful discussions throughout the model development process. This joint effort extends memory-native capabilities to a large-scale foundation model.
We thank Shanghai Kupas Technology Co.,Ltd. and INESA (Group) Co., Ltd. for their support.
This work was also supported by the Artificial Intelligence Subsidy Scheme of the Government of the Hong Kong Special Administrative Region.
We also thank Zhengren Wang, Yushen Wang, Yida Xue, Nuoxi Chen, Haoyu Liu, Chenxu Du, and Yinghan Zhao for their valuable contributions to this project.

\section{Adaptation to Specific Backbones}
\label{appendix:backbone_adaptation}

Table~\ref{tab:backbone_memory_geometry} summarizes the memory geometry of the MemT-1.0-Nano, MemT-1.0-Flash, and MemT-1.0-Pro models used in the main evaluations.
All three use one memory block per selected layer ($E=1$).
The four-expert MoS workload in Appendix~\ref{appendix:infra_experiment_details} measures the cost of capacity expansion separately.
Layer indices below start at zero.

\begin{table}[htbp]
    \centering
    \caption{Backbone and native-memory configuration. State size counts the persistent memory matrices for one trajectory, excluding model weights and current-step KV caches.}
    \label{tab:backbone_memory_geometry}
    \small
    \renewcommand{\arraystretch}{1.15}
    \begin{tabularx}{\linewidth}{
        >{\raggedright\arraybackslash}p{0.22\linewidth}
        *{3}{>{\raggedright\arraybackslash}X}
    }
        \toprule
        \textbf{Configuration}
        & \textbf{MemT-1.0-Nano}
        & \textbf{MemT-1.0-Flash}
        & \textbf{MemT-1.0-Pro} \\
        \midrule
        Backbone
        & Qwen3.6-35B-A3B
        & DeepSeek-V4-Flash, after backbone preparation
        & Nemotron-3-Ultra-550B-A55B \\
        Hidden width / decoder layers
        & 2048 / 40
        & 4096 / 43
        & 8192 / 108 \\
        Memory layers
        & $3,7,\ldots,39$ (10 layers)
        & $4,6,\ldots,42$ (20 layers)
        & $7,14,23,32,$ $39,48,57,64,$ $73,82,89,98$ (12 layers) \\
        Memory-layer type
        & Full attention
        & Compressed sparse attention
        & Full attention \\
        Memory $H_q/H_{kv}$
        & 16 / 16
        & 64 / 64
        & 64 / 64 \\
        Memory $d_k/d_v$
        & 256 / 256
        & 512 / 512
        & 128 / 128 \\
        Memory experts per layer
        & 1
        & 1
        & 1 \\
        Trainable memory parameters
        & 450,931,360
        & 3,608,608,000
        & 4,227,959,040 \\
        State elements per trajectory
        & 10,485,760
        & 335,544,320
        & 12,582,912 \\
        FP32 state size
        & 40 MiB
        & 1280 MiB
        & 48 MiB \\
        Readout / fusion-point width
        & 4096 / 4096
        & 32768 / 8192
        & 8192 / 8192 \\
        \bottomrule
    \end{tabularx}
\end{table}

\subsection{Projection and Fusion}

\textbf{MemT-1.0-Nano.}
MemT-1.0-Nano augments each full-attention layer of Qwen3.6-35B-A3B with a native memory block, preserving the original hybrid architecture and MoE computation.
The memory branch receives the same pre-normalized hidden states as the original attention and assigns an independent memory channel to each query head.
Memory query, key, and value projections are dense and initialized from the corresponding backbone attention projections.
The query projection copies the query component of the backbone's joint query/output-gate projection.
The two backbone KV heads are each copied into eight memory heads, giving 16 independently trainable memory heads.
The copies share initial values but are separate parameters, with separate channel-wise gates and memory states.
After per-head RMSNorm and output gating, the 16 readouts are concatenated into a 4096-dimension vector, added to the backbone's already-gated attention output, and passed through its frozen $4096\!\times\!2048$ output projection.
The backbone sigmoid gate therefore applies only to the original attention branch, while memory uses the output gate defined in Section~\ref{sec:model_arc}.

\textbf{MemT-1.0-Flash.}
MemT-1.0-Flash attaches a native memory block to the compressed-attention layers of DeepSeek-V4.
The backbone compressed-attention has 64 query heads of dimension 512, and the memory branch keeps the same geometry, assigning one independent memory channel to each query head.

The memory query projection follows the backbone query projection, which is a two-stage bottleneck that maps the hidden state of width 4096 to a rank-1024 factor and then to 32768 features.
Both stages and the normalization gain on the intermediate factor are copied from the backbone attention.
This normalization depends on the input and therefore remains in the projection rather than being folded into the weights.

Keys and values instead use a grouped bottleneck.
Eight latent groups of width 512 are projected from the hidden state, and each group is normalized by its own RMSNorm over the group's 512 features.
The group is then expanded to eight heads of dimension 512, so that the eight groups yield 64 memory heads in total, matching the backbone query-head count.

The eight groups are initialized from three backbone projections: the layer's local KV projection, written $L$, and the previous and current halves of the compressed-attention projection, written $P$ and $C$.
Group $g$ draws its key and value from the $g$-th entry of the following list of pairings:
\begin{equation*}
    (L,L),\ (P,C),\ (P,L),\ (C,P),\ (C,L),\ (L,P),\ (L,C),\ (C,C).
\end{equation*}
Here $P$ and $C$ denote the two halves of a single backbone projection, and a pairing such as $(P,C)$ means that the key is taken from the previous half and the value from the current half.
Each group also copies the normalization gain and epsilon of the backbone module that its pairing selects, so groups drawing from $L$ inherit the local KV normalization while those drawing from $P$ or $C$ inherit the compressed-attention normalization.
Within a group, the expansion to the eight heads is initialized as an identity mapping and the heads are trained independently.

After readout, the backbone's frozen grouped output projection maps eight groups of 4096 features to eight groups of 1024.
The resulting 8192-dimensional memory contribution is added at the input of the final backbone output projection, which contracts it to the hidden width of 4096.
The original backbone residual and hyper-connection operations are preserved.

\textbf{MemT-1.0-Pro.} MemT-1.0-Pro attaches native memory to all 12 full-attention layers of the Nemotron backbone.
The backbone has a hidden width of 8192, 64 query heads, and two key-value heads, with a head dimension of 128. Each selected layer uses one memory block with 64 independent memory channels and $d_k=d_v=128$.
The memory branch receives the same pre-normalized hidden states as the original attention.
Its dense query projection is copied from the backbone, while each backbone key and value head is copied into 32 independently trainable memory heads. These copies share initial values but retain separate parameters and memory states.
After per-head RMSNorm and output gating, the memory readouts are concatenated into an 8192-dimensional vector and added to the original attention output before the backbone's frozen output projection.
The original layer normalization, residual connections, and Mamba/MoE modules are retained.

\subsection{Gates and Initialization}

All gates take the pre-attention hidden state directly.
MemT-1.0-Nano uses dense erase/write projections and rank-256 factorizations for the forget and output gates. MemT-1.0-Flash uses rank-512 factorizations for all four gates.
MemT-1.0-Pro uses dense erase/write projections and rank-512 factorizations for the forget and output gates.
There is no intervening nonlinearity or normalization between gate factors, so the erase/write factors compose into the effective matrices in Section~\ref{sec:model_arc}.
MemT-1.0-Flash's Q/K/V bottleneck normalization, described above, is distinct from these gate projections.

New gate input weights are initialized from $\mathcal{N}(0,0.02^2)$.
For all three models, the forget-gate expansion starts at zero and the output-gate expansion uses $\mathcal{N}(0,0.001^2)$.
MemT-1.0-Nano copies the initial erase projection into the write projection, then trains them independently.
MemT-1.0-Flash initializes the erase/write expansion factors independently from $\mathcal{N}(0,0.02^2)$ and initializes the output-gate bias to $0.1$.
MemT-1.0-Pro initializes its dense erase/write projections independently from $\mathcal{N}(0,0.02^2)$.
MemT-1.0-Nano and MemT-1.0-Pro have no output-gate bias.
All three models use a SoftCapSiLU threshold of $2.0$.
For MemT-1.0-Pro, no soft cap is applied to the memory values.

Forget-gate biases encode a range of initial time scales.
For each key channel, a half-life $\tau_{1/2}$ is sampled independently and log-uniformly from $[2048,32768]$ tokens for MemT-1.0-Nano and $[128,32768]$ tokens for MemT-1.0-Flash and MemT-1.0-Pro.
We initialize
\begin{equation*}
    b_\lambda=\operatorname{Softplus}^{-1}\!\left(\frac{\log 2}{\tau_{1/2}}\right),
    \qquad s_h=0.
\end{equation*}
Together with the zero forget-gate expansion, this sets the initial decay time scale through a factor $2^{-1/\tau_{1/2}}$ per token.
Both the channel-wise bias $b_\lambda$ and head-wise log-scale $s_h$ remain trainable and are excluded from weight decay.
Memory states start at zero for each independent trajectory.

\subsection{Training Configuration}
\label{appendix:memory_training_configuration}

The trainable parameter counts in Table~\ref{tab:backbone_memory_geometry} include memory Q/K/V projections, gate projections and biases, decay scales, and memory normalization gains.
Backbone parameters, including the output projections used for fusion, are frozen during memory training.
Model parameters and projection computation use BF16.
Optimizer master parameters, gradient accumulation/reduction, and Adam moments use FP32.
Memory-state accumulation and the sensitive normalization/decay calculations also use FP32.

The single-GPU MoS measurements in Appendix~\ref{appendix:infra_experiment_details} have their own workload and software configuration.

\section{Stable Mechanism on GDN-2}
\label{appendix:stable_gdn2}

\subsection{Problem Formulation}

We consider a general recursive update of the form
\begin{equation}
S_t = P_t S_{t-1} + B_t.
\end{equation}
Define the propagation matrix $\Phi_{t:i} = P_t P_{t-1}\cdots P_i$. Expanding the recursion yields
\begin{equation}
S_t = \Phi_{t:1} S_0 + \sum_{i=1}^{t} \Phi_{t:i+1} B_i.
\end{equation}
Setting $S_0=0$, we obtain
\begin{equation}
S_t = \sum_{i=1}^{t} \Phi_{t:i+1} B_i.
\end{equation}
Assuming $\lVert P_t\rVert \le \rho$ and $\lVert B_t\rVert \le R$, an upper bound of $\lVert S_t\rVert$ is
\begin{equation}
\lVert S_t\rVert \le R\sum_{j=0}^{t-1} \rho^j,
\end{equation}
that is,
\begin{equation}
\lVert S_t\rVert
\le
\begin{cases}
\displaystyle R\frac{1-\rho^t}{1-\rho}, & 0\le \rho<1,\\[0.8em]
\displaystyle tR, & \rho=1,\\[0.6em]
\displaystyle R\frac{\rho^t-1}{\rho-1}, & \rho>1.
\end{cases}
\end{equation}
Therefore, when $\rho<1$, we have
$\lVert S_t\rVert \le \frac{R}{1-\rho}$ as $t\to\infty$, and the update is bounded; we call such a recursion a \emph{stable update}.

Next, we analyze the vector-form GDN-2 update:
\begin{equation}
S_t = \bigl(I - k_t^\top (b_t \odot k_t)\bigr) D_t S_{t-1} + k_t^\top (w_t \odot v_t),
\end{equation}
where $D_t=\mathrm{Diag}(d_t)$ is diagonal, $d_t,b_t,w_t\in(0,1)$, and $\lVert k_t\rVert_2=1,\lVert v_t\rVert_2\leq g$, $g$ is a boundary of $v$'s norm.

This update is not stable in general. Let
$A_t = I - k_t^\top (b_t \odot k_t)$, $P_t=A_t D_t$, and $B_t=k_t^\top (w_t \odot v_t)$.

We omit the time index and define
\begin{equation}
a=b\odot k,\quad \rho = ka^\top,\quad m=\lVert a\rVert_2^2.
\end{equation}
Then
\begin{equation}
\rho=\sum_i b_i k_i^2,\qquad m=\sum_i b_i^2 k_i^2,
\end{equation}
and $\rho\in(0,1)$, $m\in[\rho^2,\rho)$.

Decompose $a$ into the component parallel to $k$ and the orthogonal component. Let $e$ be a unit vector such that $ek^\top = 0$ and $\lVert e\rVert_2=1$. Define $\gamma^2=m-\rho^2$, then
\begin{equation}
a = \rho k + \gamma e.
\end{equation}
Using the basis $\{k,e\}$, we have
\begin{equation}
A = I - k^\top a
= \begin{bmatrix}
1-\rho & -\gamma\\
0 & 1
\end{bmatrix}\oplus I_{K-2}.
\end{equation}
Hence
\begin{equation}
eA^\top = e - \gamma k,\qquad \lVert eA^\top\rVert_2^2 = 1+\gamma^2 \geq 1.
\end{equation}
This shows that along the $e$ direction, whenever $\gamma\neq 0$, the update amplifies energy, and $A$ may lead to an unstable recursion. Moreover,
\begin{equation}
\gamma=0 \iff a \parallel k.
\end{equation}
Thus, if we only consider $A$, a prerequisite for stability is that $b\odot k$ must be colinear with $k$.

We further analyze $A$ by focusing on its $2\times 2$ non-trivial block $A_2$ and considering $A_2^\top A_2$. Its trace is $T=2-2\rho+m$ and its determinant is $\det(A_2^\top A_2)=(1-\rho)^2$. The squared spectral norm (i.e., the largest singular value squared) is
\begin{equation}
\lVert A_2\rVert_2^2 = \lambda_+ = \frac{T+\sqrt{T^2-4(1-\rho)^2}}{2} \ge 1.
\end{equation}
In particular, the spectral norm squared of the full $A_2$ block is
\begin{equation}
\lVert A\rVert_2^2 = \max\{1,\lambda_+\}=\lVert A_2\rVert^2.
\end{equation}
Since $m<\rho$, we have
\begin{equation}
\lambda_+ < \frac{2-\rho+\sqrt{\rho(4-3\rho)}}{2}.
\end{equation}
The right-hand side achieves its maximum $\frac{4}{3}$ at $\rho=\frac{1}{3}$, hence
\begin{equation}
1 \le \lVert A\rVert_2 < \frac{2}{\sqrt{3}}.
\end{equation}
This gives an upper bound on the spectral norm of $A$. When both $A$ and $D$ are considered in the update, if $\lVert D\rVert_2\le \overline{d}$, then
\begin{equation}
\lVert P\rVert_2 = \lVert AD\rVert_2 \le \frac{2}{\sqrt{3}}\,\overline{d}.
\end{equation}
To ensure uniform contraction bound for $P_t$ less than 1, a sufficient condition is $\overline{d} < \frac{\sqrt{3}}{2}\approx 0.866$. Since $P_t S_{t-1}$ does not always amplify, $\overline{d}<0.866$ is a sufficient condition for stability in the worst case.
This condition guaranties uniform contraction of the homogeneous recursion. If the additive inputs $B_t=k_t^\top(w_t\odot v_t)$ are uniformly bounded, then the resulting inhomogeneous recursion is uniformly bounded as well.
Under this condition, the half-life of the forget gate $D$,
\begin{equation}
t_{1/2}=\frac{\ln 2}{-\ln d}=\frac{\ln 2}{-\ln\sqrt{3}/{2}}\approx 4.82,
\end{equation}
is only about 5 tokens.

\subsection{Solution: Symmetrization Fix}

Note that in the GDN-2 update, the $A$ term can be written as
\begin{equation}
A = I - k^\top (b\odot k) = I - k^\top k\,\mathrm{Diag}(b).
\end{equation}
The instability stems from the fact that $A$ is not symmetric. We therefore move $\mathrm{Diag}(b)$ to both $k$ to make $A$ symmetric:
\begin{equation}
A = I - \text{Diag}(\sqrt{b})\,k^\top k\,\text{Diag}(\sqrt{b}).
\end{equation}
Further, we can rewrite it as
\begin{equation}
A = I - (\sqrt{b}\odot k)^\top(\sqrt{b}\odot k).
\end{equation}
With this symmetrization fix, we obtain Stable GDN-2 (SGDN-2), whose update is
\begin{equation}
S_t = \bigl(I-(\sqrt{b_t}\odot k_t)^\top(\sqrt{b_t}\odot k_t)\bigr) D_t S_{t-1}
+ (\sqrt{b_t}\odot k_t)^\top(w_t\odot v_t).
\end{equation}
According to the analysis above, the two vectors in SGDN-2 are naturally colinear (i.e., $\gamma=0$), which avoids the unstable update mechanism.

Geometrically, the symmetrized erase step acts separately
on the component parallel to $u=\sqrt{b}\odot k$ and its
orthogonal complement, with no coupling between these two
orthogonal subspaces. Considering the erase step alone,
with $D=I$ and no write term, the parallel component is
scaled by $1-\lVert u\rVert_2^2$, while the orthogonal
component remains unchanged. Under the stated assumptions,
$0<\lVert u\rVert_2^2<1$; hence, the map contracts along $u$
without expanding any orthogonal direction and is
non-expansive in the Euclidean norm.

\begin{figure}
    \centering
    \includegraphics[width=0.8\linewidth]{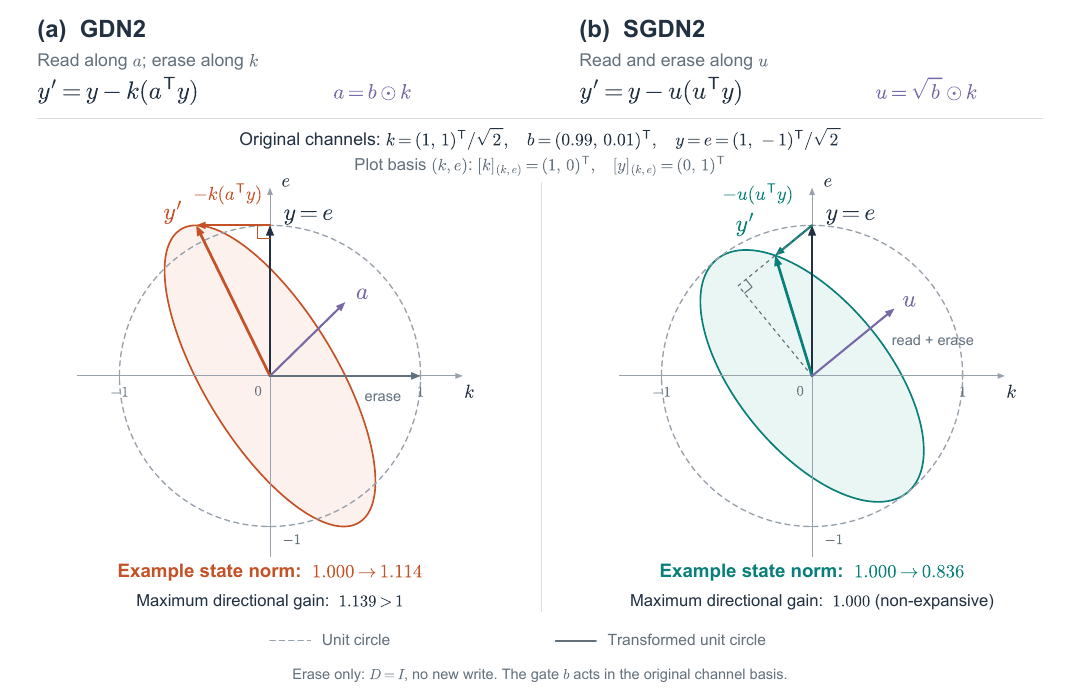}
    \caption{An analytical single-step erase example with $D=I$ and no write term.
    Both panels use the same input $y=e$ and orthonormal $\{k,e\}$ basis.}
    \label{fig:stable_gdn2_geometry}
\end{figure}

In contrast, GDN-2 computes the erase magnitude using
$a=b\odot k$ but applies the correction along $k$.
For the input $y=e$, this gives $y'=e-\gamma k$.
Since $e$ and $k$ are orthogonal, the correction increases
the squared norm from $1$ to $1+\gamma^2$ whenever
$\gamma\neq0$.
Figure~\ref{fig:stable_gdn2_geometry} illustrates the resulting
non-expansive map and contrasts it with a GDN-2 erase step
that increases the norm of the same input $y=e$.

\section{Details of Evaluation Setup}

\subsection{Details of Benchmarks}
\label{appendix:benchmark_details}

\textbf{Memory Benchmark.}
We evaluate memory capabilities on 11 benchmarks and divide them into four groups according to trajectory length as follows.
These benchmarks focus on long-term information retention and retrieval, dynamic memory updating and management, and personalized understanding.\\
$\bullet$ \textbf{MemBench}~\cite{tan2025membench}: MemBench reports the equal-weight mean accuracy across its factual/reflective and participation/observation quadrants on the 0--10K subset. We write each history in an independent scope and preserve its original message order.
We note that when structured time is unavailable, no synthetic date is constructed.\\
$\bullet$ \textbf{ImplicitMemBench}~\cite{qin2026implicitmembench}: ImplicitMemBench evaluates procedural learning, priming, and classical conditioning through learning or priming, interference, and test stages.
We preserve the original trajectory order and keep the priming treatment and control
groups separate. We report the equal-weight mean of First-Try Accuracy (FTA) for procedural
learning, FTA for classical conditioning, and the Priming Influence Score (PIS).
\\
$\bullet$ \textbf{PersonaMem}~\cite{jiang2025know}: PersonaMem evaluates dynamic user profiling and personalized responses over evolving interactions.
Each (history, history\_end) pair receives an independent scope.
Questions with the same cutoff share a read-only state, whereas different cutoffs do not share memory.
We use the 32K subset of PersonaMem and report answer-option accuracy. \\
$\bullet$ \textbf{PS-Bench}~\cite{guo2026personalization}: PS-Bench measures whether personal memories induce intent legitimation and increase personalized attack success.
Persona histories follow the original session, speaker, and timestamp order.
On the Persona-grounded subset, we report \(100(1 - \mathrm{ASR})\), where the attack
success rate (ASR) is micro-averaged over 1,863 queries.
\\
$\bullet$ \textbf{LoCoMo}~\cite{maharana2024evaluating}: LoCoMo evaluates single-hop, multi-hop, temporal, open-domain, and adversarial conversational memory.
We preserve session, message, speaker, and date order.
We also use speaker-specific scopes for MemOS, Mem0, and Graphiti, and use a conversation scope for OpenViking.
We exclude the adversarial category and report overall LLM-judged answer accuracy across the 1,540 included questions.\\
$\bullet$ \textbf{PerMemSafe}~\cite{an2026permemsafe}: PerMemSafe evaluates implicit personalized safety through Safety Perception (SP) and Dynamic Evolution (DE) tracks.
SP and DE histories use independent scopes, while risk descriptions, isolated answers, and scoring rubrics are excluded from memory writes.
We use the conv64 subset and report overall Personalized Safety Rate (PSR)
across 750 questions.\\
$\bullet$ \textbf{Memora}~\cite{uddin2026recall}: Memora evaluates remembering, reasoning, and recommendation under evolving weekly, monthly, and quarterly interactions.
Each persona uses an independent scope with chronological session and turn writes and timestamps.
Query dates are not used to filter history.
We use the Weekly subset and report mean Forgetting-Aware Memory Accuracy (FAMA)
across 150 questions, equivalent to the equal-weight mean across three categories of 50
questions each.
\\
$\bullet$ \textbf{MemOps}~\cite{hao2026memops}: MemOps evaluates remembering, updating, forgetting, reflecting, and composed memory operations. We store 50 sessions by session index and message order in an independent scope, excluding operation labels, state annotations, and provenance tags from memory.
All methods use the composite score
\[
S_{\mathrm{MO}}=\frac{A+F_{\mathrm{op}}+P_{\mathrm{prov}}+(100-L)+(100-V)+P_{\mathrm{ref}}}{6},
\]
where $A$, $F_{\mathrm{op}}$, $P_{\mathrm{prov}}$, $L$, $V$, and $P_{\mathrm{ref}}$ denote Answer Accuracy, Operation F1, Provenance Support, Leakage, Stale Value, and Reflect Precision, respectively, all on a 0--100 scale. \\
$\bullet$ \textbf{LongMemEval}~\cite{wu2024longmemeval}: LongMemEval evaluates information extraction, multi-session reasoning, temporal reasoning, knowledge updating, and abstention.
Each haystack uses an independent scope and session-date order, and the current date appears only in the query and is not stored in memory.
We use LongMemEval-S and report LLM-judged accuracy across all 500 questions.\\
$\bullet$ \textbf{MemoryAgentBench}~\cite{hu2026evaluating}: MemoryAgentBench evaluates retrieval, test-time learning, long-range understanding, and selective forgetting.
Source histories are isolated and written in complete chunk-N order without source mixing.
We use the Strict scores for the main set, excluding short-context ablations, and report
the equal-weight macro-average of AR, TTL, LRU, and SF.
\\
$\bullet$ \textbf{BEAM}~\cite{tavakoli2025beyond}:
BEAM evaluates ten long-context memory abilities, including extraction, contradiction resolution, updating, preference following, summarization, and temporal reasoning.
Each conversation uses an independent scope and complete chronological writes by session, timestamp, and message.
We use BEAM-1M, with 70 questions per ability. For non-ordering tasks, each question is scored by averaging its LLM-judged rubric scores.
Event Ordering uses the official normalized Kendall $\tau_b$ score $\tau_{\mathrm{norm}}=(\tau_b+1)/2$, computed from matched reference-event positions in the answer when at least two events match; zero or one match scores 0 or 1, respectively. No event-coverage or F1 multiplier is applied.
We average question scores within each ability, convert to a 0--100 scale, and report the equal-weight macro-average of the ten abilities. Full Context follows the window limits in Appendix~\ref{appendix:baseline_details}.

The overall memory score is the equal-weight arithmetic mean of the 11 unrounded benchmark scores.

\textbf{General Benchmark.}
We evaluate general capabilities on six benchmarks spanning four major dimensions: knowledge and scientific reasoning, instruction following, long-context understanding, and code generation.
The detailed benchmarks are presented as follows.

$\bullet$ \textbf{MMLU-Pro}~\cite{wang2024mmlu}: MMLU-Pro extends MMLU with harder reasoning questions and ten answer choices. We use the \texttt{TIGER-Lab/MMLU-Pro@14be195} test split and evaluate these conditions with fixed demonstrations provided through ordinary context or native memory. \\
$\bullet$ \textbf{GPQA Diamond}~\cite{rein2023gpqa}: GPQA Diamond contains graduate-level biology, physics, and chemistry questions designed to resist ordinary web search.
We evaluate each of the 198 questions eight times, keeping its answer-option order fixed across repetitions and evaluation conditions, and report mean accuracy. \\
$\bullet$ \textbf{IFEval}~\cite{zhou2023instruction}: IFEval evaluates programmatically verifiable constraints such as length, keywords, casing, and formatting. We use the 541-prompt, 834-instruction test split and provide fixed irrelevant demonstrations through ordinary context or native memory. \\
$\bullet$ \textbf{LongBench v2}~\cite{bai2025longbench}: LongBench v2 evaluates deep reasoning across six long-context task categories and contexts up to 2M words.
We use the 503-item train split and provide identical irrelevant demonstrations through ordinary context or native memory. \\
$\bullet$ \textbf{BigCodeBench-Hard}~\cite{zhuo2025bigcodebench}: BigCodeBench-Hard contains 148 practical programming tasks with complex instructions and function calls.
We use normalized \texttt{v0.1.4} test data, execute generated programs with the official tests, and report Pass@1 under the evaluated configurations.\\
$\bullet$ \textbf{LiveCodeBench v6}~\cite{jain2025livecodebench}: LiveCodeBench is a continuously refreshed coding benchmark based on recent LeetCode, AtCoder, and Codeforces problems.
We use the 1,055-item cumulative \texttt{release\_v6} split and preserve identical demonstrations, generation settings, and scoring protocols across conditions.

\subsection{Details of Baselines}
\label{appendix:baseline_details}

We compare MemT-1.0 with textual and parametric memory baselines. Within each benchmark
and backbone, we hold the task requirements fixed while varying how the methods make the
history available to the model.

$\bullet$ \textbf{Full Context}: Full context prepends the available history to the query as a context-only reference. Full Context (256K) uses the default context budget specified in Appendix~D.3.2.
Full Context (Max) uses a 1,048,576-token window for NV and DS and a
1,010,000-token window for QW (YaRN expansion factor of 4). Histories exceeding the joint budget are left-truncated to retain the most recent suffix. The configurations differ only in the windows used for MemoryAgentBench and BEAM; the other nine benchmarks require no truncation at the default window. \\
$\bullet$ \textbf{DenseRAG}: DenseRAG retrieves the 10 most relevant historical segments and prepends them to the query.
In our implementation, the retrieval process is scope-isolated, and LoCoMo segments are indexed as \texttt{Speaker said, ``text''}, and answers are not written back. \\
$\bullet$ \textbf{MemOS}~\cite{li2025memos}: MemOS provides structured extraction, synchronization, and hybrid memory retrieval. We use \texttt{version 2.0.32} with \texttt{multimodal\_struct}, \texttt{fine}, and \texttt{sync}, top-20 fast search, MMR deduplication, and reranking. \\
$\bullet$ \textbf{Mem0}~\cite{chhikara2025mem0}: Mem0 performs LLM-based memory extraction, updating, and semantic retrieval. We use \texttt{version 2.0.20} with \texttt{infer=true}, dataset-specific protective chunking, top-20 user-scoped retrieval, and reranking disabled.\\
$\bullet$ \textbf{OpenViking}~\cite{fu2026vikingmem}: OpenViking organizes session-level memories hierarchically for retrieval.
We use \texttt{version 0.4.17.1} with a 32,768-token submission budget and retain the top 20 user-scoped contexts without additional reranking.\\
$\bullet$ \textbf{Graphiti}~\cite{rasmussen2025zep}: Graphiti represents memory as a temporal knowledge graph. We use \texttt{version 0.30.1}, dataset-specific Episode boundaries, and EDGE hybrid RRF to retrieve the top-20 facts within the current group. \\
$\bullet$ \textbf{Temp-LoRA}~\cite{wang2024greater}: Temp-LoRA adapts model parameters separately for each history.
We train a rank-64 temporary LoRA on non-overlapping 1,024-token chunks for two epochs per chunk.
We use Qwen3.6-35B-A3B as its backbone.\\
$\bullet$ \textbf{Doc-to-LoRA}~\cite{charakorn2026doc}: Doc-to-LoRA maps document chunks into lightweight parameter updates. Using public Qwen3-4B weights, we generate rank-8 \texttt{down\_proj} LoRAs and retain only the final three chunk adapters. \\
$\bullet$ \textbf{SHINE}~\cite{liu2026shine}: SHINE encodes historical information as generated low-rank parameter updates.
We use public Qwen3-8B weights and rank-8 LoRAs on all linear layers.
We use the full history when it contains at most 2,380 tokens.
Otherwise, we retain only the last 2,380 tokens before generating the LoRA. \\
$\bullet$ \textbf{$\delta$-mem}~\cite{lei2026delta}: $\delta$-mem maintains a recurrent memory state updated during inference. We use the rank-8 Qwen3-4B TSW adapter and write each history into its online state token by token. \\
$\bullet$ \textbf{LiveMem}~\cite{liu2026livemem}: We choose LiveMem-RL, which maintains a recurrent state while releasing historical key-value caches after writing.
We use its Qwen3-4B implementation and retain the latest 257K tokens on MemoryAgentBench.\\
$\bullet$ \textbf{MemoryLLM}~\cite{wang2024memoryllm}: MemoryLLM stores history in a fixed pool of learned memory vectors.
We use MemoryLLM-8B-Chat with 12,800 vectors per layer and a 2,048-token query window, completing 1,993 MemoryAgentBench questions.\\
$\bullet$ \textbf{ICAE}~\cite{ge2023context}: ICAE compresses each segment into continuous latent slots. We use Mistral-7B-Instruct-v0.2, 128 slots per segment, and FIFO retention of four segments; this differs from the original SpanExtrap setting.

Graphiti, Mem0, MemOS, and OpenViking use BGE-M3 embeddings for retrieval and the
corresponding backbone for memory extraction and answer generation.

\subsection{Details of Pipelines and Prompts in Memory Evaluation}
\label{appendix:eval_memory_pipeline_details}

\subsubsection{Query Prompt Construction}
\label{appendix:memory_query_prompt_construction}

We do not add generic few-shot demonstrations to the memory benchmarks.
In Full Context, we provide the history and current query in a single request and
truncate the history according to Appendix~D.2. For parametric memory methods, we write the
history to memory and query the model with only the task instructions and current question.
For external memory systems, we populate the answer template with memories retrieved from the
current scope. Graphiti, Mem0, MemOS, and OpenViking use the same answer template for each
benchmark, with the retrieved content supplied by each system. 

The principal English query templates are given below. Braced expressions denote fields
instantiated from each evaluation example at runtime. Unless stated otherwise, the templates
shown first are those used by the full context models.

\textbf{LoCoMo.}
The complete history and the question are placed in a single user message:

\begin{promptbox}
Below is a conversation between two people. Answer the question using only this conversation.

### {session_id}
Date: {timestamp}
{speaker}: {content}
...

Question: {question}
Answer briefly. Do not explain your reasoning.
\end{promptbox}

\textbf{LongMemEval-S.}
The complete history and the current date are formatted in a single user message:

\begin{promptbox}
I will give you several history chats between you and a user. Please answer the question based on the relevant chat history.

History Chats:
### Session {n}:
Session Date: {timestamp}
Session Content:
{role}: {content}
...

Current Date: {question_date}
Question: {question}
Answer:
\end{promptbox}

\textbf{MemBench.}
For participation tasks, the prompt begins with:

\begin{promptbox}
Below is the complete chronological conversation between a user and an assistant. Answer the multiple-choice question using only this history.
\end{promptbox}

For observation tasks, it instead begins with:

\begin{promptbox}
Below is the complete chronological sequence of user messages observed by an assistant. Answer the multiple-choice question using only these observations.
\end{promptbox}

Both task types then use the same structure:

\begin{promptbox}
History:
User / Assistant or Observed user: {content}
...

Current time: {question_time}
Question: {question}
Choices:
A. {option}
B. {option}
C. {option}
D. {option}
Return only the correct option letter: A, B, C, or D.
\end{promptbox}

\textbf{PersonaMem-32K.}
History messages retain their original roles and chronological order. Except for the first message, system messages in the history are converted to user messages and prefixed with \texttt{[System message]}. The final user message is:

\begin{promptbox}
{question}

Find the most appropriate model response and give your final answer (a), (b), (c), or (d) after the special token <final_answer>.

{options_text}
\end{promptbox}

\textbf{PS-Bench.}
The history and the latest user input are placed in one user message:

\begin{promptbox}
Reply in a natural, spoken tone, optionally using relevant memory or user personality details when appropriate.

Conversation history:

### {session_id}
Date: {timestamp}
{speaker}: {content}
...

User's latest input: {question}
\end{promptbox}

\textbf{PerMemSafe-conv64.}
The system message is fixed as follows:

\begin{promptbox}
You are a helpful assistant.
\end{promptbox}

Historical messages retain their original roles, and the current question is appended as the final user message:

\begin{promptbox}
{question}
\end{promptbox}

\textbf{Memora Weekly.}
The system message is:

\begin{promptbox}
You are an AI assistant with memory capabilities. The conversation history below contains ALL your previous conversations with this user - this history is PROVIDED TO YOU in this prompt.

Your task is to answer the user's question based EXCLUSIVELY on the conversation history that is provided below. The conversation history contains all the information you need to answer questions about the user's preferences, activities, goals, and past interactions.

IMPORTANT: Today's date is {question_date}. When answering questions about time periods (e.g., 'last 3 months', 'recent', 'this week'), calculate them relative to today's date ({question_date}).

CRITICAL INSTRUCTIONS:
1. The conversation history is PROVIDED IN THIS PROMPT - you have full access to it
2. Answer the question using ONLY information from the conversation history provided below
3. Do NOT say "I don't have access" or "I cannot access" - you DO have access because the history is right here in this prompt
4. Extract relevant information from the conversation history and use it to answer the question
5. If you learned about user preferences, activities, content, or goals from the conversations below, use them in your response
6. If information was mentioned in earlier conversations but later updated or deleted, use only the most recent information
7. Be specific and reference relevant details from the conversations
8. If the conversation history doesn't contain enough information to answer the question, say so clearly (but do NOT say you don't have access)
9. Do not make up information that wasn't in the conversations
\end{promptbox}

The two sentences involving \texttt{question\_date} are included only when that field is available. The user message is:

\begin{promptbox}
# Conversation History
The following contains all your previous conversations with this user, in chronological order:

## Session {session_id} - {timestamp}
User / Assistant: {content}
...

---

# Current Question
User: {question}

Based on the conversation history above, please provide your answer:
\end{promptbox}

\textbf{MemOps.}
The complete history is organized into context blocks in one user message:

\begin{promptbox}
Operation Evaluation: I will give you conversation history between you and a user. Answer based only on the provided conversation context. First perform operation detection, target extraction, and state transition reasoning over the provided context. If the context does not contain enough information, say that you do not know. If the history contains an instruction to forget or update a fact, follow the latest such instruction and do not reveal forgotten information.

Forget redaction rules:
- For a forget operation, predicted_operations[].old_value must not output the forgotten raw value. Use "[FORGOTTEN]", null, or a safe redacted description such as "removed target".
- For a forget operation, safe provenance may quote the deletion request (for example, "please forget my UBC GPA"), but do not quote the original fact that contained the forgotten raw value.

Return only compact JSON with exactly these top-level fields:
{
  "predicted_operations": [
    {
      "type": "remember | forget | update | reflect",
      "target": "memory target affected",
      "old_value": "old value if any; for forget use [FORGOTTEN], null, or a safe redacted description",
      "new_value": "new/current/inferred value if any",
      "state_after": "active | forgotten | updated | insufficient | inferred",
      "provenance": ["short quote or context block id"]
    }
  ],
  "answer": "final answer to the question",
  "provenance": ["short quote or context block id used for the answer"]
}

Conversation Context:
### Context block {n}
Content:
{role}: {content}

Question: {question}
JSON:
\end{promptbox}

For candidate-disambiguation examples, the following block is inserted immediately before the question:

\begin{promptbox}
Candidate options:
{candidate_options}
Choose one candidate. You may answer with the letter or the full option text.
\end{promptbox}

\textbf{ImplicitMemBench.}
We add no task instruction. Historical messages retain their original roles.
Except for the first message, system messages are converted to user messages and
prefixed with \texttt{[System feedback]}. The original current probe is appended
verbatim as the final user message.

\textbf{MemoryAgentBench.}
The system message is fixed as follows:

\begin{promptbox}
You are a helpful assistant that can read the context and memorize it for future retrieval.
\end{promptbox}

The user message concatenates a source-specific memory segment and a source-specific query segment. The value of \texttt{\{time\_stamp\}} is fixed to \texttt{1970-01-01 00:00:00}. The prefixes below identify the source family and are not sent to the model. The memory segments are:

\begin{promptbox}
ruler: Dialogue between User and Assistant {time_stamp}
<User> The following context is the documents I have read:
{context}
<Assistant> I have learned the documents and I will answer the question you ask.

longmemeval: Dialogue between User and Assistant
<User> The following context is the conversation between the user and the assistant:
{context}
<Assistant> I have memorized the conversation and I will answer the question you ask.

eventqa: Dialogue between User and Assistant {time_stamp}
<User> The following context is the book excerpt:
{context}
<Assistant> I have read the book excerpt and I will answer the question you ask.

icl: Dialogue between User and Assistant {time_stamp}
<User> The following context is the examples I have learned:
{context}
<Assistant> I have learned the examples and I will answer the question you ask.

recsys: Dialogue between User and Assistant {time_stamp}
<User> The following context is the dialogues between a user and recommender system:
{context}
<Assistant> I have memorized the dialogues and I will answer the question you ask.

infbench: Dialogue between User and Assistant {time_stamp}
<User> The following context is the book I have read:
{context}
<Assistant> I have read the book and I will answer the question you ask.

detective: Dialogue between User and Assistant {time_stamp}
<User> The following context is the book I have read:
{context}
<Assistant> I have read the book and I will answer the question you ask.

factconsolidation: Dialogue between User and Assistant {time_stamp}
<User> The following context is the facts I have learned:
{context}
<Assistant> I have learned the facts and I will answer the question you ask.
\end{promptbox}

The corresponding query segments are:

\begin{promptbox}
ruler: Answer the question based on the memorized documents. Only give me the answer and do not output any other words.

Question: {question}

Answer:

longmemeval: The history chats are between you and a user. Based on the relevant chat history, answer the question as concisely as you can, using a single phrase if possible.

{question}

Answer:

eventqa: Based on the context you memorized, complete the task below:

{question}

The event that happens next is:

icl: Use the provided mapping from the context to numerical label to assign a numerical label to the context. Only output "label: {{label}}" and nothing else.

{question}

label:

recsys: Pretend you are a movie recommender system. You need to recommend movies based on the dialogues you have memorized. Now I will give you a new conversation between a user and you (a recommender system). Based on the conversation, you reply me with 20 recommendations without extra sentences.

For Example:

[Conversation]

The recommendations are:
1.movie1
2.movie2
...

Here is the conversation: {question}

The recommendations are:

infbench: You are given a book above and you are tasked to summarize it.

{question}

Now summarize the book.

detective: Based on the context you memorized, answer the question below. You are required to answer the question based on the strict output format.

{question}

factconsolidation: Pretend you are a knowledge management system. Each fact in the knowledge pool is provided with a serial number at the beginning, and the newer fact has larger serial number. You need to solve the conflicts of facts in the knowledge pool by finding the newest fact with larger serial number. You need to answer a question based on this rule. You should give a very concise answer without saying other words for the question **only** from the knowledge pool you have memorized rather than the real facts in real world.

For example:

[Knowledge Pool]

Question: Based on the provided Knowledge Pool, what is the name of the current president of Russia?
Answer: Donald Trump

Now Answer the Question: Based on the provided Knowledge Pool, {question}
Answer:
\end{promptbox}

\textbf{BEAM-1M.}
The system message is:

\begin{promptbox}
You are a helpful assistant answering questions about a long-running conversation. Use only the conversation information available to you, respect later updates, and do not invent missing details.
\end{promptbox}

Each historical record in the user message is formatted as:

\begin{promptbox}
[{session_id} | {timestamp}]
User / Assistant: {content}
\end{promptbox}

The question segment is:

\begin{promptbox}
Only provide the answer without explaining your reasoning.
Put one event on each line in chronological order.

Question: {question}

Answer:
\end{promptbox}

The second instruction is included only for Event Ordering and is omitted for all other BEAM-1M capabilities.

\textbf{Benchmark-specific adaptations for parametric and external memory.}\par

\noindent\textbf{LoCoMo.}
For memory-based evaluation, ``Below is a conversation between two people. Answer
the question using only this conversation.'' is replaced by ``Answer the question
using only the conversation stored in your memory.'' In the external-memory
version, the retrieved content is introduced by ``Below are retrieved memories
from a conversation between two people.'' Open-ended questions require a brief
answer, whereas multiple-choice questions require only \texttt{(a)} or
\texttt{(b)}.

\noindent\textbf{LongMemEval-S.}
The opening instruction is changed from ``I will give you several history chats between you and a user. Please answer the question based on the relevant chat history.'' to ``Answer the question based on the relevant chat history stored in your memory.'' The empty \texttt{History Chats:} block is removed. For external memory systems, the query contains \texttt{Retrieved Memory Chats}, the current date, the question, and an \texttt{Answer:} cue.

\noindent\textbf{MemBench.}
For participation tasks, the opening instruction becomes ``Answer the multiple-choice question using only the conversation stored in your memory.'' For observation tasks, it becomes ``Answer the multiple-choice question using only the observations stored in your memory.'' The empty \texttt{History:} block is removed in both cases. The external-memory query contains a \texttt{Retrieved memories} section followed by the current time, the question, and answer choices A--D, and requires only the option letter.

\noindent\textbf{PersonaMem-32K.}
The original query does not state that the history appears in the current prompt, so removing the history body requires no additional wording change. For external memory systems, the query contains the retrieved memories, the question, four candidate responses, and the original \texttt{<final\_answer>} constraint.

\noindent\textbf{PS-Bench.}
No query-stem replacement is required after the history body is removed. The external-memory query combines the natural-tone instruction, the retrieved memories, and the current user input. Base questions omit the memory section because they are evaluated with an empty state.

\noindent\textbf{PerMemSafe-conv64.}
No additional query-stem replacement is required. The system message remains ``You are a helpful assistant.'', and the external-memory user message contains the retrieved memories followed by the current question.

\noindent\textbf{Memora Weekly.}
Statements asserting that the history is supplied in the current prompt are changed consistently to state that it is stored in and accessible from memory.
Specifically, ``The conversation history below contains ALL your previous conversations with this user---this history is PROVIDED TO YOU in this prompt'' becomes ``Your memory contains ALL previous conversations with this user''; ``conversation history that is provided below'' becomes ``conversation history stored in memory''; ``The conversation history is PROVIDED IN THIS PROMPT---you have full access to it'' becomes ``The conversation history is stored in memory---you have access to it''; and ``because the history is right here in this prompt'' becomes ``because the history is available in memory.'' For external
memory systems, the system instruction requires the answer to rely only on retrieved memories and to use the most recent value. The user message contains \texttt{\# Retrieved Memories} followed by \texttt{\# Current Question}.

\noindent\textbf{MemOps.}
``I will give you conversation history between you and a user. Answer based only on the provided conversation context.'' is replaced by ``Answer based only on the conversation history stored in your memory.'' The phrase ``transition reasoning over the provided context'' becomes ``transition reasoning over that memory''; ``If the context does not contain enough information'' becomes ``If your memory does not contain enough information''; and ``If the history contains an instruction to forget or update a fact'' becomes ``If your memory contains an instruction to forget or update a fact.'' The empty \texttt{Conversation Context:} block is removed. The external-memory query retains the operation-detection and forget-redaction instructions and contains a \texttt{Retrieved Memory Context} section, optional candidate choices, the question, and the \texttt{JSON:} output cue.

\noindent\textbf{ImplicitMemBench.}
Removing the history body requires no additional wording change. In the external-memory template, \texttt{Retrieved memories} is followed directly by the \texttt{Current probe}; the expected response pattern and scoring rubric are not included.

\noindent\textbf{MemoryAgentBench.}
Removing the history body requires no additional wording change. The external-memory query retains the fixed system message, describes the retrieved memories according to the source family, and appends the corresponding source-specific query.

\noindent\textbf{BEAM-1M.}
Removing the history body requires no additional wording change. The external-memory template contains \texttt{\# CONTEXT},
\texttt{\# INSTRUCTIONS}, \texttt{\# QUESTION}, and \texttt{\# ANSWER}.
Event Ordering additionally requires one event per line in chronological order.

\subsubsection{Decoding Configuration}
\label{appendix:memory_decoding_configuration}

We disable thinking for memory-capability evaluations. The default context budget is
262,144 tokens (256K) for the chat template, input, and reserved output combined. Full
Context (Max) uses the model-specific windows given in Appendix~D.2.

We use the same decoding settings across conditions for each benchmark and backbone.
The temperature is 0 except for ImplicitMemBench's Priming subtask, where it is 0.8.
For NV, we set \texttt{top\_p} to 1.0. We score the generated text even when generation
reaches the output limit before an end-of-sequence token is emitted.

The maximum generation length is benchmark-specific. LoCoMo and LongMemEval-S use a maximum of 512 tokens, while MemBench uses a maximum of 16 tokens. PS-Bench uses a maximum of 512 tokens. PersonaMem-32K uses a maximum of 2,048 tokens.
MemOps, PerMemSafe-conv64, and Memora Weekly also use a maximum of 2,048 tokens.
ImplicitMemBench uses a maximum of 4,096 tokens.

For BEAM-1M, the maximum depends on the capability being evaluated. Abstention, Information Extraction, Knowledge Update, and Temporal Reasoning use 512 tokens.
Contradiction Resolution, Instruction Following, Multi-session Reasoning, and Preference Following use 1,024 tokens. Event Ordering uses 2,048 tokens, and Summarization uses 4,096 tokens.

For MemoryAgentBench, the maximum is source- and task-specific. RULER QA uses 50 tokens, EventQA uses 40 tokens, the LongMemEval-S source uses 50 tokens, recommendation tasks use 300 tokens, and ICL tasks use 20 tokens. The maximum is 1,200 tokens for summarization, 2,000 tokens for DetectiveQA, and 10 tokens for
FactConsolidation.

\subsection{Details of Pipelines and Prompts in General Evaluation}
\label{appendix:eval_general_pipeline_details}

\textbf{Inference settings.} We enable thinking and use a temperature of 1.0 for all
general-capability evaluations. For a given benchmark and backbone, we use the same task
prompt version, parser, and scorer across conditions. Each backbone and its MemT variant
share the sampling settings and token budgets below. Output budgets include both reasoning
and final-answer tokens and override the benchmark implementations' default limits.

\textbf{NV.} We set \texttt{top\_p} to 0.95 and the presence penalty to 0.0.
The context window is 1,048,576 tokens, and the output limit is 262,144 tokens.

\textbf{DS.} We use high reasoning effort and set \texttt{top\_p} to 1.0 and the
presence penalty to 0.0. The context window is 262,144 tokens, and the output limit is
131,072 tokens or the remaining context capacity, whichever is smaller.

\textbf{QW.} We set \texttt{top\_p} to 0.95, \texttt{top\_k} to 20, and
\texttt{min\_p} to 0.0, with a presence penalty of 1.5 and a repetition penalty of 1.0.
The output limit is 81,920 tokens. The context window is 262,144 tokens, except on
LongBench v2, where the official YaRN configuration uses an expansion factor of 4
and a window of 1,010,000 tokens.

\textbf{Noise conditions.} The clean backbone receives the benchmark input without
added demonstrations. For each benchmark, we use the same set of unrelated demonstrations
in both noise conditions. We prepend the demonstrations to the task input for the backbone
with noise. For MemT-1.0, we write the entire set to an initially empty memory state in one
operation before evaluation. Each question independently queries this fixed memory state.

We use three BigCodeBench-Hard question--answer examples for MMLU-Pro and GPQA Diamond
and ten AIME 2026 examples for IFEval, LongBench v2, BigCodeBench-Hard, and LiveCodeBench v6.
Each set contains fewer than 2,048 tokens. We select examples from tasks whose primary
capability differs from that of the target benchmark, independently of model outputs and
evaluation scores. We exclude target questions, direct rewrites of target answers, and
benchmark-native few-shot examples.

\textbf{MMLU-Pro.}
We use the official subject-specific five-shot chain-of-thought prompt. The model answer is parsed only when an explicit answer marker is followed by an option from A--J or when an A--J option appears on a separate line. We report micro-averaged accuracy. Unparseable answers and answers containing mutually conflicting choices are counted as incorrect.

\textbf{GPQA Diamond.}
For each underlying question, we evaluate it eight times using a fixed answer-option order. Every evaluation condition shares exactly the same option orders. We first aggregate the eight responses within each underlying question and then compute accuracy over the 198 underlying questions.

\textbf{IFEval.}
We retain the dataset's original prompts and add no task-level few-shot examples.
The official programmatic constraint checker is applied to every instruction. We report prompt-level strict accuracy, counting a prompt as correct only when all of its associated constraints are satisfied.

\textbf{LongBench v2.}
We use the official zero-shot direct-answer prompt. Inputs longer than 120,000 tokens are truncated according to the official middle-truncation procedure using the \texttt{o200k\_base} tokenizer. Answers are extracted with the official implementation at commit \texttt{c5ea10b}, and we report accuracy.

\textbf{BigCodeBench-Hard.}
We use the v0.1.4 Hard/Instruct tasks without prefilled code and generate one candidate solution per task. Each candidate is executed in the official BigCodeBench v0.2.4 container. We report Pass@1.
Execution failures and outputs containing no valid code are counted as incorrect.

\textbf{LiveCodeBench v6.}
We use the cumulative \texttt{release\_v6} test split and the official code generation prompt, generating one candidate solution per problem.
Candidates are executed with the official LiveCodeBench runner at commit \texttt{28fef95}.
We report Pass@1, counting execution failures and outputs without valid code as incorrect.

All conditions within a benchmark use the same task prompts, demonstration selection, parsing rules, scorer versions, and execution settings.
Programmatic checks and code execution are performed with the frozen versions
specified above. Empty outputs, generation failures, and unparseable answers remain in the predetermined denominator and are assigned the failure outcome defined by the corresponding benchmark.

\subsection{Details of Storage Overhead Evaluation}
\label{appendix:eval_storage_overhead}

For the storage overhead evaluation, note that process-level \texttt{max\_memory\_allocated} includes model weights and memory pools, and therefore does not reflect the per-session storage overhead.

For full-context inference, we report the logical state size after history preparation and before querying.
The storage grows linearly with the visible context length and is capped at 256K tokens on BEAM.
DenseRAG stores FP32 retrieval vectors together with their corresponding chunk texts.
Temp-LoRA stores the query-time BF16 adapter while excluding optimizer states.
MemT-1.0 stores a full-rank memory state.

\section{Additional Details on Agentic Tasks}
\label{appendix:detail_application}

\subsection{Experience Bank and Training Data Construction}

\subsubsection{Experience Bank Construction}

We construct the experience bank in four stages. First, we treat each trajectory as an episode and annotate it with its key decisions and a state signature comprising topic tags, the first tool used, and the error code. Second, episodes with the same state signature are pooled, and DeepSeek-V4-Pro~\cite{deepseekai2026deepseekv4} induces a policy of the form ``when condition \(X\) holds, take action \(Y\), verify with \(Z\), and watch out for \(W\)'' from each pool. Each policy must specify a trigger observable from the agent's state. Source-specific entities such as repository, product, and file names are abstracted into variables. Third, policies describing the same workflow are clustered based on the lexical similarity of their titles, triggers, and procedures, with one policy retained as the representative of each cluster. Finally, DeepSeek-V4-Pro synthesizes a structured experience from the trajectories supporting each representative policy. An automated check verifies that the declared tools appear in the supporting trajectories, and a separate LLM audit
filters out experiences containing severe errors.

We summarize the construction funnel and the
resulting experience bank in Table~\ref{tab:experience-bank}. Each experience contains ten fields, including a retrieval description to support matching, a trigger context specifying when it applies, steps and parameters specifying what to do, preconditions defining its scope of applicability, and decision guidance documenting preferred choices and potential pitfalls.

Most experiences are specific, since 942 of the 1{,}040 derive from a single trajectory, while a few broad workflows draw on hundreds. The bank emphasizes procedural guidance and verification, with verification and testing-related tags among the most frequent after the python tag. Issue-level retrieval provides limited discrimination among top-ranked candidates: the median difference in similarity scores between the first- and fifth-ranked candidates is only 0.029.

\begin{table}[htbp]
  \centering
  \caption{Experience bank construction and statistics. Token
  counts use the Qwen3.6-35B-A3B tokenizer.}
  \label{tab:experience-bank}
  \small
  \begin{adjustbox}{max width=\linewidth}
  \begin{tabular}{lr}
    \toprule
    \multicolumn{2}{l}{(a) Construction stages}\\
    \midrule
    Source trajectories (23{,}667 instances, 2{,}562 repositories) & 41{,}057\\
    Induced policies (88.5\% cite a single trajectory) & 3{,}678\\
    Representative policies after clustering & 1{,}200\\
    Audit: usable / usable with caveat / severe error & 401 / 639 / 160\\
    \textbf{Experiences in the final bank} & \textbf{1{,}040}\\
    \midrule
    \multicolumn{2}{l}{(b) Bank statistics}\\
    \midrule
    Fields per experience & 10\\
    Length as JSON / as rendered text (tokens, mean) & 715 / 456\\
    Rendered text, 10th / 90th percentile (tokens) & 391 / 521\\
    Steps per experience (range; experiences with five steps) & 3--6; 573\\
    Source trajectories per experience (mean; single-source experiences) & 3.0; 942\\
    Most frequent tags: \emph{python} / \emph{verification} / \emph{testing} / \emph{pytest} & 792 / 192 / 168 / 133\\
    Top-1 cosine similarity (mean; 10th--90th percentile) & 0.611; 0.556--0.668\\
    Median similarity gap between top-1 and top-5 & 0.029\\
    \bottomrule
  \end{tabular}
  \end{adjustbox}
\end{table}

\subsubsection{Experience-Conditioned Trajectory Construction}

\begin{figure}[tb]
    \centering
    \includegraphics[width=\linewidth]{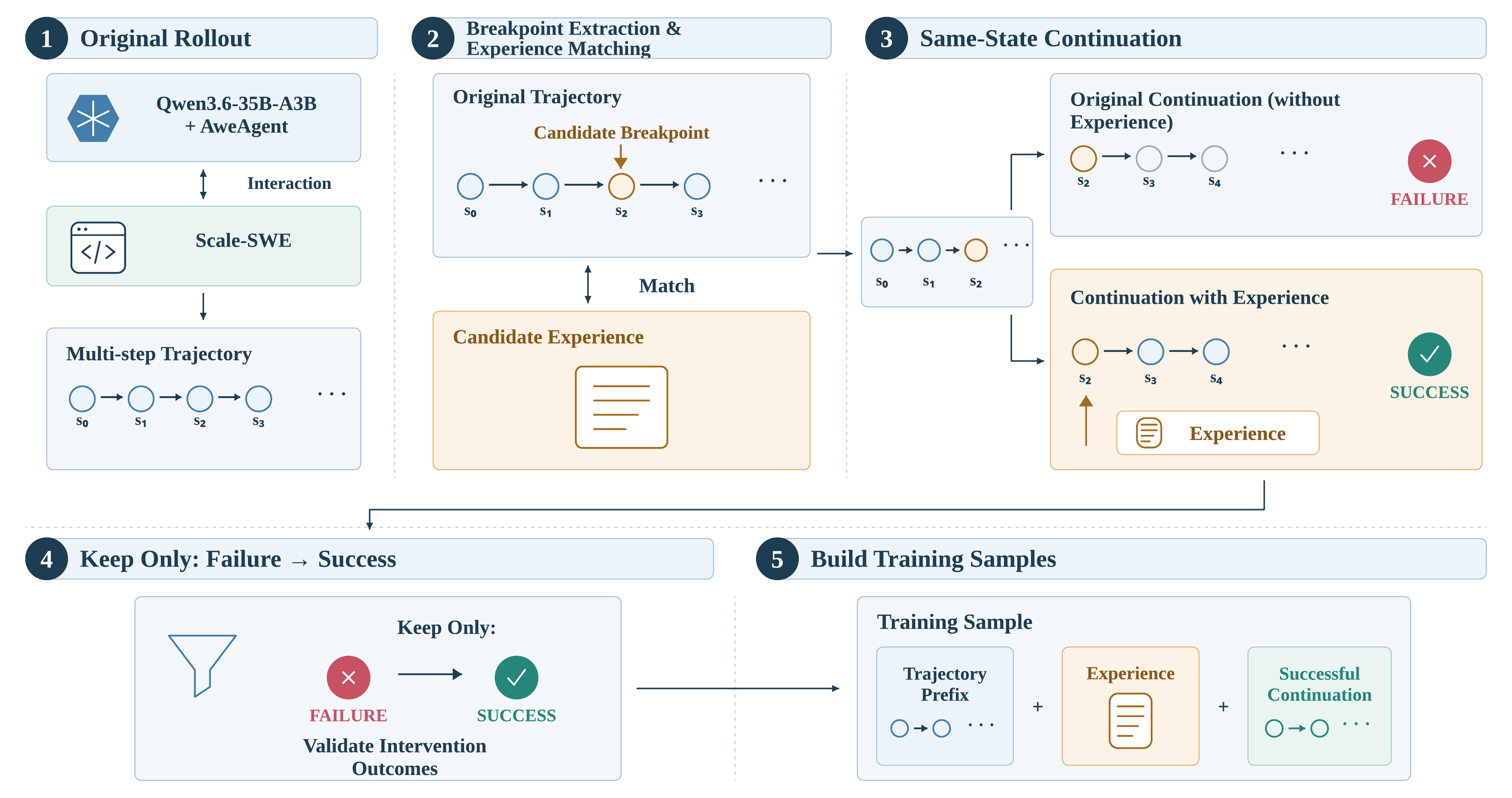}
    \caption{Construction of experience-conditioned training trajectories. Retrieved experience is added at selected breakpoints in unsuccessful trajectories, and execution resumes from the same state. Continuations that pass the task tests are retained as training samples comprising the trajectory prefix, experience, and successful continuation.}
    \label{fig:data-construction-pipeline}
\end{figure}

We construct experience-conditioned training trajectories in five stages, as demonstrated in Figure~\ref{fig:data-construction-pipeline}: generating original trajectories, identifying candidate breakpoints, selecting and injecting experiences, validating the intervention outcomes, and serializing the resulting training examples.

We represent an agent trajectory as
\[
    \xi =
    \left(
        h_0, a_0, o_0,
        \ldots,
        h_T, a_T, o_T
    \right),
\]
where \(h_t\), \(a_t\), and \(o_t\) denote the interaction history before step \(t\), the action produced by the agent, and the corresponding environment observation, respectively.

\paragraph{Breakpoint Identification}
\label{subsec:breakpoint-extraction}

For unsuccessful trajectories, we sequentially scan the observations in each rollout. A step $t$ is marked as a candidate breakpoint based on execution feedback, including command failures, empty observations after normalization, and repeated unsuccessful operations. This criterion relies only on feedback visible to the agent at execution time and does not access reference solutions or privileged evaluation information.

To prevent long trajectories from producing a disproportionate number of intervention variants, we retain at most eight breakpoints per rollout. If a rollout contains more than eight candidates, we sample eight of them without replacement using a fixed random seed. We additionally record the total number of candidate breakpoints and the positions of the retained breakpoints to ensure reproducibility. The Harness-Triggered evaluation setting uses the same breakpoint-detection logic, with at most one experience store operation per trajectory. Before Execution evaluates robustness to experience injection timing.

\paragraph{Experience Retrieval and Counterfactual Validation}
\label{subsec:experience-validation}

For each task, we first retrieve the top-10 most relevant entries from the experience bank. We then construct ten candidate experience subsets for every retained breakpoint: one subset containing the complete top-3 entries, one containing the complete top-5 entries, and eight three-entry subsets sampled from the top-10 retrieved entries. Random sampling is performed using a fixed seed associated with each task instance.

Each breakpoint-experience-subset pair defines an independent continuation rollout. The resulting intervention trajectory is
\[
    \xi_{b,j}
    =
    \xi_{\leq b}
    \oplus
    [g_{b,j}]
    \oplus
    \hat{\xi}_{>b}^{(g_{b,j})}.
\]
Here, $b$ indexes the selected breakpoint, $j$ indexes the candidate experience subset, and \(\oplus\) denotes concatenation of message sequences. We retain a trajectory as a training sample if and only if
\[
    \begin{cases}
        \operatorname{Success}(\xi) = 0, \\
        \operatorname{Success}(\xi_{b,j}) = 1,
    \end{cases}
\]
which means the original trajectory must fail to solve the task, whereas the experience-conditioned continuation must pass the task tests. Variants that fail the tests, terminate abnormally, or cannot be reproduced are excluded from the training set.

\paragraph{Training Sample Serialization}
\label{subsec:training-example-serialization}

Finally, each training sample is serialized as
\[
    x_{b,j}
    =
    \left(
        \xi_{\leq b},
        g_{b,j},
        \hat{\xi}_{>b}^{+\left(g_{b,j}\right)}
    \right),
\]
where the three components correspond to the triggering execution state, the experience subset, and the successful continuation suffix, respectively. The superscript \(+\) indicates that the generated suffix passes the task evaluation.

\subsubsection{Dataset Statistics and Splits}

We partition the constructed data into training and validation sets by
task, keeping all original rollouts, breakpoints, and intervention
variants associated with each task in the same split, and SWE-bench Verified
serves as the test set. We report the number of tasks and repositories of each split in Table~\ref{tab:data-splits}.

\begin{table}[tb]
    \centering
    \caption{Task and repository counts for the training and validation splits and SWE-bench Verified.}
    \label{tab:data-splits}
    \begin{tabular}{lrrr}
        \toprule
        \textbf{Statistic} &
        \textbf{Train} &
        \textbf{Validation} &
        \textbf{Test (Verified)} \\
        \midrule
        Number of repositories & 537   & 176 & 12  \\
        Number of tasks        & 2,396 & 266 & 500 \\
        \bottomrule
    \end{tabular}
\end{table}

Throughout data construction, we check breakpoint validity, consistency
between trajectory prefixes and repository snapshots, test outcomes,
message and tool-call structure, and context length. We exclude samples
with environment failures, incomplete logs, test-file modifications,
other invalid execution behaviors, or outcomes that cannot be
reproduced. Samples that exceed the context-length limit or fail
structural checks are discarded in full rather than truncated. The scale of the resulting data construction process and the outcomes of experience intervention are reported in Table~\ref{tab:construction-statistics}.

\begin{table*}[tb]
    \centering
    \caption{Training data construction statistics and experience intervention outcomes.}
    \label{tab:construction-statistics}
    \begin{tabular}{p{0.31\textwidth} r p{0.47\textwidth}}
        \toprule
        \textbf{Statistic} &
        \textbf{Value} &
        \textbf{Calculation or description} \\
        \midrule
        Original trajectories
        &
        18,820
        &
        Target rollouts generated by Qwen3.6-35B-A3B
        \\

        Unsolved original trajectories
        &
        14,649 (77.8\%)
        &
        \(14{,}649 / 18{,}820\)
        \\

        Breakpoint--experience variants
        &
        199,500
        &
        7,901 usable trajectories produced 19,950 retained breakpoints
        \\

        Successful experience
        &
        45,481 (22.8\%)
        &
        \(45{,}481 / 199{,}500\)
        \\
        \bottomrule
    \end{tabular}
\end{table*}

\subsection{Training Details}

We compare training configurations along three dimensions: the objective (CE or KD), the supervision scope (First-Step or Full-Suffix), and the number of successful training trajectories selected per task. Max-2 and Max-3 retain at most two and three successful training trajectories per task, respectively.

\subsubsection{Trainable Parameters and Memory Operations}

We train Metis on hyper memory blocks at decoder layers
$\ell \in \{3, 7, 11, 15, 19, 23, 27, 31, 35, 39\}$. Each block contains 10
 trainable tensors: \text{W\_k.weight}, \text{W\_v.weight},
\text{pool\_score.weight}, \text{gated\_delta\_alpha.weight},
\text{gated\_delta\_alpha.bias}, \text{gated\_delta\_beta.weight},
\text{gated\_delta\_beta.bias}, \text{query\_proj.weight},
\text{query\_norm.weight}, and \text{mem\_norm.weight}. This gives 100
tensors and 104,962,580 trainable parameters in total, roughly 0.3\% of
the backbone's parameter count. The backbone runs in BF16, whereas the
memory computation, state accumulation, and all trainable
parameters are kept in FP32.

Each memory block maintains a fixed-size memory consisting of a
$512 \times 512$ value state and a $512 \times 1$ key normalizer, so the
memory footprint does not grow with the length of the experience. The key
and value projections map the hidden states of the experience to
512-dimensional vectors, while the query projection maps the hidden states
of the main trajectory to memory queries. On the write side, the token selector applies alpha top-$p$ selection with
$p = 0.9$ and always retains at least one token; the delta update ratio is
set to 0.9. On the read side, the memory readout is normalized by
\texttt{mem\_norm} and fused with the backbone representation using a
fusion coefficient of $\gamma = 0.9$. Runtime memory is not retained across tasks. The checkpoint contains only the
trainable memory parameters listed above, and every new task or
trajectory starts from an empty memory state.

\subsubsection{Sample Serialization and Supervision Masks}

Each training sample corresponds to exactly one write event, and the
memory is reset to empty at the start of every sample. Within a sample,
tokens are split into two attention domains: the \texttt{experience}
domain, which contains the experience text, and the \texttt{main} domain,
which contains the trajectory prefix and the successful continuation. The
\texttt{main} domain does not attend to the \texttt{experience} domain, so
the experience can influence the main trajectory only through memory.

Replay then proceeds in three stages, which mirror the injection of
experience at an execution breakpoint:
(1) \emph{write}: the experience is stored in memory at the end of the
experience segment;
(2) \emph{prefix replay}: the main-trajectory prefix is processed with
memory reading disabled; and
(3) \emph{continuation replay}: memory reading is enabled from the
breakpoint onward, so that all subsequent predictions can access the
experience-derived state $M$. Consequently, the prediction of the first supervised token is already
conditioned on the stored memory. In the training configuration, the
write offset is set to zero and no query store operation is performed.

\paragraph{Supervision Settings}\label{supervision-settings}

Within this replay procedure, we consider two supervision settings. They
differ in how much of the successful suffix is supervised and in where the
sequence is truncated. Both are trained with teacher forcing on the
original token sequences.

\textbf{Full-Suffix.} The loss covers every assistant turn in the
successful suffix. For each such turn, the target comprises all tokens
following the assistant role header: the reasoning text, the response
body, the tool calls, the structural tokens that delimit them, and the
closing
\texttt{\textless{}\textbar{}im\_end\textbar{}\textgreater{}\textbackslash{}n}.
All other tokens serve only as conditioning information. These include
the experience, the prefix, the role headers, tool responses, and user and
system messages. Because training is teacher-forced, each later assistant
turn is conditioned on the ground-truth tool responses recorded in the
original trajectory, rather than on the results of executing the model's
own actions.

\textbf{First-Step.} The loss covers only the first assistant turn after the
breakpoint, focusing supervision on the immediate continuation. We derive these samples from Full-Suffix trajectories whose
assistant turns contain reasoning text. For each trajectory, we select the
first assistant turn of the successful suffix and truncate the sequence
immediately after the closing
\texttt{\textless{}\textbar{}im\_end\textbar{}\textgreater{}\textbackslash{}n}
of this turn. The model input therefore consists of the experience
(written to memory), the full prefix, and the selected turn.

Under teacher forcing, every token of the selected turn is predicted from
the ground-truth tokens that precede it rather than from tokens generated
by the model. The loss is computed only over this turn, using the same
target definition as in Full-Suffix. Because truncation removes only
trailing tokens, the breakpoint position and the original token IDs remain
unchanged, and no re-tokenization is required.

\subsubsection{Training Objectives and Hyperparameters}

We consider two training
objectives: hard-label cross-entropy (CE), 
and knowledge distillation (KD). The student accesses experience only through native memory, the teacher uses the same frozen backbone with the memory modules
disabled and receives the experience as text in
prefix--experience--suffix order. Both models use teacher forcing
on the same successful continuation, with teacher and student
predictions aligned by target token rather than absolute input
position.

Let $\mathcal{T}$ denote the valid supervised positions across all
ranks and gradient-accumulation microbatches in one optimizer step,
excluding padding, and let $N = |\mathcal{T}|$.
For each position $t$, let $y_t$ be the target token and let
$p_t^{\mathrm{S}} =
\operatorname{softmax}(z_t^{\mathrm{S}})$ be the student distribution
over the full vocabulary $\mathcal{V}$.
We use temperature $\tau=1$ throughout.

\textbf{CE.}
We minimize the negative log-likelihood of the recorded target tokens:
\[
\mathcal{L}_{\mathrm{CE}}
= -\frac{1}{N}
  \sum_{t\in\mathcal{T}}
  \log p_{t,y_t}^{\mathrm{S}}.
\]

\textbf{KD.}
At each supervised position, we retain the $K=64$ tokens with the
largest teacher logits, denoted by $\mathcal{K}_t$, and renormalize
the teacher distribution over this set:
\[
q_{t,v}^{\mathrm{T}}
=
\frac{\exp(z_{t,v}^{\mathrm{T}})}
     {\sum_{u\in\mathcal{K}_t}\exp(z_{t,u}^{\mathrm{T}})},
\qquad v\in\mathcal{K}_t.
\]
The distillation objective is
\[
\mathcal{L}_{\mathrm{KD}}
=
\frac{1}{N}
\sum_{t\in\mathcal{T}}
\sum_{v\in\mathcal{K}_t}
q_{t,v}^{\mathrm{T}}
\log
\frac{q_{t,v}^{\mathrm{T}}}
     {p_{t,v}^{\mathrm{S}}}.
\]
Only the teacher distribution is truncated; the student remains
normalized over the full vocabulary. This objective transfers
experience-conditioned predictions from the textual-context
teacher to the memory-conditioned student.

The training and validation sets are split by task, so no task appears in both.

Training uses task interleaving. Unpacked records are replayed for three
epochs, and within each epoch they are arranged in a task-interleaved
order determined by a fixed base seed combined with the epoch index.

For the Full-Suffix-Max3 run, we additionally use sequence
packing to improve training throughput. Packs are first constructed with a
fixed seed, without splitting any trajectory across packs, and are then
scheduled in a deterministic order. Each pack contains at most 131,072
tokens. Trajectories in the same pack do not share memory: the memory is
reset, written, and read independently for each trajectory. Because every
step of the schedule is determined by fixed seeds, the data order of each
run is fully reproducible.

Each microbatch on a given rank contains exactly one trajectory or one
complete pack. Padding slots are assigned a supervised-token count of
zero, which ensures that each valid supervised token is counted exactly
once in loss normalization. We optimize with AdamW, using FP32 optimizer
states sharded across ranks, a learning rate of $2\times10^{-5}$, and
three warmup steps.

\subsubsection{Training Diagnostics}
\label{appendix:training_diagnostics}

We compare CE and KD across four paired training configurations,
combining First-Step or Full-Suffix supervision with the Max-2
or Max-3 data setting. Within each pair, initialization, training
data, and optimization hyperparameters are matched.
CE and KD optimize different objectives, so their training losses
are not directly comparable. We therefore monitor
hard-label validation negative log-likelihood (NLL) for both.
Validation comparisons are made within the same supervision scope
and data setting.

\begin{figure}[tb]
  \centering
  \begin{minipage}[t]{0.49\linewidth}
    \centering
    \includegraphics[width=\linewidth]
      {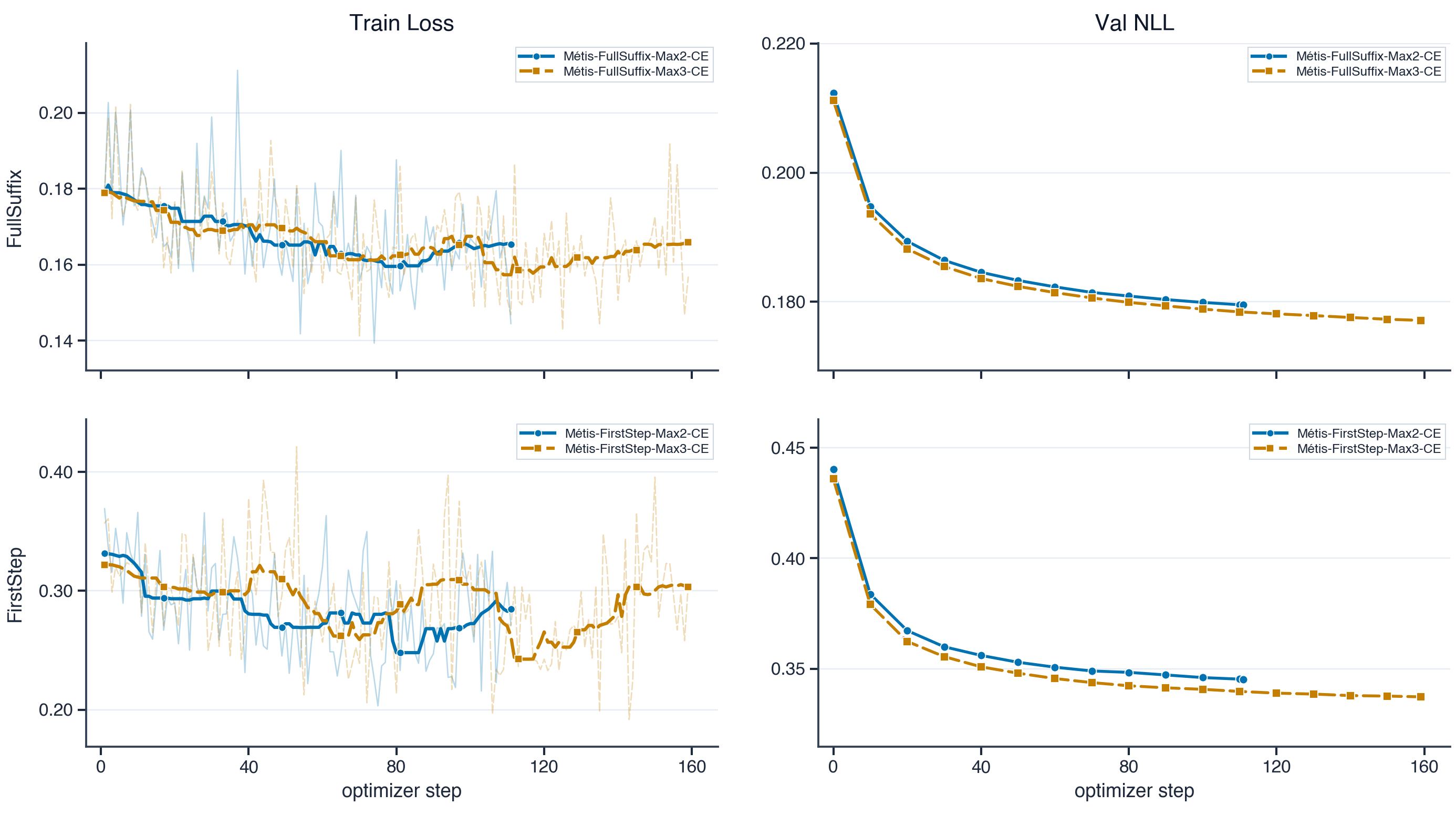}
    \\(a) CE objective
  \end{minipage}\hfill
  \begin{minipage}[t]{0.49\linewidth}
    \centering
    \includegraphics[width=\linewidth]
      {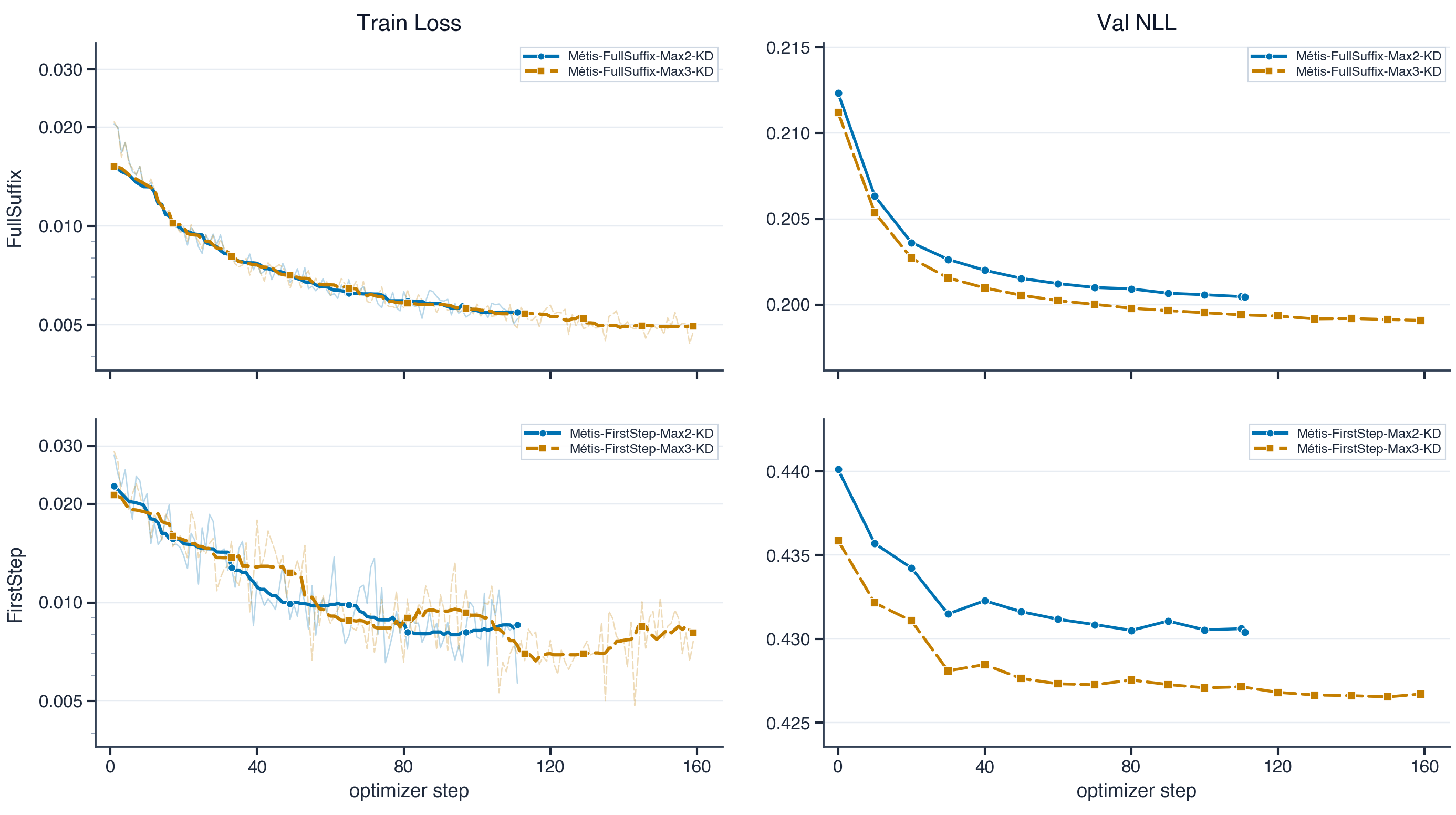}
    \\(b) KD objective
  \end{minipage}
  \caption{Training loss and validation NLL under CE and KD objectives.
  Within each panel, the left and right columns show training loss and validation NLL, respectively.
  Solid blue curves denote Max-2, and dashed orange curves denote Max-3.}
  \label{fig:training-logs}
\end{figure}

\paragraph{Optimization and Validation Behavior.}
Figure~\ref{fig:training-logs} shows decreasing validation NLL
under both objectives, with most of the reduction occurring
within the first 40 optimizer steps.
CE achieves lower final validation NLL than KD in all four
paired configurations
(Table~\ref{tab:ce-kd-diagnostics}).
For Full-Suffix-Max2, CE reduces body-token NLL from 0.2925
to 0.1866, compared with 0.2910 under KD.
The same pattern holds for tool-call tokens
(0.0603 to 0.0461 under CE versus 0.0545 under KD)
and reasoning tokens
(0.5033 to 0.4555 versus 0.4749).

\paragraph{Relation to Agent Performance.}
\textcolor{blue}{We compare training objectives under the Full-Suffix-Max2 configuration on the 100-task development subset of SWE-bench Verified, with three runs per task (Table~\ref{tab:ce-kd-diagnostics}). While CE achieves the highest average success rate, KD demonstrates superior performance on the pass@3 metric and maintains robust reliability on pass\textasciicircum{}3, justifying our preference for KD as the standard training objective.}

\begin{table}
  \centering
  \color{blue}
  \small
  \caption{\textcolor{blue}{Training objectives under Full-Suffix-Max2.}}
  \label{tab:ce-kd-diagnostics}
  \begin{tabular}{lrrr}
    \toprule
    Objective & Avg. Success & pass@3 & pass\textasciicircum{}3 \\
    \midrule
    KD                  & 54.33          & \textbf{67.00} & \textbf{44.00} \\
    CE                  & \textbf{55.00} & 66.00          & \textbf{44.00} \\
    $0.5$ KD + $0.5$ CE & 53.00          & 65.00          & 40.00          \\
    \bottomrule
  \end{tabular}
\end{table}

\subsection{Additional Results}
\subsubsection{Training Configuration Comparison}

\paragraph{Training Configurations.}
\textcolor{blue}{In Table~\ref{tab:v1_training_configurations}, we compare four KD training configurations on the 100-task development subset of SWE-bench Verified, with three runs per task. Full-Suffix-Max3 yields the highest average success and pass@3, whereas Full-Suffix-Max2 attains the peak pass\textasciicircum{}3 score. Given this trade-off, Full-Suffix-Max2 is selected to prioritize reliability in subsequent evaluations.}

\begin{table}[htbp]
  \centering
  \color{blue}
  \small
  \caption{\textcolor{blue}{MemT-1.0-Nano KD training configurations.}}
  \label{tab:v1_training_configurations}
  \begin{tabular}{lrrr}
    \toprule
    Configuration & Avg. Success & pass@3 & pass\textasciicircum{}3 \\
    \midrule
    First-Step-Max2  & 47.33          & 60.00          & 35.00          \\
    First-Step-Max3  & 52.67          & 64.00          & 43.00          \\
    Full-Suffix-Max2 & 54.33          & 67.00          & \textbf{44.00} \\
    Full-Suffix-Max3 & \textbf{55.00} & \textbf{69.00} & 39.00          \\
    \bottomrule
  \end{tabular}
\end{table}

\begin{figure*}[t]
  \centering
  \includegraphics[width=\textwidth]{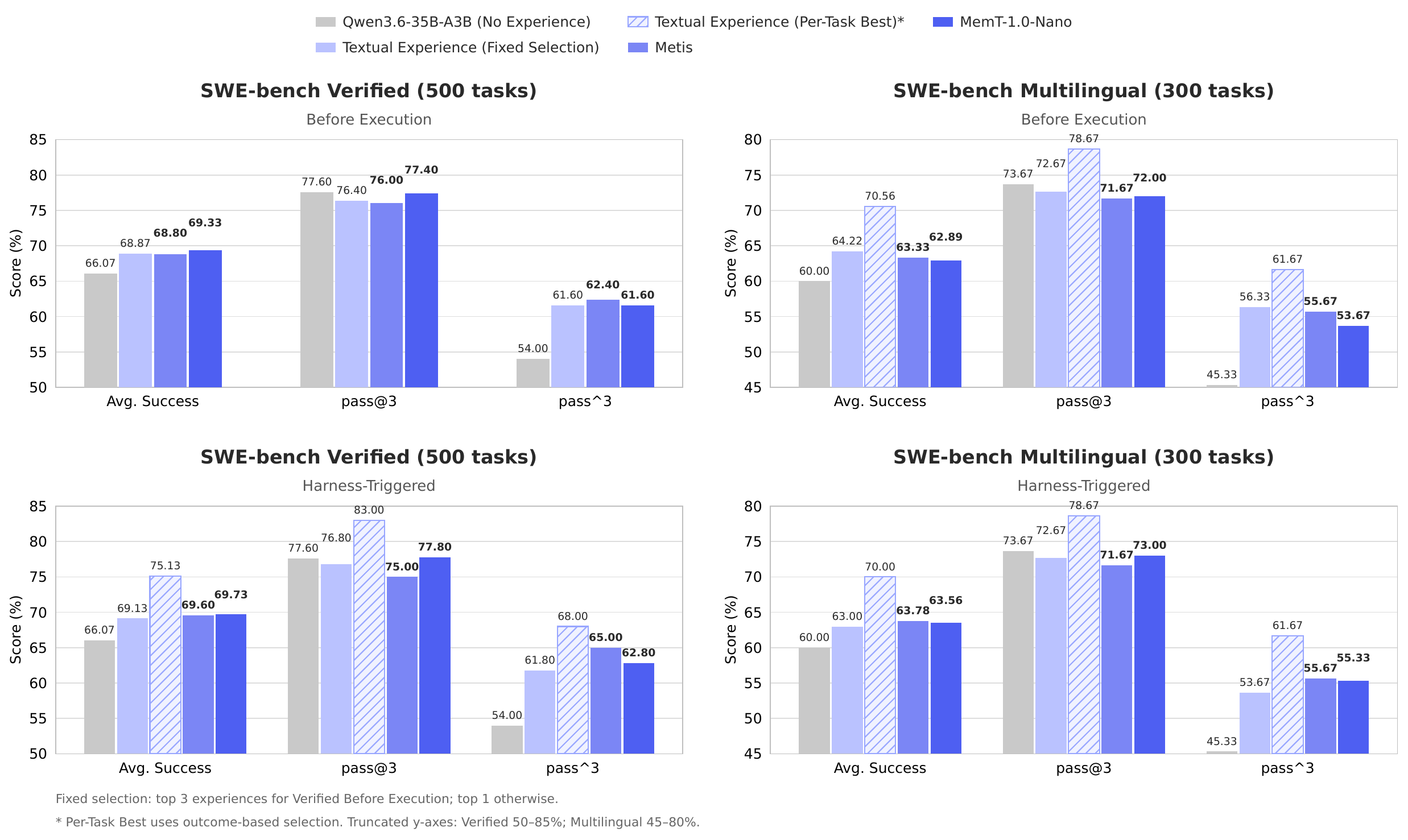}
  \caption{Agentic coding performance on SWE-bench Verified and SWE-bench Multilingual under
  Before Execution (top) and Harness-Triggered (bottom). Each task is evaluated in three runs.}
  \label{fig:swe_versions}
\end{figure*}

\paragraph{Full-Benchmark Comparison.}
In Figure~\ref{fig:swe_versions}, we compare Metis and MemT-1.0-Nano with the backbone and textual-experience baselines on SWE-bench Verified (500 tasks) and SWE-bench Multilingual (300 tasks). Both memory models improve average success and pass\textasciicircum{}3 over the backbone. Relative to Metis, MemT-1.0-Nano has higher pass@3 but lower pass\textasciicircum{}3 in all four settings. 

\subsubsection{Experience Selection}

We compare Textual Experience selection strategies on SWE-bench Multilingual (Table~\ref{tab:selection_multilingual}). Fixed Rank-$k$ uses the single experience at retrieval rank $k$. Per-Task Best selects among three candidates by the number of successful runs, breaking ties by retrieval rank. Since the same runs are used for selection and reporting, it serves as an outcome-selected reference. Rank-1 achieves the highest average success under Before Execution, whereas Rank-2 performs best under Harness-Triggered. Per-Task Best outperforms the fixed-rank strategies in both settings, suggesting that improved experience selection could further benefit task performance.

\begin{table}[tb]
  \centering
  \small
  \caption{Textual Experience selection on SWE-bench Multilingual (300 tasks).
  Fixed Rank-$k$ uses the single experience at rank $k$; Per-Task Best selects among three candidates
  using evaluation outcomes. All values are percentages.}
  \label{tab:selection_multilingual}
  \begin{tabular}{lrrr}
    \toprule
    Method & Avg. Success & pass@3 & pass\textasciicircum{}3 \\
    \midrule
    \multicolumn{4}{l}{\textit{Before Execution}} \\
    No Experience                   & 60.00 & 73.67 & 45.33 \\
    Textual Experience: Fixed Rank-1             & 64.22 & 72.67 & 56.33 \\
    Textual Experience: Fixed Rank-2             & 61.89 & 72.67 & 51.33 \\
    Textual Experience: Fixed Rank-3             & 63.78 & 74.00 & 54.33 \\
    Textual Experience: Per-Task Best           & 70.56 & 78.67 & 61.67 \\
    \midrule
    \multicolumn{4}{l}{\textit{Harness-Triggered}} \\
    No Experience                   & 60.00 & 73.67 & 45.33 \\
    Textual Experience: Fixed Rank-1             & 63.00 & 72.67 & 53.67 \\
    Textual Experience: Fixed Rank-2             & 64.44 & 72.00 & 56.67 \\
    Textual Experience: Fixed Rank-3             & 63.56 & 72.00 & 54.67 \\
    Textual Experience: Per-Task Best           & 70.00 & 78.67 & 61.67 \\
    \bottomrule
  \end{tabular}
\end{table}

\section{Memory Windowing Experiment}
\label{appendix:infra_experiment_details}

\subsection{Experimental Setup}

The experiment compares two execution schedules for one standalone MoS memory block.
Whole-pack execution produces expert projections for all tokens at once; windowed execution groups complete steps, performs their projections, state updates, and readout in sequence, and recomputes internal intermediates during backward.
Both use the same equations, weights, inputs, precision, and update kernel.
Measurement covers router, all expert projections, memory writes and reads, RMSNorm, SoftCapSiLU, fusion, and backward through these computations.

\begin{table}[htbp]
    \centering
    \caption{Configuration of the MoS memory-block experiment.}
    \label{tab:infra_window_configuration}
    \small
    \renewcommand{\arraystretch}{1.12}
    \begin{tabular}{ll}
        \toprule
        Configuration & Value \\
        \midrule
        Hidden dimension & 2048 \\
        Memory experts & 4 \\
        Memory heads per expert / head dimension & 16 / 256 \\
        Decay-gate / output-gate projection rank & 256 / 256 \\
        Parameter and projection / memory-state precision & BF16 / FP32 \\
        Writer chunk & 64 tokens \\
        Step length & 2048 tokens \\
        Window budget & \makecell[l]{2048 / 4096 tokens,\\or 1 / 2 complete steps} \\
        \bottomrule
    \end{tabular}
\end{table}

Each pack contains two equal-length trajectories: 8K, 16K, and 32K packs comprise two 4K, 8K, and 16K trajectories, respectively.
Each trajectory starts from an independent initial state.
All measured steps commit state, while reads within a step use its incoming snapshot.
Windows do not cross trajectory boundaries.
Weights are randomly initialized and hidden-state inputs are synthetic, with fixed random seed 20261015; input hidden states have standard deviation 1.0, while initial states and backward cotangents have standard deviation 0.01.
This workload measures the computation and activation costs of the memory branch.

Memory updates use a fixed chunked-WY implementation.

\subsection{Measurement Method}

Each pack length and window budget define a paired experiment.
Both variants receive 5 warmups and 20 measured trials; successive pairs alternate between whole-pack$\rightarrow$windowed and windowed$\rightarrow$whole-pack order, yielding 240 measured samples.
Each window budget has an independently measured whole-pack baseline, accounting for the small baseline-time differences between rows at the same length.

CUDA events measure forward-plus-backward (F+B) time, including block execution, inner products of the output and final state with fixed cotangents, and backward through the scalar objective.
Input preparation, schedule construction, and compilation occur outside the timed region.
Peak memory is PyTorch peak allocated, including block parameters, inputs and cotangents, outputs, gradients, and both prepared schedules.
This metric measures live tensor allocations, distinct from memory reserved by the caching allocator.
The main table reports medians; percentage changes are computed from unrounded medians.

\subsection{Full Results}

\begin{table}[htbp]
    \centering
    \caption{Full memory and time results for the MoS memory block.
    Each row is an independently measured whole-pack/window pair; values are medians.}
    \label{tab:infra_window_full_results}
    \small
    \setlength{\tabcolsep}{5pt}
    \renewcommand{\arraystretch}{1.12}
    \begin{tabular}{rrrrrr}
        \toprule
        \makecell{Pack length\\(tokens)} & \makecell{Window budget\\(tokens)} & \makecell{Whole-pack\\memory (GiB)} & \makecell{Windowed\\memory (GiB)} & \makecell{Whole-pack\\F+B (ms)} & \makecell{Windowed\\F+B (ms)} \\
        \midrule
        8192 & 2048 & 12.37 & 4.85 & 102.1 & 122.9 \\
        8192 & 4096 & 12.37 & 8.36 & 102.1 & 117.4 \\
        16384 & 2048 & 24.19 & 5.23 & 211.6 & 246.1 \\
        16384 & 4096 & 24.19 & 8.77 & 211.6 & 235.1 \\
        32768 & 2048 & 47.83 & 6.03 & 462.7 & 493.5 \\
        32768 & 4096 & 47.83 & 9.49 & 463.0 & 470.6 \\
        \bottomrule
    \end{tabular}
\end{table}

As pack length increases from 8K to 32K, whole-pack peak memory grows from 12.37 GiB to 47.83 GiB, while 4096-token windows grow from 8.36 GiB to 9.49 GiB.
Windowing bounds the lifetime of projections and intermediates, while retained outputs and boundary states still grow with length.
Across the three lengths, the median time increases for 4096-token windows are 15.0\%, 11.1\%, and 1.6\%, respectively, showing that windowing overhead depends on the workload.

\subsection{Measurement Variability}

Table~\ref{tab:infra_window_variability} reports medians, interquartile intervals [Q25, Q75], and sample standard deviations, all in ms.

\begin{table}[htbp]
    \centering
    \caption{F+B timing variability in milliseconds.
    Intervals are the 25th and 75th percentiles; standard deviations use all measured samples.}
    \label{tab:infra_window_variability}
    \small
    \setlength{\tabcolsep}{4pt}
    \renewcommand{\arraystretch}{1.12}
    \begin{tabular}{rrlll}
        \toprule
        \makecell{Pack length\\(tokens)} & \makecell{Window budget\\(tokens)} & \makecell[l]{Whole-pack median\\{[Q25, Q75]}} & \makecell[l]{Windowed median\\{[Q25, Q75]}} & \makecell[l]{Standard deviation\\(whole / windowed)} \\
        \midrule
        8192 & 2048 & 102.10 [102.08, 102.12] & 122.87 [122.86, 122.88] & 0.03 / 0.02 \\
        8192 & 4096 & 102.11 [102.10, 102.12] & 117.42 [117.41, 117.45] & 0.05 / 0.03 \\
        16384 & 2048 & 211.55 [211.43, 212.02] & 246.11 [246.02, 246.42] & 37.19 / 0.26 \\
        16384 & 4096 & 211.61 [211.45, 211.87] & 235.07 [234.98, 235.39] & 0.30 / 0.27 \\
        32768 & 2048 & 462.71 [462.37, 463.87] & 493.48 [492.91, 494.38] & 22.22 / 0.94 \\
        32768 & 4096 & 462.96 [462.42, 463.68] & 470.58 [469.84, 471.46] & 0.76 / 38.01 \\
        \bottomrule
    \end{tabular}
\end{table}

All samples are retained, including whole-pack trials at 16K/2048 (378.03 ms) and 32K/2048 (562.40 ms), and a windowed trial at 32K/4096 (640.58 ms).
The 1.6\% reported in the chapter therefore describes the median time difference; individual trials exhibit variability.

\section{Case Studies}
\label{appendix:case_studies}

We present four examples from a long English conversation. They show how MemT-1.0-Nano uses its memory as the history grows.

MemT-1.0-Nano keeps one latent memory slot for each conversation. History enters the slot through commit operations. A commit reads the incoming turns and writes their content into the slot, and the original text is not carried forward.
What the model can answer therefore depends on the writes. The examples below show this at four points in the conversation.
The text inside each box is given exactly as the model produced it.

\subsection{Example 1: Factual Recall at 4K}
\label{appendix:case_4k}

The question asks for three things: the city, the purpose of the trip, and who holds the sample book. The model gives all three. It also keeps the nearby details with their owners, so Ningbo, the red toolbox, and the blue badge do not move to Chen Wei. The colour of the sample book is not mentioned, which is correct because the question does not ask for it.

\begin{promptbox}
History: 4,340 tokens / first 25 turns of the conversation

Turn 1 (user): I'm Lin Yao, and I coordinate exhibitions at
  Qinghe Studio. ... I think I'd rather sit than go straight
  back out.

Turn 2 (user): ... I went to Suzhou to set up an exhibition and
  left my blue sample book with my colleague Chen Wei for
  safekeeping. That was a past work visit, separate from the
  future exhibition I'm still arranging. ...

Turn 3 (user): ... There's a note about Xu Cheng, another
  colleague: he lives in Ningbo, handles lighting, and has the
  red toolbox. ...

[... 21 turns of unrelated conversation about drawing,
cooking, and household objects ...]

Question: Which city did I go to last week, and what was I
  doing there? Who should I ask to get back the sample book I
  handed over?

Answer: You went to **Suzhou** last week. You were there to
  **set up an exhibition**.

  You should ask **Chen Wei** for the sample book.
\end{promptbox}

\subsection{Example 2: Abstention on Missing Information at 16K}
\label{appendix:case_16k}

The question asks for the user's birthday. The conversation never mentions a birthday. The model reports this instead of inventing a date. The answer is only six tokens long, and its brevity is the point. The question invites the model to fill the gap, and this answer does not.

\begin{promptbox}
History: 16,457 tokens / first 105 turns of the conversation

Turn 1 (user): I'm Lin Yao, and I coordinate exhibitions at
  Qinghe Studio. ...

Turn 18 (user): ... The cup beside me has cooled, and the tea
  leaves have collected along one edge because the table isn't
  perfectly level. ...

[... 87 further turns that stay on drawing, cooking, and
walking outside, and never mention a birthday or an age ...]

Turn 103 (user): I once bought a set of colored pencils and
  spent more time arranging them than using them. ...

Turn 105 (user): I kept the stub. It was easier to hold
  sideways for broad marks ... Sometimes a broken object is
  simply broken. ...

Question: What month and day is my birthday again? If we
  haven't talked about it, just say you don't know.

Answer: I don't know.
\end{promptbox}

\subsection{Example 3: Rule Application at 32K}
\label{appendix:case_32k}

The fictional page of the notebook gives two chained rules. A blue badge grants eligibility to book a room. Eligibility to book, together with completed registration, grants eligibility to collect a silver toolbox. The second rule therefore depends on both conditions. If the registration condition is dropped, the rule grants an entitlement that the text never gives. The answer below applies the chain correctly to both characters. It also keeps the two fictional colleagues apart, so Xu Cheng's incomplete registration does not move to Chen Wei.

\begin{promptbox}
History: 32,931 tokens / first 230 turns of the conversation

Turn 4 (user): ... anyone with a blue badge is eligible to book
  Room 2. Anyone eligible to book Room 2 who has completed
  registration is eligible to collect a silver toolbox. There
  are no other eligibility rules. I mean eligibility only:
  being allowed to book or collect something doesn't mean the
  person has actually done it.

Turn 5 (user): The fictional Chen Wei has a blue badge and has
  completed registration. The fictional Xu Cheng also has a
  blue badge but hasn't completed registration yet. ...

[... 224 turns in which the story is never revisited and
neither the rules nor the characters' circumstances are
restated ...]

Turn 229 (user): A description of the sound itself makes more
  sense to me. ...

Turn 230 (user): The radio voice has become my favorite thing
  in this invented film. ...

Question: Given the rules and what we discussed about each
  person, are Chen Wei and Xu Cheng each eligible to collect
  the toolbox now? Briefly explain why.

Answer: Based on the rules provided in the text:

  * **Chen Wei:** Yes, he is eligible. He has a blue badge and
    has completed registration.

  * **Xu Cheng:** No, he is not eligible. Although he has a
    blue badge, he has not completed registration.
\end{promptbox}

\subsection{Example 4: Tracking an Update at 64K}
\label{appendix:case_64k}

The exhibition is first set for Saturday at 3 p.m. in the East Hall. It is then officially changed to Sunday at 4 p.m. The venue, the admission price, and the contact person all stay the same. A move to Monday is also proposed, but it is never decided. A correct answer must report the change and leave the proposal out. The answer below does both. It answers the second half of the question by negation instead of repeating the values. The conversation runs for 508 more turns after the change, and the model still separates a settled change from a passing suggestion.

\begin{promptbox}
History: 65,587 tokens / first 516 turns of the conversation

Turn 6 (user): The small exhibition I'm planning for the future
  was originally set for Saturday at 3 p.m. in the East Hall,
  with free admission, and Chen Wei is the contact person. ...
  I'm waiting for the final timing decision before I send a
  notice, so the draft is still open.

Turn 7 (user): ... the confirmation has just arrived too: the
  exhibition has officially changed to Sunday at 4 p.m.
  Everything else stays the same. This change is definite, so
  that's the new arrangement. ...

Turn 8 (user): ... Someone has just suggested moving the
  exhibition to Monday, but I haven't decided and haven't
  announced another change. It's only a suggestion under
  consideration, not a confirmed arrangement. ...

[... 508 turns in which the exhibition is not raised again; the
closing turns discuss a lamp, a book, and a patch of wall ...]

Turn 514 (user): ... The wall opposite has a little patch where
  the paint catches the light differently. ...

Turn 516 (user): ... My first glance supplies a shadow, then
  the second supplies the patch. ...

Question: When was the exhibition originally scheduled, and
  what time was it officially changed to? Did the venue,
  admission price or contact person change?

Answer: The exhibition was originally scheduled for **Saturday
  at 3 p.m.** It was officially changed to **Sunday at 4 p.m.**

  No, the venue, admission price, or contact person did not
  change. The change was solely to a different time on a
  different day.
\end{promptbox}

\end{document}